\documentclass{article}
\usepackage{enumitem}
\usepackage{arxiv}
\usepackage[utf8]{inputenc} % allow utf-8 input
\usepackage[T1]{fontenc}    % use 8-bit T1 fonts
\usepackage{authblk}
\usepackage[colorlinks,citecolor=blue]{hyperref}       % hyperlinks
\usepackage{url}            % simple URL typesetting
\usepackage{booktabs}       % professional-quality tables
\usepackage{amsfonts}       % blackboard math symbols
\usepackage{nicefrac}       % compact symbols for 1/2, etc.
\usepackage{microtype}     
\usepackage{graphicx}
\usepackage[square,sort,comma,numbers]{natbib}
\usepackage{doi}
\usepackage{multicol}
\usepackage{amssymb,amsmath}
\usepackage[thmmarks,amsmath]{ntheorem}
\usepackage{mathrsfs}
\usepackage{algorithm}
\usepackage{algorithmic}
\usepackage{subfigure}
\usepackage{caption}
\usepackage{multicol}
\usepackage{geometry}
\usepackage{caption}
\title{Training Generative Adversarial Networks with Adaptive Composite Gradient}

\author{Huiqing Qi}
\author{Fang Li}
\author{Shengli Tan}
\author{Xiangyun Zhang \thanks{Corresponding author:xyzhang@math.ecnu.edu.cn}}
\affil{School of Mathematical Sciences, East China Normal University, Shanghai 200241, China}

\renewcommand{\undertitle}{}
\newtheorem{theorem}{Theorem}[section]

\newtheorem{proposition}[theorem]{Proposition}
\newtheorem{definition}[theorem]{Definition}
\qedsymbol{\ensuremath{\square}}{
   \theoremheaderfont{\bfseries}
   \theorembodyfont{\normalfont}
   \theoremsymbol{\ensuremath{\square}}
   \newtheorem*{proof}{Proof}
}
\hypersetup{
pdftitle={paper version of \today},
pdfauthor={Huiqing Qi},
pdfkeywords={First keyword, Second keyword, More},
}
\begin{document} \small
\maketitle

\begin{abstract}
The wide applications of Generative adversarial networks benefit from the successful training methods, guaranteeing that an object function converges to the local minima. Nevertheless, designing an efficient and competitive training method is still a challenging task due to the cyclic behaviors of some gradient-based ways and the expensive computational cost of these methods based on the Hessian matrix. This paper proposed the adaptive Composite Gradients (ACG) method,  linearly convergent in bilinear games under suitable settings. Theory and toy-function experiments suggest that our approach can alleviate the cyclic behaviors and converge faster than recently proposed algorithms. Significantly, the ACG method is not only used to find stable fixed points in bilinear games as well as in general games. The ACG method is a novel semi-gradient-free algorithm since it does not need to calculate the gradient of each step, reducing the computational cost of gradient and Hessian by utilizing the predictive information in future iterations. We conducted two mixture of Gaussians experiments by integrating ACG to existing algorithms with Linear GANs. Results show ACG is competitive with the previous algorithms.  Realistic experiments on four prevalent data sets (MNIST, Fashion-MNIST, CIFAR-10, and CelebA) with DCGANs show that our ACG method outperforms several baselines, which illustrates the superiority and efficacy of our method.
\end{abstract}

% keywords can be removed
%%\keywords{First keyword \and Second keyword \and More}
\begin{multicols}{2}
\section{Introduction}
Gradient Descent-based machine learning and deep learning methods have been widely used in various computer science tasks over the past several decades. Optimizing a single objective problem with gradient descent is easy to converge to a saddle point in some cases \cite{lee2017}. Meanwhile, there is a growing set of multi-objective problems that need to be optimized in numerous fields, such as deep reinforcement learning \cite{Li2020,vezhnevets2017}, Game Theory, Machine Learning and Deep Learning. Generative Adversarial Networks \cite{goodfellow2014} is a kind of classical multi-objective problem in Deep Learning. GANs have a wide range of applications \cite{Hong2019} because of their capability, which can learn to generate complex and high dimensional target distribution. The existing literature about GANs can be divided into  four categories, including music generation \cite{guimaraes2018,lee2018, yu2017}, natural languages \cite{hsu2017,lin2018,haidar2019,Croce2020}, methods of training GANs \cite{neyshabur2018,qin2020,Peng_2020,keke2020},images processing \cite{ledig2017,wu2017,zhu2020,Wang_2018}. GANs have obtained remarkable progress in image processing, such as  video generation \cite{walker2017,NIPS2016_04025959,tulyakov2017}, noise removal \cite{yue2020dual}, deblur \cite{DeblurGAN}, image to image translation \cite{kim2017,yi2018}, image super-resolution \cite{ledig2017}, medical image processing \cite{dai2017,yang2017,8417964}.

Generative adversarial networks’ framework consists of two deep neural networks: generator network and discriminator network correspondingly. The generator network is given a noise sample from a simple known distribution as input, and it can produce a fake sample as output. The generator learns to make such fake samples, not by directly using real data, just by adversarial training with a discriminator network.  The essence of GANs is a zero-sum game between the generator and discriminator. The object function of GANs \cite{goodfellow2014} is often formulated as a two-player min-max game with a Nash equilibrium at the saddle points:
\end{multicols}
\vspace{-0.6cm}
\begin{equation}
    \mathop{\min}\limits_{G} \mathop{\max}\limits_{D} V(D,G) = \mathbb{E}_{x \sim P_{X}(x)}[\log D(x)] + \mathbb{E}_{z \sim P_Z(z)}[\log(1-D(G(z)))].
\end{equation}

\begin{figure}[H]
\centering
\subfigure{\includegraphics[width=0.48\linewidth]{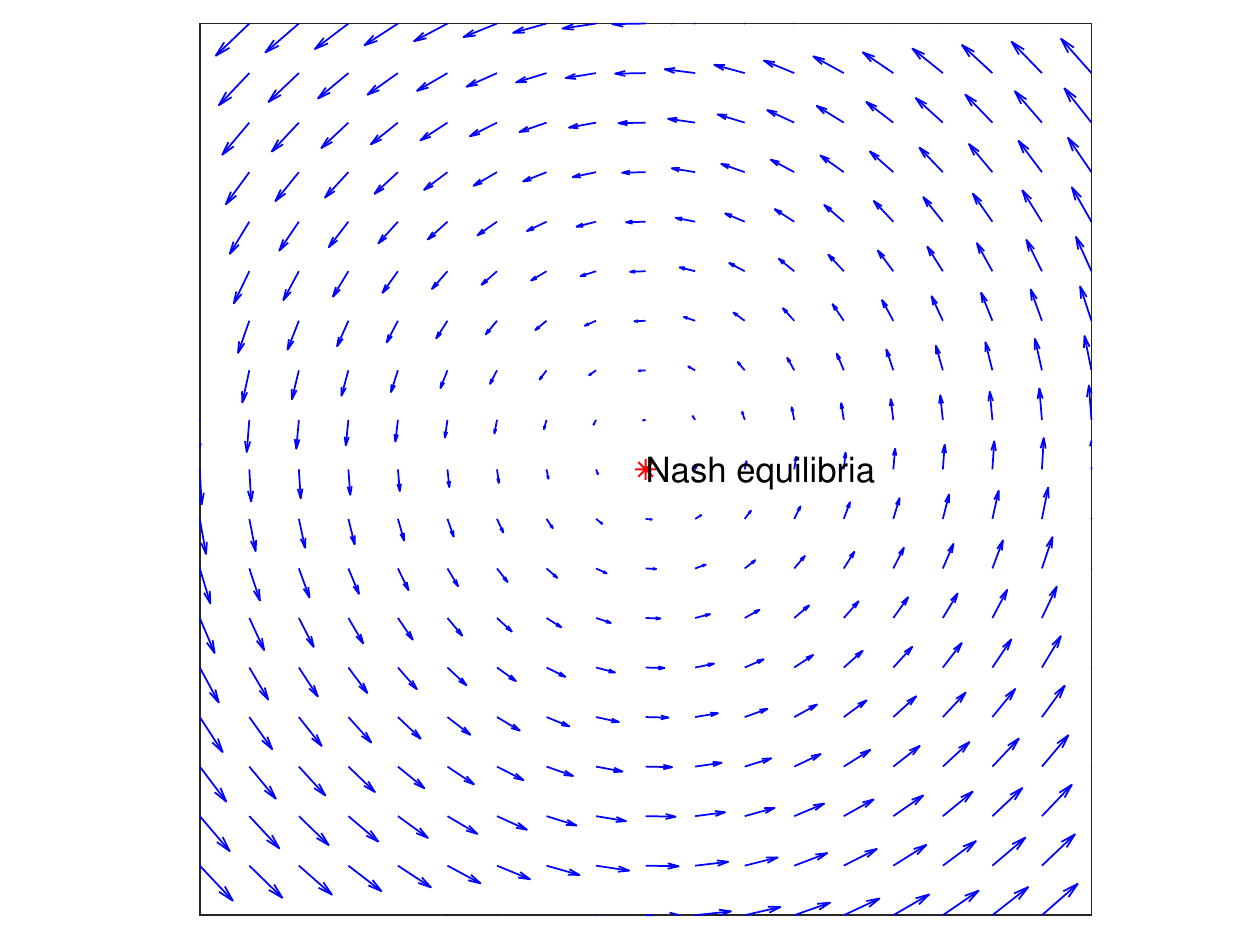}}
\hfill
\subfigure{\includegraphics[width=0.48\linewidth]{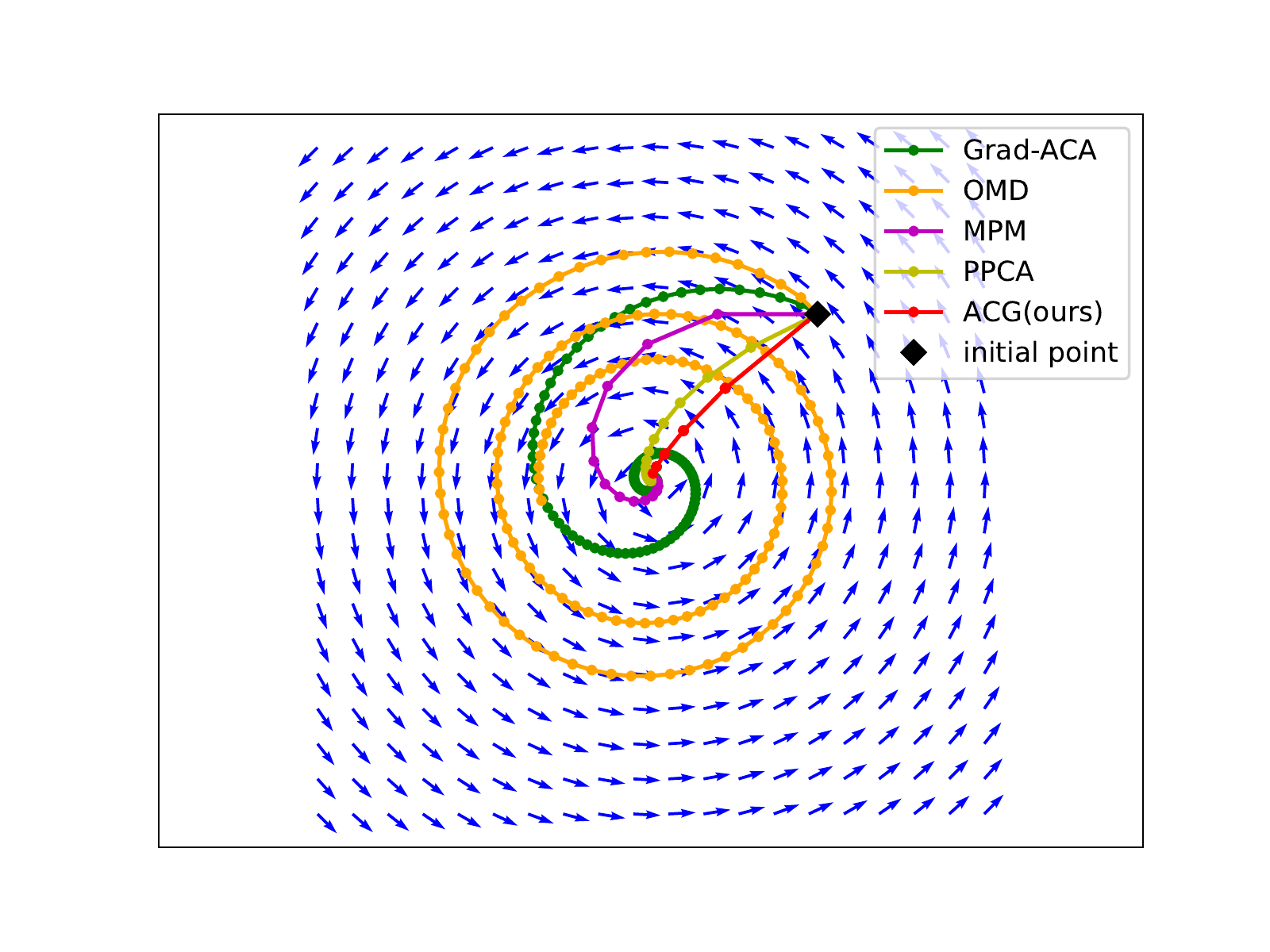}}
\captionsetup{font={small}}
\caption{Left: the strong gradients rotational filed around Nash equilibria. Right: comparison of convergence behaviors among several recently proposed methods. It is obvious ours ACG method converges faster than others. For more details in Section 6.1.}
\label{fig:fig1} 
\end{figure}

Where $x\sim P_{X}(x)$ denotes an actual data sample and $z\sim P_{Z}(z)$ denotes a sample from a noise distribution(often using uniform distribution or Gaussian distribution). More different forms of GANs object function are mentioned in \cite{wang2020}. Though GANs have achieved remarkable applications, training stable and fast GANs \cite{odena2019,goodfellow2017} still is a challenging task. Since it suffers from the strongly associated gradient vector field rotating around a Nash equilibrium (see Figure \ref{fig:fig1}). Moreover, those gradient descent ascent-based methods used to optimize object function of GANs tend to lead the limit oscillatory behavior because of imaginary components in the Jacobian eigenvalues. 
\begin{multicols}{2}
In recent years, there are great amount of remarkable studies on proposing novel algorithms for training GANs. \cite{qin2020} considered the dynamics as a continuous-time process and proposed using ordinary differential equations to train GANs. Consensus optimization \cite{mescheder2018} with Jacobian information diverts gradient updates to the descent direction of the field magnitudes. \cite{pmlr-v80} developed a method called Symplectic Gradient Adjust(SGA).  Motivated by SGA, \cite{Peng_2020} propose centripetal acceleration method and their altering centripetal acceleration version. \cite{keke2020} based on some predictive methods \cite{pmlr-v108,9054014,chavdarova2020,gidel2020,pmlr-v108M,zhang2020} proposed the projection predictive gradient centripetal method.

The main idea of this work is to reduce the computing cost of the Hessian matrix in consensus optimization and SGA. Motivated by \cite{Peng_2020,keke2020} and \cite{2019Liang}, we propose a novel Adaptive Composite Gradient method, which can be used to calibrate and accelerate the traditional methods such as SGD, RMSProp, Adam, consensus optimization, and SGA. The ACG method exploits three aspects of information in the iteration process, which includes gradients information of the past iteration steps, adaptive and predictive information for future iteration steps, and the projection information of the current iteration step mentioned in \cite{keke2020}. We fuse this information as the composite gradient to update the parameters in our algorithm, which can be deployed in deep networks and used to train GANS.

\textbf{Contributions.} \quad The main contributions of this paper are as follows:
\begin{itemize}[leftmargin=*]
\item We proposed the Adaptive Composite Gradient (ACG) method, which can alleviate the cyclic behaviors around the Nash equilibria in games. Meanwhile, we prove its convergence in bilinear games. Our algorithm can be used to train GANs.
\item Our ACG method is not only applied in bilinear games but also used in general game problems. Furthermore, we experimentally demonstrate applicability for three-player game problems by toy model.
\item The Adaptive Composite Gradient method can reduce the computing cost of the Hessian when it calibrates SGA and consensus optimization or some other Hessian-based methods. And also, ACG can reduce the computing cost of gradients when calibrating gradient descent-based methods. In other words, our method is a novel semi-gradient-free algorithm.
\item We conducted sufficient numerical simulations in training GANs and Deep Convolutional Generative Adversarial Networks, demonstrating that our method can be used to train ordinary GANs and deployed into a  generic deep convolution network framework.
\end{itemize}
\vspace{-0.3cm}

\section{Related Work}
There are several distinctive approaches to improve the training of GANs, but they show more or fewer limitations in some cases. Some of these are dependent closely on the previous assumptions, which leads to these methods not being valid. Moreover, some of these need to payoff the computing cost of the Hessian in the dynamics. Next, we will discuss some related researches in this section.

\textbf{Symplectic Gradient Adjustment (SGA) \cite{pmlr-v80}:} Compared with the traditional games, do not constrain the players' parameter sets or require the loss functions to be convex. The general games can be decomposed into a potential game, and a Hamiltonian game in \cite{pmlr-v80}. To introduce our method, we first recall the SGA method as follows.
\vspace{-0.1cm}
\begin{definition}\label{D1}
  A game is a set of players $[p] = \lbrace 1,2,...,n \rbrace $, and the loss functions satisfy twice continuously differentiable $\lbrace \ell_{i}: \mathbb{R}^{d} \rightarrow \mathbb{R} \rbrace_{i=1}^{n}$. Players' parameters are $w = (w_{1},w_{2},...,w_{n}) \in \mathbb{R}^{d}$  with $w_{i} \in \mathbb{R}^{d_{i}} $ where $\sum_{i}^{n}d_{i}=d$. The $i^{th}$ player can control $w_{i}$.
\end{definition}
\vspace{-0.2cm}
Using the $g(w)$ notes the simultaneous gradient which is the gradient of the losses withe respect to players' parameters  $g(w) = (\nabla_{w_{1}} \ell_{1},\nabla_{w_{2}} \ell_{2},...,\nabla_{w_{n}} \ell_{n})$. For a bilinear game, it requires the losses to satisfy $\sum_{i=1}^{n} \ell_{i} \equiv 0$ such as follow:
$$\ell_{1}(\textup{x},\textup{y}) = \textup{x}^{\mathrm{T}} \textup{C} \textup{y}   \quad and \quad   \ell_{2}(\textup{x},\textup{y}) = -\textup{x}^{\mathrm{T}} \textup{C} \textup{y}$$
This kind of games have a Nash equilibrium at $(\textup{x},\textup{y}) = (\textup{\textbf{0}},\textup{\textbf{0}})$. The simultaneous gradient $g(\textup{x},\textup{y}) = ( \textup{C} \textup{y},-\textup{C}^{\mathrm{T}} \textup{x})$ rotates around the Nash equilibrium shown in Figure \ref{fig2}.

We can derive the Hessian of a $n$-players game with the simultaneous gradient $g(w)$. The formula of  Hessian is $\textup{H}(w) =\nabla_{w} \cdot g(w)^{\mathrm{T}}=(\frac{\partial g_{i}(w)}{\partial w_{j}})_{i,j=1}^{d}$, where $\textup{H} \in \mathbb{R}^{d\times d}$. Further, the matrix formula of Hessian is as follows:
\begin{equation}
\textup{H}(w)=\left[
\begin{array}{cccc}
    \nabla_{w_{1}}^{2} \ell_{1}  &  \nabla_{w_{2},w_{1}}^{2} \ell_{2} & \cdots & \nabla_{w_{n},w_{1}}^{2} \ell_{n}\\
    \nabla_{w_{1},w_{2}}^{2} \ell_{1} & \nabla_{w_{2}}^{2} \ell_{2}  & \cdots  & \nabla_{w_{n},w_{2}}^{2} \ell_{n}\\
    \vdots & \vdots & \ddots & \vdots \\
    \nabla_{w_{1},w_{n}}^{2} \ell_{1} & \nabla_{w_{2},w_{n}}^{2} \ell_{2}  & \cdots  & \nabla_{w_{n}}^{2} \ell_{n}
\end{array}
\right]\label{Hess}.
\end{equation}
Applying the generalized Helmholtz decomposition[\textbf{Lemma 1} in \cite{pmlr-v80}] to the above mentioned Hessian of the game $\textup{H}(w) = \textup{S}(w)+\textup{A}(w)$. David et al(2018) \cite{pmlr-v80} pointed that a game is a potential game if $\textup{A}(w) \equiv 0$. It is a Hamiltonian game if $\textup{S}(w) \equiv 0$. Potential games and Hamiltonian games are both well-studied, and they are easy to solve. Since the cyclic behavior around the Nash equilibrium is caused by simultaneous gradient, David et al \cite{pmlr-v80} proposed the Symplectic Gradient Adjustment method which is as follows:$$g_{\lambda}:=g+\lambda \cdot \textup{A}^{\mathrm{T}}g.$$ Where $\textbf{A}$ is from the Helmholtz decomposition of Hessian. $g_{\lambda}$ is used to replace the gradient among the iterates in the GAD-based methods, and using $g_{\lambda}$ to train GANs can alleviate the cyclic behaviors. If we consider the players in a bilinear game as GANs, the SGA algorithm needs to pay an expensive computing cost of Hessian which can lower the algorithm efficiency.

\textbf{Centripetal Acceleration\cite{Peng_2020}:}  The simultaneous gradient shows cyclic behaviors around the Nash. Hamiltonian games obey a conservation law in these gradient descent-based methods, so the cyclic behaviors can be considered a uniform circular motion process. As well known, the direction of the centripetal acceleration of a consistent circular motion process points to the center of the circle. Using this characteristic modifies the direction of the simultaneous gradient vector field to alleviate the cyclic behaviors. Based on these observations, Peng et al. (2020) \cite{Peng_2020} propose the Centripetal acceleration methods, which are derived to two versions methods named Simultaneous Centripetal Acceleration (Grad-SCA) and  Alternating Centripetal Acceleration (Grad-ACA) which are used to train GANs. 

\begin{figure}[H]
    \centering
    \includegraphics[width=0.95\linewidth]{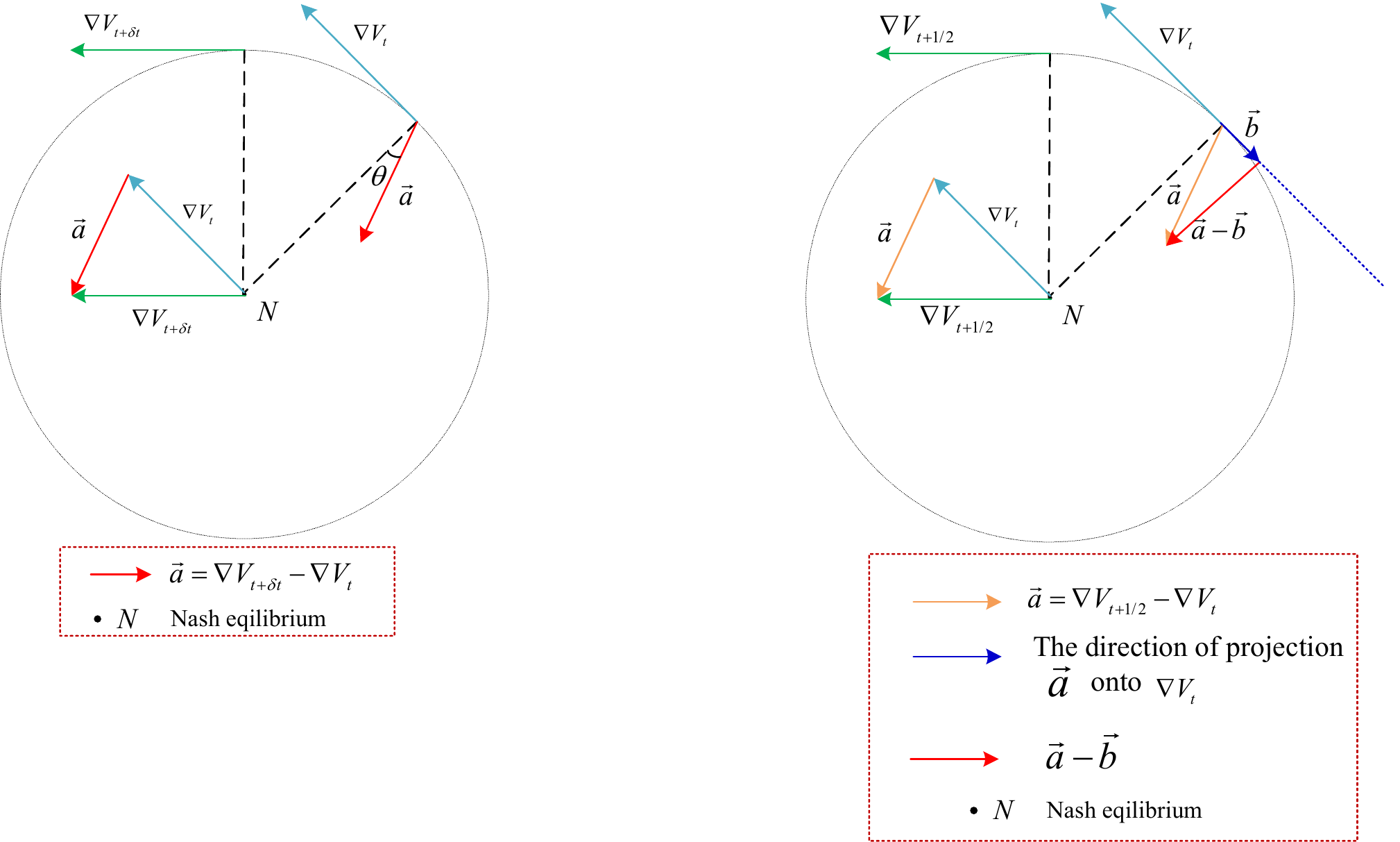}
    \captionsetup{font={small}}
    \caption{Left: the basic intuition of centripetal acceleration methods in \cite{Peng_2020}.Right: the basic intuition of PPCA methods in \cite{keke2020}}
    \label{fig:int}
\end{figure}

Given a bilinear game, the losses of this game are $\ell_{1}(\theta, \phi)$ , $\ell_{2}(\theta, \phi)$ corresponding to player 1 and player 2. The parameter space is defined in $\theta \times \phi$, where $\theta,\phi \in \mathbb{R}^{n}$. Player 1 can control the parameter $\theta$ and tries to minimize payoff function $\ell_{1}$ while player 2 can control parameter $\phi$ and tries to minimize payoff function $\ell_{2}$ under the non-cooperative situation. The game is a process in which the two players adjust their parameters to find a local Nash equilibrium that satisfies the following two requirements:
$$\theta^{'} \in \mathop{\arg} \mathop{\min} \limits_{\theta}  \ell_{1}(\theta, \phi^{'}) \quad and \quad  \phi^{'} \in \mathop{\arg} \mathop{\min} \limits_{\phi} \ell_{2}(\theta^{'}, \phi). $$
The centripetal acceleration methods require that the two-player game is differentiable. Then, the above two payoff functions can be combined into a joint payoff function because of the zero-sum property of the game: 
$$(\theta^{'}, \phi^{'}) \in \mathop{\min} \limits_{\theta} \mathop{\max}\limits_{\phi} \mathbf{V}(\theta, \phi).$$
In order to introduce the Centripetal Acceleration methods, we fist review the simultaneous gradient descent method in \cite{NIPS2016} is $$\theta_{t+1} = \theta_{t} - \alpha \nabla_{\theta}\mathbf{V}(\theta_{t}, \phi_{t}), \quad \phi_{t+1} = \phi_{t} + \alpha \nabla_{\phi}\mathbf{V}(\theta_{t}, \phi_{t}).$$
And the simultaneous gradient descent based alternating version is $$\theta_{t+1} = \theta_{t} - \alpha \nabla_{\theta}\mathbf{V}(\theta_{t}, \phi_{t}), \quad \phi_{t+1} = \phi_{t} + \alpha \nabla_{\phi}\mathbf{V}(\theta_{t+1}, \phi_{t}),$$
where $\alpha$ is learning rate in the algorithms.
While the Centripetal Acceleration methods directly utilize the item of centripetal acceleration to adjust the simultaneous gradient descent. Then gradient descent with simultaneous centripetal acceleration is introduced as:
\begin{align*}
    \theta_{t+1}  = \theta_{t}-\alpha_{1} \nabla_{\theta}&\mathbf{V}(\theta_{t}, \phi_{t})\\
    &-\beta_{1}[\nabla_{\theta}\mathbf{V}(\theta_{t}, \phi_{t}) - \nabla_{\theta}\mathbf{V}(\theta_{t-1}, \phi_{t-1})]\\
    \phi_{t+1} = \phi_{t} +\alpha_{2} \nabla_{\phi}&\mathbf{V}(\theta_{t}, \phi_{t})\\
    &+ \beta_{2}[\nabla_{\phi}\mathbf{V}(\theta_{t}, \phi_{t}) - \nabla_{\phi}\mathbf{V}(\theta_{t-1}, \phi_{t-1})].
\end{align*}
We can also obtain the gradient descent with the alternating centripetal acceleration method:
\begin{align*}
    \theta_{t+1} = \theta_{t}-\alpha_{1} \nabla_{\theta}&\mathbf{V}(\theta_{t}, \phi_{t})\\
    &-\beta_{1}[\nabla_{\theta}\mathbf{V}(\theta_{t}, \phi_{t}) - \nabla_{\theta}\mathbf{V}(\theta_{t-1}, \phi_{t-1})]\\
    \phi_{t+1} = \phi_{t} +\alpha_{2} \nabla_{\phi}&\mathbf{V}(\theta_{t+1}, \phi_{t})\\
    &+ \beta_{2}[\nabla_{\phi}\mathbf{V}(\theta_{t+1}, \phi_{t}) - \nabla_{\phi}\mathbf{V}(\theta_{t}, \phi_{t-1})],
\end{align*}
where $\alpha_{1},\beta_{1},\alpha_{2},\beta_{2}$ in above two versions of the centripetal acceleration methods are hyper-parameters. The centripetal acceleration methods can calibrate other gradient-based methods. The intuitive understanding of the centripetal acceleration method is shown in Figure \ref{fig:int}.

\textbf{Predictive Projection Centripetal Acceleration (PPCA) \cite{keke2020}:} From the viewpoint of centripetal acceleration methods, it use the last iterative step information to update $(\theta_{t+1}, \phi_{t+1})$. However, there are some methods which utilize the predictive step information to update $(\theta_{t+1}, \phi_{t+1})$, such as MPM,OMD,OGDA. MPM is introduced by Liang et al.(2019) \cite{liang19b} and its dynamics are as follows:
\begin{equation*}
\begin{split}
    predictive \quad step: \quad \theta_{t+\frac{1}{2}} &= \theta_{t}-\alpha \nabla_{\theta} \mathbf{V}(\theta_{t}, \phi_{t})\\
                     \phi_{t+\frac{1}{2}} &= \phi_{t}+\alpha \nabla_{\phi} \mathbf{V}(\theta_{t}, \phi_{t});\\
    gradient \quad step: \quad \theta_{t+1} &= \theta_{t}-\beta \nabla_{\theta} \mathbf{V}(\theta_{t+\frac{1}{2}}, \phi_{t+\frac{1}{2}})\\
                         \phi_{t+1} &= \phi_{t}+\beta \nabla_{\phi} \mathbf{V}(\theta_{t+\frac{1}{2}}, \phi_{t+\frac{1}{2}})
\end{split}
\end{equation*}
Motivated by MPM and centripetal acceleration methods, Li et al.(2020) \cite{keke2020} propose the predictive projection centripetal acceleration methods. They also consider approximately the cyclic behavior around a Nash as a uniform circular motion process. However, it is different from the Grad-SCA and Grad-ACA. They construct the item of centripetal acceleration to use the predictive step information replacing that of the last step. Meanwhile, they argue that the approximated centripetal acceleration term points to the matched center. To make the centripetal acceleration item point to the center precisely, they propose the projection centripetal acceleration term at time $t$. PPCA can modify the gradient descent ascent and the alternating gradient descent ascent directly. We can understand PPCA intuitively from Figure \ref{fig:int}. The dynamics of predictive projection centripetal acceleration are the following formula:
\vspace{-0.1cm}
\begin{align*}
    predictive \; step: \; \theta_{t+\frac{1}{2}} = \theta_{t}-\gamma \nabla_{\theta} \mathbf{V}(\theta_{t}, \phi_{t})&;\\
                     \phi_{t+\frac{1}{2}} = \phi_{t}+\gamma \nabla_{\phi} \mathbf{V}(\theta_{t}, \phi_{t})&;\\
    gradient \; step: \; \begin{pmatrix} \theta_{t+1}\\ \phi_{t+1} \end{pmatrix} = \begin{pmatrix} \theta_{t}\\ \phi_{t} \end{pmatrix} + \alpha \nabla \Bar{\mathbf{V}}(\theta_{t}, \phi_{t})&\\ 
    +\beta \{ [\nabla \Bar{\mathbf{V}}(\theta_{t+\frac{1}{2}}, \phi_{t+\frac{1}{2}})-\nabla \Bar{\mathbf{V}}(\theta_{t}, \phi_{t})]&\\
                             - \prod_{\nabla \Bar{\mathbf{V}}(\theta_{t}, \phi_{t})} [\nabla \Bar{\mathbf{V}}(\theta_{t+\frac{1}{2}}, \phi_{t+\frac{1}{2}}) - \nabla \Bar{\mathbf{V}}(\theta_{t}, \phi_{t})] \}&.
\end{align*}
Where $\nabla \Bar{\mathbf{V}}(\theta_{t}, \phi_{t}) = \begin{pmatrix} -\nabla_{\theta} \mathbf{V}(\theta_{t}, \phi_{t})\\ \nabla_{\phi} \mathbf{V}(\theta_{t}, \phi_{t}) \end{pmatrix} $ is the signed gradient vector at time $t$. The $\prod_{\nabla \Bar{\mathbf{V}}(\theta_{t}, \phi_{t})} [\nabla \Bar{\mathbf{V}}(\theta_{t+\frac{1}{2}}, \phi_{t+\frac{1}{2}}) - \nabla \Bar{\mathbf{V}}(\theta_{t}, \phi_{t})] $ is the projection of the centripetal acceleration term $[\nabla \Bar{\mathbf{V}}(\theta_{t+\frac{1}{2}}, \phi_{t+\frac{1}{2}})-\nabla \Bar{\mathbf{V}}(\theta_{t}, \phi_{t})]$ onto the vector $\nabla \Bar{\mathbf{V}}(\theta_{t}, \phi_{t}).$ Li et al(2020) propose two versions of the PPCA methods by constraining the coefficient matrix which must be full rank in bilinear games under the specified situation [\textbf{Lemma 3.2 in \cite{keke2020}}]. The form of PPCA method for bilinear game is 
\begin{align*}
    predictive \; step: \; \theta_{t+\frac{1}{2}} = \theta_{t}-\gamma \nabla_{\theta} \mathbf{V}(\theta_{t}, \phi_{t})&;\\
                     \phi_{t+\frac{1}{2}} = \phi_{t}+\gamma \nabla_{\phi} \mathbf{V}(\theta_{t}, \phi_{t})&;\\
    gradient \; step: \; \theta_{t+1} = \theta_{t}-\alpha \nabla_{\theta}\mathbf{V}(\theta_{t}, \phi_{t})&\\
    - \beta[\nabla_{\theta}\mathbf{V}(\theta_{t+\frac{1}{2}}, \phi_{t+\frac{1}{2}}) -\nabla_{\theta}&\mathbf{V}(\theta_{t}, \phi_{t})];\\
    \phi_{t+1} = \phi_{t}+\alpha \nabla_{\phi}\mathbf{V}(\theta_{t}, \phi_{t})&\\
    + \beta[\nabla_{\phi}\mathbf{V}(\theta_{t+\frac{1}{2}}, \phi_{t+\frac{1}{2}}) -\nabla_{\phi}&\mathbf{V}(\theta_{t}, \phi_{t})].
\end{align*}
And also, we can get the alternating PPCA formula is as follow:
\begin{align*}
    predictive \; step: \; \theta_{t+\frac{1}{2}} = \theta_{t}-\gamma \nabla_{\theta} \mathbf{V}(\theta_{t}, \phi_{t});&\\
                     \phi_{t+\frac{1}{2}} = \phi_{t}+\gamma \nabla_{\phi} \mathbf{V}(\theta_{t}, \phi_{t});&\\
    gradient \; step: \;  \theta_{t+1} = \theta_{t}-\alpha \nabla_{\theta}\mathbf{V}(\theta_{t}, \phi_{t})&\\
    - \beta[\nabla_{\theta}\mathbf{V}(\theta_{t+\frac{1}{2}}, \phi_{t+\frac{1}{2}}) -\nabla_{\theta}&\mathbf{V}(\theta_{t}, \phi_{t})];\\
    \phi_{t+1} = \phi_{t}+\alpha \nabla_{\phi}\mathbf{V}(\theta_{t+1}, \phi_{t})&\\
    + \beta[\nabla_{\phi}\mathbf{V}(\theta_{t+\frac{1}{2}}, \phi_{t+\frac{1}{2}}) -\nabla_{\phi}&\mathbf{V}(\theta_{t}, \phi_{t})];
\end{align*}
where the all of $\gamma,\alpha,\beta$ are hyper parameters.
\vspace{-0.5cm}
Although the mentioned methods have achieved significant work in training GANs, some require high computing costs and high computer memory. The rest of them depend closely on the approximate circular motion process. If the practical numerical experiments do not satisfy this approximate theory, these methods will not be valid. In contrast to our approach, we propose the adaptive composite gradient method, which can reduce the computing cost and solve the limitation brought by the approximate circular motion process.
\begin{figure}[H]
    \centering
    \includegraphics[width=0.95\linewidth]{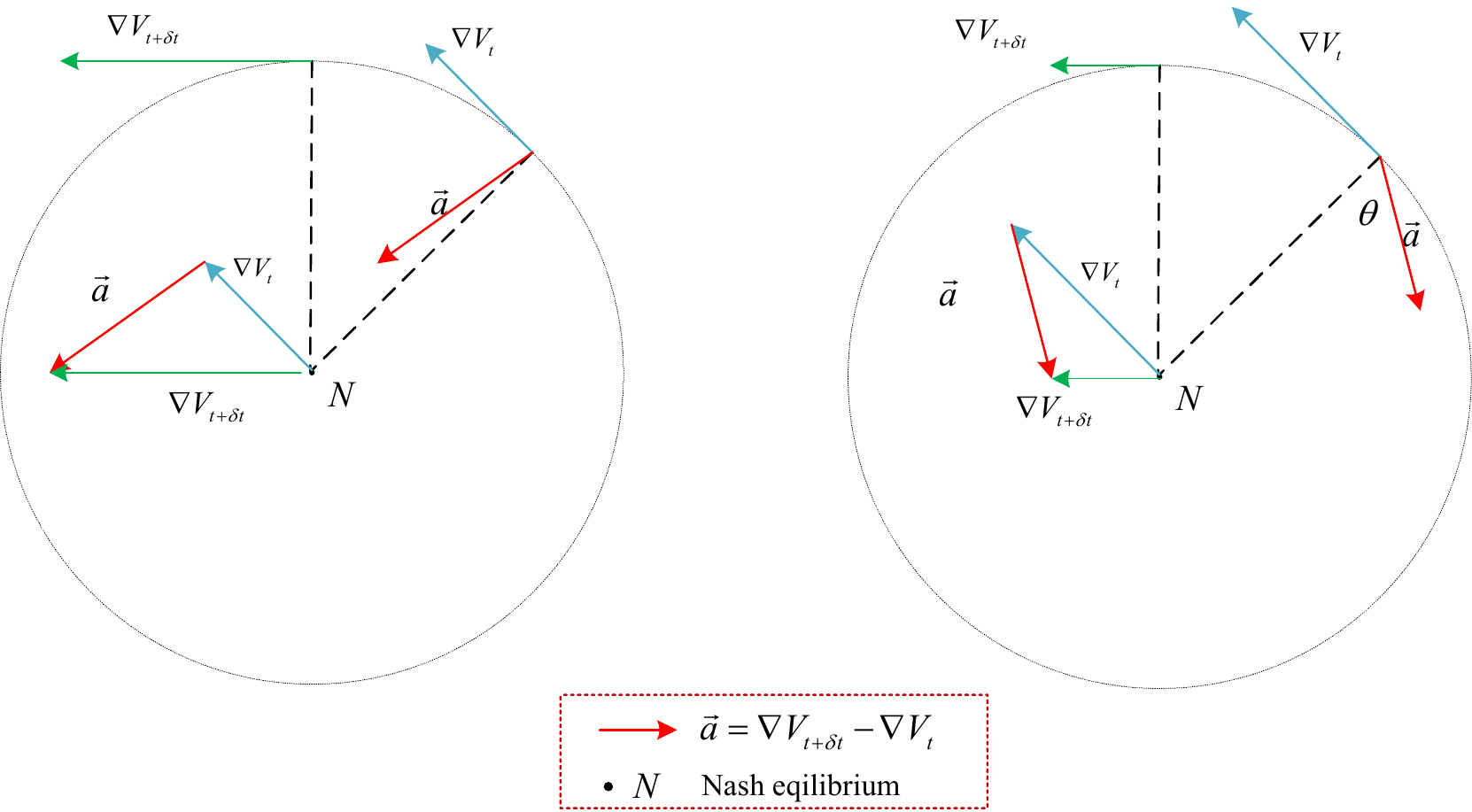}
    \captionsetup{font={small}}
    \caption{The limitations of centripetal acceleration method. Left: case:$|\nabla V_{t+\delta t}|>|\nabla V_{t}|$, Right: case:$|\nabla V_{t+\delta t}|<|\nabla V_{t}|$}
    \label{fig:limt1}
\end{figure}
\vspace{-0.4cm}
\section{Motivation}
In this section, firstly, we mainly focus on the limitations of mentioned methods in Section 2. Then, we describe the theory that motivates our proposed method in the next section. 
\subsection{Limitation Analysing}
These Hessian-based methods are used to optimize $n$ players' game problem, which bring high computing costs, such as Consensus Optimization and Symplectic Gradient Adjustment. The dynamics of SGA is $g_{\lambda}:=g+\lambda \cdot \textup{A}^{\mathrm{T}}g $. Before updating the parameters, the SGA method must obtain the Hessian matrix. However, the time complexity of computing Hessian is $O(n^{3})$ and space complexity of that is $O(n^{2})$ just for one layer in the deep neural network. Supposing there are generative adversarial networks with $m$ depth and $n$ width, the maximum iteration number is $I$. It is well known the costs and computer memory of computing Hessian are expensive. Compared with our method, we use predictive information to update the dynamics, reducing the computing cost of gradients and improving the efficiency of training deep networks because of our method's semi-gradient-free characteristic.

The Centripetal Acceleration methods depend closely on the cyclic behavior around a Nash equilibrium which is approximately considered a uniform circular motion process around the origin. In contrast, the realistic experiments show that the cyclic behavior is not a uniform circular motion process. The centripetal acceleration methods can change the direction of the gradient vector field. It makes the direction far away from Nash equilibrium under exceptional cases shown in  Figure \ref{fig:limt1}. If the assumption is not satisfied, the Centripetal Acceleration method will be invalid.

The PPCA method is an improved version Centripetal Acceleration method. The PPCA method also argues that the cyclic behavior around the origin is approximated to a circular motion process. The PPCA method uses the projection of the centripetal acceleration item, which points to the origin precisely to make up the limitation of Centripetal Acceleration methods, as shown in Figure \ref{fig:int}. But the PPCA method is reduced to be centripetal acceleration method with the full rank coefficient matrix $\mathbf{A}$ in bilinear games (shown in Figure \ref{fig:sp}), since the projection of the PPCA method is zero [\textbf{Lemma 3.2 in \cite{keke2020}}]. And others situations are not discussed in the PPCA method. Meanwhile, the Centripetal Acceleration methods and PPCA methods are only applied in these two-player games. Our proposed Adaptive Compose Gradient method can apply to $n$-player games.
\vspace{-0.2cm}
\subsection{Motivational Theory}
Our idea is motivated by $\mathbf{A^{3}DMM}$\cite{2019Liang}. Then, we review the $\mathbf{A^{3}DMM}$ method. Give a optimisation problem$$ \mathop{\min}\limits_{x \in \mathbb{R}^{n}, y \in \mathbb{R}^{m}} \mathbf{R}(x) + \mathbf{J}(y) \quad s_{\cdot} t_{\cdot} \quad \mathbf{A}x+\mathbf{B}y=b,$$ where the essential assumptions are proposed 
\begin{itemize}[leftmargin=*]
    \item $\mathbf{R} \in \Gamma_{0}(\mathbb{R}^{n})$ and $\mathbf{J} \in \Gamma_{0}(\mathbb{R}^{m})$ are proper convex and lower semi-continuous functions.
    \item $\mathbf{A},\mathbf{B}$ are injective linear operators.
    \item $ri(dom(\mathbf{R}) \cap dom(\mathbf{J})) \neq \varnothing$ and the set of minimizers is non-empty.
\end{itemize}
In order to derive the $\mathbf{ADMM}$ iteration, consider the augmented Lagrangian to rewrite the optimisation problem which reads$$\mathcal{L}(x,y;\Psi) \overset{def}{=} \mathbf{R}(x) + \mathbf{J}(y) + \langle \Psi,\mathbf{A}x+\mathbf{B}y-b \rangle +\frac{\gamma}{2} \| \mathbf{A}x+\mathbf{B}y-b \|^{2}.$$
Where $\gamma > 0$ and $\Psi$ is the Lagrangian multiplier, then we can derive the $\mathbf{ADMM}$ iteration form as follow:
\begin{equation*}
\begin{split}
    x_{k} &= \mathop{\arg\min}_{x \in \mathbb{R}^{n}} \mathbf{R}(x)+\frac{\gamma}{2} \| \mathbf{A}x+\mathbf{B} \textup{y}_{k-1}-b + \frac{1}{\gamma} \Psi_{k-1} \|^{2} \\
    \textup{y}_{k} &= \mathop{\arg\min}_{y \in \mathbb{R}^{m}} \mathbf{J}(y)\\
    &+\frac{\gamma}{2} \| \mathbf{A}x_{k}+\mathbf{B}y-b + \frac{1}{\gamma} \Psi_{k-1} \|^{2}\\
    \Psi_{k} &= \Psi_{k-1} +\gamma(\mathbf{A}x_{k}+\mathbf{B} \textup{y}_{k}-b).
\end{split}
\end{equation*}
We can rewrite the above iteration into the following formula by introducing a new variable $Z_{k}\overset{def}{=} \Psi_{k-1}+\gamma \mathbf{A}x_{k}$,
\begin{equation*}
\begin{split}
    x_{k} &= \mathop{\arg\min}_{x \in \mathbb{R}^{n}} \mathbf{R}(x)+\frac{\gamma}{2} \| \mathbf{A}x - \frac{1}{\gamma}(Z_{k-1} -2\Psi_{k-1} \|^{2} \\
    Z_{k} &= \Psi_{k-1}+\gamma \mathbf{A}x_{k}\\
    \textup{y}_{k} &= \mathop{\arg\min}_{y \in \mathbb{R}^{m}} \mathbf{J}(y)+\frac{\gamma}{2} \| \mathbf{B}y + \frac{1}{\gamma} (Z_{k}-\gamma b) \|^{2}\\
    \Psi_{k} &= Z_{k}+\gamma(\mathbf{B}\textup{y}_{k}-b).
\end{split}
\end{equation*}
\begin{figure}[H]
\centering
\subfigure{\includegraphics[width=0.55\linewidth]{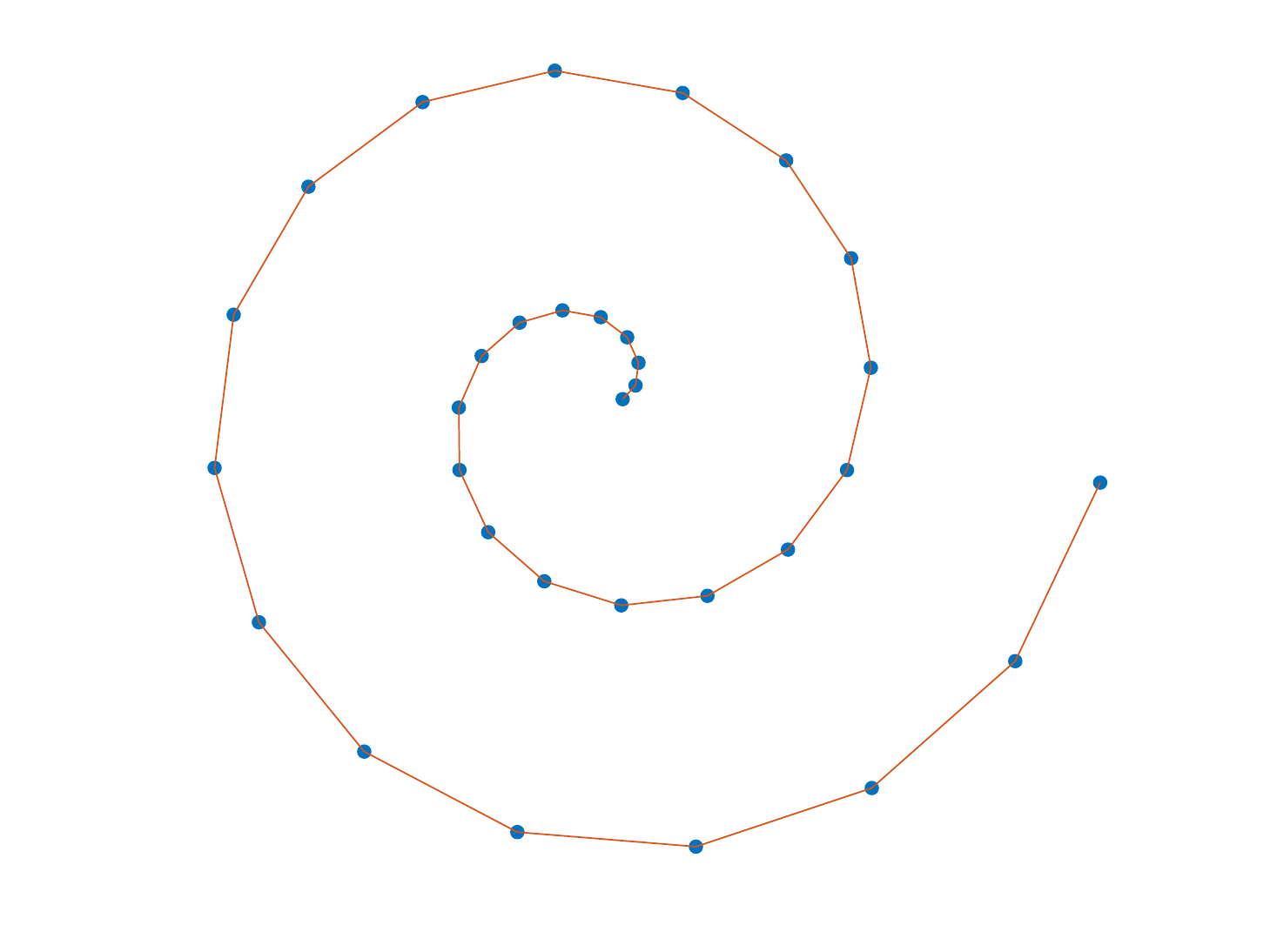}}
\hfill
\subfigure{\includegraphics[width=0.35\linewidth]{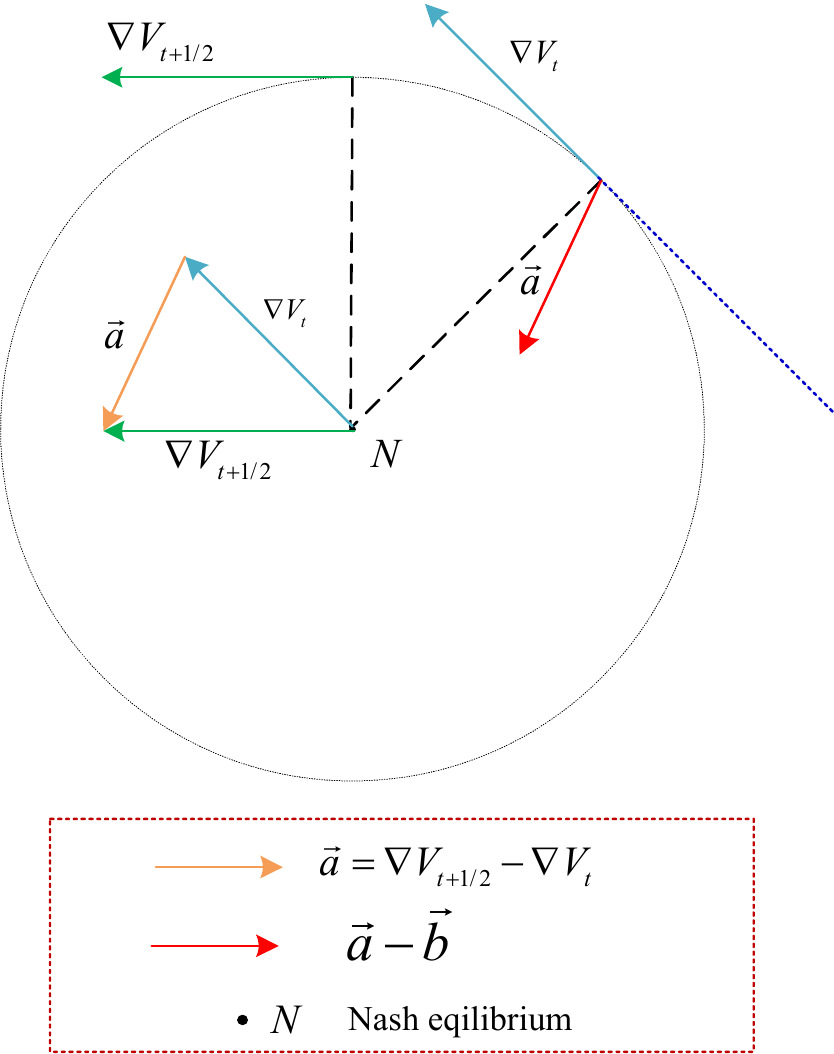}}
\captionsetup{font={small}}
\caption{Left: The spiral trajectory of $Z_{k}$. Right: Degenerated PPCA method in case $|\vec{b}|=0$. The degenerated PPCA is the same as centripetal acceleration method, if $\delta t = 1/2$.}
\label{fig:sp} 
\end{figure}
\vspace{-0.5cm}
The trajectory of the sequence $Z_{k}$ dependents closely on the value of $\gamma$, where $k \in \mathbb{N}$. If selecting a proper $\gamma$, the eventual trajectory of $Z_{k}$ is spiral as shown in Figure \ref{fig:sp}. since the trajectory of $Z_{k}$ has the characteristic of spiral, which provides a way that using the previous $q$ iterates predicts the future $s$ iterates. The update $Z_{k}$ of $\mathbf{ADMM}$ can be estimated by $\Bar{Z}_{k,s}$ which is defined as follow:
$$\Bar{Z_{k}}=\mathcal{F}(Z_{k},Z_{k-1},\cdots,Z_{k-q}),$$
for the choice of $s=1$. Given a sequence $Z_{k-i},i=1,2,\cdots,q+1$, we can define $v_{i}=Z_{i}-Z_{i-1}$ then obtain the past $v_{k-1},v_{k-2},\cdots,v_{k-q}$ which can be used the $v_{k-1},v_{k-2},\cdots,v_{k-q}$ to estimate $v_{k}$. Let $V_{k-1}=[v_{k-1},v_{k-2},\cdots,v_{k-q}]\in \mathbb{R}^{n\times q}$ and $C_{k}= \mathop{\arg \min}_{C \in \mathbb{R}^{q}} \|V_{k-1}C-v_{k} \|^{2}=\|\sum_{i=1}^{q}C_{i}v_{k-i}-v_{k} \|^{2}$. Then we can use the $V_{k}C_{k}$ to approximate $v_{k+1}$, that is $V_{k}C_{k} \approx v_{k+1},$ we can compute $\Bar{Z}_{k+1}=Z_{k}+V_{k}C \approx Z_{k+1}.$ By iterating $s$ times, we can obtain $\Bar{Z}_{k,s} \approx Z_{k+s}.$ This method is proposed by Clarice Poon and Jingwei Liang(2019) \cite{2019Liang}. 

Our Adaptive Composite Gradient method is motivated by two aspects. Firstly, we still consider the cyclic behavior around a Nash as a circular motion process around an origin but not a uniform circular motion process. Therefore, similar to the centripetal acceleration method, we modify the directions by adding the projection of the centripetal acceleration term at time $t$. Secondly, The $\mathbf{A^{3}DMM}$ provides an idea that can use the past iterates to predict the future iterates because the trajectory of the sequence $Z_{k}$ is either straight liner or spiral. However, the cyclic behavior around a Nash is also approximately a spiral. In our method, to reduce the computing cost and accelerate the iteration, we consider the controlled parameters by players as a spiral trajectory as shown in Figure \ref{fig:fig1}. By the two motivations, we propose a \textbf{Semi-Gradient-Free} method named the Adaptive Composite Gradient method to optimize games' problems, which also can alleviate the cyclic behaviors and be used to train GANs.

\section{Adaptive Composite Gradient Method}
In this section, we mainly introduce proposed the Adaptive Composite Gradient method. Firstly, in order to more easily understand our method, we make some symbol conventions throughout the paper. let $\vec{a} \bigodot \vec{b}$ is considered as the projection of $\vec{a}$ onto $\vec{b}$, where $\vec{a},\vec{b} \in \mathbb{R}^{n}$ and $\bigodot$ denotes the projecting calculation between two vectors. $w_{t}^{i}$ denotes the $i^{th}$ player controlling  parameter at time $t$ and $W_{t}=(w_{t}^{1},w_{t}^{2},\cdots,w_{t}^{n})$. we use $\lbrace \ell_{i}: \mathbb{R}^{d} \rightarrow \mathbb{R} \rbrace_{i=1}^{n}$ to denote the losses corresponding to $n$ players which is same as mentioned in \textbf{Definition 1.} and we can obtain $\mathscr{L}(W_{t})=(\ell_{1}(w_{t}^{1}),\ell_{2}(w_{t}^{2}),\cdots,\ell_{n}(w_{t}^{n}))$ which is the payoff vector of the $n$ players at time $t$.

We consider a bilinear game problem with the following form 
\begin{equation}
\begin{split}
    \ell_{1}(w^{1},w^{2})&=(w^{1})^{\mathrm{T}} \textup{A}w^{2} \\
    \ell_{1}(w^{1},w^{2})&=-(w^{1})^{\mathrm{T}} \textup{A}w^{2}.
\end{split}
\label{eq3}
\end{equation}
\textbf{Desiderata.} This two players game has a Nash equilibrium which must satisfy 
\begin{description}
  \item[D1.] The two losses satisfy: $\sum_{i=1}^{2}\ell_{i}(w^{1},w^{2}) \equiv 0;$
  \item[D2.] For $\ell_{i}$ is differentiable over the parameter space $\Omega(w^{1}) \times \Omega(w^{2}),$ where $\Omega(w^{1}) \times \Omega(w^{2}) \subseteq \mathbb{R}^{n} \times \mathbb{R}^{n}.$
\end{description}
The player 1 which holds the parameter $w^{1}$ tries to minimize the loss $\ell_{1}$, while the player 2 which holds the parameter $w^{2}$ tries to minimize the loss $\ell_{2}$. From the \textup{D1.} we can get $\ell_{1} = -\ell_{2}$, the equation \eqref{eq3} can be rewritten as $$\mathop{\min}\limits_{w^{1} \in \Omega(w^{1})} \mathop{\max} \limits_{w^{2} \in \Omega(w^{2})} \mathscr{L}((w^{1},w^{2})).$$ As well known, the dynamics of traditional gradient descent ascent based method reads $$w_{t+1}^{1}=w_{t}^{1}- \alpha \nabla_{w^{1}}\mathscr{L}((w_{t}^{1},w_{t}^{2})) $$
$$w_{t+1}^{2}=w_{t}^{2}+ \alpha \nabla_{w^{2}}\mathscr{L}((w_{t}^{1},w_{t}^{2})).$$
According the previous section motivational theory we exploit the spiral characteristic to design the proposed Adaptive Composite Gradients method(ACG). The ACG method involves three parts which are consist of the composite gradients. Firstly, we introduce the predictive aspect. In this section, $W_{t}=(w_{t}^{1},w_{t}^{2})$ is the parameter vectors at time $t$. Similar to $\mathbf{A^{3}DMM}$, we utilize the $W$ of the previous $q$ iterations to predict the future $s$ iterations which can be denoted by $\Bar{W}_{t,s}$. Then we can get the following formula of $\Bar{W}_{t,s}:$
$$\Bar{W}_{t}=\mathcal{F}({W}_{t},{W}_{t-1},\cdots,{W}_{t-q}),$$ for the value of $s=1$. Define $v_{i}=W_{i}-W_{i-1}$, $W_{i}$ is from sequence $\lbrace W_{k-i} \rbrace_{i=0}^{q}.$ We can get the past $v_{k-1},v_{k-2},\cdots,v_{k-q}$ which can be used to approximate the last $v_{k}$. Let define $V_{k-1}=[v_{k-1},v_{k-2},\cdots,v_{k-q}]$ and $c_{k}=\mathop{\arg\min}_{c \in \mathbb{R}^{q}}\|V_{k-1}c-v_{k}\|^2=\|\sum_{i=1}^{q}c_{i}v_{k-i}-v_{k}\|^2$. The behind of idea is $V_{k}c_{k}\approx v_{k+1}$ and $\Bar{W}_{k,1}=W_{k}+V_{k}c_{k}\approx W_{k+1}$,By looping of s times, we can get $\Bar{W}_{k,s} \approx W_{k+s}.$ The second part and third part of our ACG method are the $\nabla \mathscr{L}((w_{t}^{1},w_{t}^{2}))$ and the projection of centripetal acceleration term.

The dynamics of proposed Adaptive Composite Gradients is 
\begin{equation}\small
\begin{split}
composite \; gradients:\\
G_{w^{1}}=\nabla_{w^{1}} \mathscr{L}((w_{t}^{1},& w_{t}^{2})) + \frac{\beta_{1}}{\alpha}(\vec{a}_{1}-\vec{b}_{1}) + \frac{\beta_{2}}{\alpha}\Bar{w}_{t+s}^{1};\\
G_{w^{2}}=\nabla_{w^{2}} \mathscr{L}((w_{t}^{1},& w_{t}^{2})) + \frac{\beta_{1}}{\alpha}(\vec{a}_{2}-\vec{b}_{2})  +\frac{\beta_{2}}{\alpha}\Bar{w}_{t+s}^{2};\\
gradient \; step: w_{t+s}^{1}&=w_{t}^{1}-\alpha G_{w^{1}};\\
w_{t+s}^{2}&=w_{t}^{2}+\alpha G_{w^{2}}.
\end{split}
\end{equation}
Where the $\nabla_{w^{i}} \mathscr{L}((w_{t}^{1},w_{t}^{2}))$ denotes the partial derivatives of $w_{i}$ corresponding to $\ell_{i}$ at time $t$.The $\vec{a}_{i}$ denotes $\nabla_{w^{i}} \mathscr{L}((w_{t}^{1},w_{t}^{2}))-\nabla_{w^{i}} \mathscr{L}((w_{t-1}^{1},w_{t-1}^{2}))$. The $\vec{b}_{i}$ denotes $\vec{a}_{i} \bigodot \nabla_{w^{i}} \mathscr{L}((w_{t-1}^{1},w_{t-1}^{2}))$ which is the projection of $\vec{a}_{i}$ onto the vector $\nabla_{w^{i}} \mathscr{L}((w_{t-1}^{1},w_{t-1}^{2}))$.

\begin{algorithm}[H]\small
\caption{ACG-Adaptive Composite Gradients method for Bilinear game} 
\label{alg1}
\begin{algorithmic}
\REQUIRE losses $\mathscr{L}(\ell_{1},\ell_{2})$ and $W=(w^{1},w^{2})$
\ENSURE Let $s \geq 1, q \geq 1$ be integers and $k=q+1,$ learning rate $\alpha$, adaptive rate $\beta_{1},\beta_{2}$, $W_{0}=(w^{1}_{0},w^{2}_{0})$.
\WHILE{not converged}
\FOR{$t \geq 1$}
\IF{$\mathbf{mod}(t,k)==0$}
\STATE Compute $C_{t}$ and $\Bar{w}_{t+s}^{1},\Bar{w}_{t+s}^{2}:$
\begin{align*}
\Bar{w}_{t+s}^{1}=\mathcal{F}(w_{t}^{1},w_{t-1}^{1},\cdots,w_{t-q}^{1})&;\\
\Bar{w}_{t+s}^{2}=\mathcal{F}(w_{t}^{2},w_{t-1}^{2},\cdots,w_{t-q}^{2})&;
\end{align*}
\vspace{-0.5cm}
\STATE Composite gradients:
\begin{align*}
G_{w^{1}}=\nabla_{w^{1}} \mathscr{L}((w_{t}^{1},w_{t}^{2})) + \frac{\beta_{1}}{\alpha}(\vec{a}_{1}-\vec{b}_{1})&\\
+ \frac{\beta_{2}}{\alpha}\Bar{w}_{t+s}^{1}&;\\
G_{w^{2}}=\nabla_{w^{2}} \mathscr{L}((w_{t}^{1},w_{t}^{2})) + \frac{\beta_{1}}{\alpha}(\vec{a}_{2}-\vec{b}_{2})&\\  
+\frac{\beta_{2}}{\alpha}\Bar{w}_{t+s}^{2}&;
\end{align*}
\vspace{-0.5cm}
\STATE Gradient update:
\begin{align*}
    w_{t+s}^{1}=w_{t}^{1}-\alpha G_{w^{1}}&;\\
    w_{t+s}^{2}=w_{t}^{2}+\alpha G_{w^{2}}&.
\end{align*}
\ELSE 
\STATE OMD update:
\begin{align*}
w_{t+1}^{1}=w_{t}^{1}-\alpha \nabla_{w^{1}} \mathscr{L}((w_{t}^{1},w_{t}^{2}))&\\
+\beta \nabla_{w^{1}} \mathscr{L}((w_{t-1}^{1},w_{t-1}^{2}))&;\\
w_{t+1}^{2}=w_{t}^{2}+\alpha \nabla_{w^{2}} \mathscr{L}((w_{t}^{1},w_{t}^{2}))&\\
-\beta \nabla_{w^{2}} \mathscr{L}((w_{t-1}^{1},w_{t-1}^{2}))&.\\ 
//Replace\;this\;step\;with\;any\;optimizer
\end{align*}
\ENDIF
\ENDFOR
\ENDWHILE
\end{algorithmic}
\end{algorithm}
\vspace{-0.6cm}
In Algorithm \ref{alg1}, The Adaptive Composite Gradient is proposed for the bilinear game with two players. It is remarkable that the ACG method can calibrate any optimizer based on gradient descent ascent. While the ACG method can extend for a game with $n$ player. $g(W_{t})$ is the gradient of the losses for the controlling parameters of the corresponding players. It is worth noting that the losses are required to satisfy differentiable. We adopt the way same as Algorithm \ref{alg1} to compute $\Bar{W}_{t}$. The dynamics of the ACG method for $n$-players reads
$$\Bar{W}_{t+s}=\mathcal{F}({W}_{t},{W}_{t-1},\cdots,{W}_{t-q}),$$
\vspace{-0.5cm}
\begin{equation}\label{f5}
\begin{split}
composite \;gradients: \\
    G_{W}=g(W_{t})+&\frac{\beta_{1}}{\alpha} (\vec{a}-\vec{b})+ \frac{\beta_{2}}{\alpha} \Bar{W}_{t+s};\\
    gradient \;step: W_{t+s}=&W_{t}-\alpha G_{W}.
\end{split}
\end{equation}
Where the $\vec{a}$ denotes $g(W_{t})-g(W_{t-1})$, $\vec{b}$ denotes $\vec{a} \bigodot g(W_{t-1})$ which is the projection of $\vec{a}$ onto  $g(W_{t-1})$.
\vspace{-0.2cm}
\begin{algorithm}[H]\small
\caption{ACG - Adaptive Composite Gradients method for general game with $n$ players} 
\label{alg2}
\begin{algorithmic}
\REQUIRE losses $\mathscr{L}(W)$ and $W=(w^{1},w^{2},\cdots,w^{n})$
\ENSURE Let $s \geq 1, q \geq 1$ be integers and $k=q+1,$ learning rate $\alpha$, adaptive rate $\beta_{1},\beta_{2}$, $W_{0}=(w^{1}_{0},w^{2}_{0},\cdots,w^{n}_{0})$.
\WHILE{not converged}
\FOR{$t \geq 1$}
\IF{$\mathbf{mod}(t,k)==0$}
\STATE Compute $C_{t}$  and $\Bar{W}_{t+s}:$
$$
\Bar{W}_{t+s}=\mathcal{F}(W_{t},W_{t-1},\cdots,W_{t-q});
$$
\vspace{-0.5cm}
\STATE Compute composite gradients:
$$
    G_{W}=g(W_{t})+\frac{\beta_{1}}{\alpha} (\vec{a}-\vec{b})+ \frac{\beta_{2}}{\alpha} \Bar{W}_{t+s};
$$
\vspace{-0.5cm}
\STATE Gradient update:
$$W_{t+s}=W_{t}-\alpha G_{W}.$$
\ELSE 
\STATE Gradient update:
$$
W_{t+1}=W_{t}-\alpha g(W_{t}).
$$
\ENDIF
\ENDFOR
\ENDWHILE
\end{algorithmic}
\end{algorithm}
\vspace{-0.5cm}
\textbf{Remark 4.1.} The value of $k$ can be controlled both in Algorithm \ref{alg1} and Algorithm \ref{alg2}. Let $k=q+i$ where $i \in \mathbb{N}^{+}$, we can set different acceleration ratio of the algorithms by adjusting the values of $k,s.$

\textbf{Note:} \textbf{(1)} In Algorithm \ref{alg1} and Algorithm \ref{alg2}, the memory cost of storing $V_{k}$ is $nq$ and computational cost of obtaining the pseudo-inverse of $V_{k}$ is $nq^{2}$. \textbf{(2)} There is no need to calculate the gradient of each iteration because of the $\Bar{W}_{t+s}$. So that this is a \textbf{Semi-Gradient-Free} method which can reduce the computational cost of calculating gradients. \textbf{(3)} The $\beta_{1},\beta_{2}$ can be used to control the convergence of the algorithms.
\begin{figure}[H]
    \centering
    \includegraphics[width=0.90\linewidth]{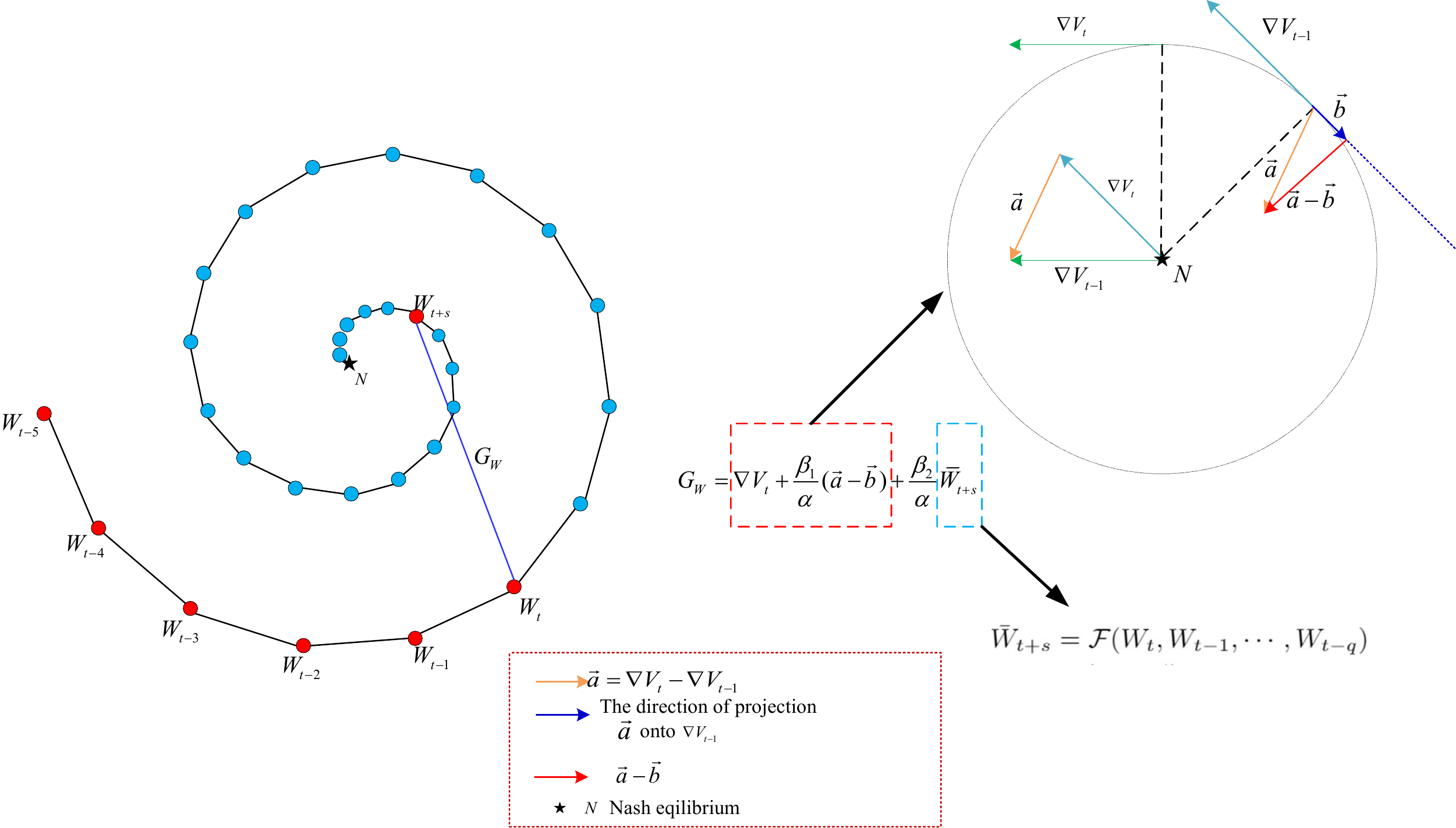}
    \captionsetup{font={small}}
    \caption{The basic intuition of our proposed Adaptive Composite Gradient method. To illustrate our approach, we chose the $s=20$ in this figure. We explored the influence of $s$ on the convergence in Figure \ref{fig8} of Appendix.}
    \label{fig:Ours}
\end{figure}
\vspace{-0.1cm}
\section{The Convergence of Adaptive Composite Gradient Method}
The proposed Adaptive Composite Gradient Method is adapted for bilinear games and $n$-players games. We discussed the convergence of the Adaptive Composite Gradient Method for the above two situations in this section.
\vspace{-0.5cm}
\subsection{The Convergence of Adaptive Composite Gradient Method for Bilinear Game}
In this subsection, we mainly discuss the convergence of Adaptive Composite Gradient Method in the bilinear game, which reads 
\begin{equation}\label{f1}
    \mathop{\min}\limits_{\theta \in \mathbb{R}^{d}} \mathop{\max} \limits_{\phi \in \mathbb{R}^{d}} \theta^{\mathrm{T}}\emph{A} \phi + \theta^{\mathrm{T}}B + C^{\mathrm{T}}\phi, \quad \emph{A} \in \mathbb{R}^{d \times d}, \quad B,C \in \mathbb{R}^{d}.
\end{equation}
For any local Nash equilibrium of the bilinear game has the point $(\theta^{*},\phi^{*})$ satisfied the following conditions:
\begin{equation}
\begin{split}
    \emph{A} \phi^{*} + B =0 \\
    \emph{A}^{\mathrm{T}} \theta^{*} + C =0.
\end{split}
\end{equation}
The local Nash equilibrium exists if and only if the ranks of $\emph{A}$ and $\emph{A}^{\mathrm{T}}$ are the same as the dimension of $B,C$. By this way, without loss of generality, we can convert $(\theta, \phi)$ to $(\theta-\theta^{*}, \phi-\phi^{*})$ which is used to rewrite bilinear game \eqref{f1} as:
\begin{equation}
    \mathop{\min}\limits_{\theta \in \mathbb{R}^{d}} \mathop{\max} \limits_{\phi \in \mathbb{R}^{d}} \theta^{\mathrm{T}}\emph{A} \phi, \quad \emph{A} \in \mathbb{R}^{d \times d}.
\end{equation}
Before analyzing the convergence property of the Adaptive Composite Gradient Method in a bilinear game, we introduce some essential theorems and propositions.
\begin{theorem}\label{t1}
       Suppose $F\in \mathbb{R}^{d\times d}$ is from the iterative system $x_{k+1}=Fx_{k}$. If $F$ is nonsingular and satisfies the spectral radius $\rho(F)<1$, then the $x_{k}$ of iterative  system converges to 0 linearly.
\end{theorem}
    
\begin{theorem}[OMD] \label{omd}
       Consider a bilinear game $V(\theta,\phi)= \theta^{\mathrm{T}} \emph{A} \phi$, where $\emph{A} \in \mathbb{R}^{d \times d} $. Assume $\emph{A}$ is full rank. Then the following dynamics,  
\begin{equation}
\begin{split}
    \theta_{t+1} &= \theta_{t} - 2\eta \nabla_{\theta} V(\theta_{t}, \phi_{t}) + \eta \nabla_{\theta} V(\theta_{t-1}, \phi_{t-1})\\
    \phi_{t+1} &= \phi_{t} + 2\eta \nabla_{\phi} V(\theta_{t}, \phi_{t}) - \eta \nabla_{\phi} V(\theta_{t-1}, \phi_{t-1}),
\end{split}
\end{equation}

with the learning rate $$\eta = \frac{1}{2\sqrt{2 \lambda_{max}(\emph{A}\emph{A}^{\mathrm{T}})}},$$
obtain an $\epsilon$-minimizer such that $(\theta_{T}, \phi_{T}) \in B_{2}(\epsilon),$ provided
$$T\geqslant T_{OMD}:= \lceil 16\frac{\lambda_{max}(\emph{A}\emph{A}^{\mathrm{T}})}{\lambda_{min}(\emph{A}\emph{A}^{\mathrm{T}})} \log \frac{4 \sqrt{2} \delta}{\epsilon}\rceil $$
under the assumption that $||(\theta_{0},\phi_{0})||, ||(\theta_{1},\phi_{1})|| \leqslant \delta$.
\end{theorem}

To discuss the convergence of the ACG method for the bilinear game, we divided the analysis process into three parts(two cases). Without loss of generality, let $t$ represent the iterative step, and $k$ is the previous steps. The convergence property of Algorithm \ref{alg1} is as the following.

\textbf{Case 1.} $\mathbf{mod}(t,k) \neq 0$ or $\mathbf{mod}(t,k) =0 \& \rho(C_{k}) \geq 1$

When the iterative $t$ and the previous step $k$ satisfied the conditions of \textbf{Case 1.}, our ACG method adopts the dynamics:
\begin{equation}
\begin{split}
w_{t+1}^{1}&=w_{t}^{1}-\alpha \nabla_{w^{1}} \mathscr{L}((w_{t}^{1},w_{t}^{2}))+\beta \nabla_{w^{1}} \mathscr{L}((w_{t-1}^{1},w_{t-1}^{2}));\\
w_{t+1}^{2}&=w_{t}^{2}+\alpha \nabla_{w^{2}} \mathscr{L}((w_{t}^{1},w_{t}^{2}))-\beta \nabla_{w^{2}} \mathscr{L}((w_{t-1}^{1},w_{t-1}^{2})). 
\end{split}
\end{equation} 
Taking $\alpha = 2 \beta = 2 \eta$ in \textbf{Case 1}, the dynamics scheme reduces to \textbf{OMD}, which can be written as:
\begin{equation}
\begin{split}
w_{t+1}^{1}&=w_{t}^{1}-2\eta \nabla_{w^{1}} \mathscr{L}((w_{t}^{1},w_{t}^{2}))+\eta \nabla_{w^{1}} \mathscr{L}((w_{t-1}^{1},w_{t-1}^{2}));\\
w_{t+1}^{2}&=w_{t}^{2}+2\eta \nabla_{w^{2}} \mathscr{L}((w_{t}^{1},w_{t}^{2}))-\eta \nabla_{w^{2}} \mathscr{L}((w_{t-1}^{1},w_{t-1}^{2})). 
\end{split}
\end{equation} 
\textbf{Theorem} \ref{omd} has assigned the condition about learning rate of \textbf{OMD} and it is exponential Convergence. The convergence of \textbf{OMD} can be found in \cite{Sim2018}[Theorem 3].
 
\textbf{Case 2.} $\mathbf{mod}(t,k) = 0 \  \& \ \rho(C_{k}) < 1$ 

From the Algorithm \ref{alg1}, firstly we shall compute the $\Bar{w}_{t+s}$ by:
\begin{equation}\label{f12}
\begin{split}
      \Bar{w}_{t+s}^{1}&=\mathcal{F}(w_{t}^{1},w_{t-1}^{1},\cdots,w_{t-q}^{1})\\
      \Bar{w}_{t+s}^{2}&=\mathcal{F}(w_{t}^{2},w_{t-1}^{2},\cdots,w_{t-q}^{2}).
\end{split}
\end{equation}
Using the fixed-point formulation of ADMM, (\ref{f12}) can be written as an unified $\Bar{w}_{t} = \varepsilon(\Bar{w}_{t-1})$, let $V_{t}c_{t} = \sigma_{t}$, then $\Bar{w}_{t} = \varepsilon(\Bar{w}_{t-1} + \sigma_{t})$. We can obtain the convergence of (\ref{f12}), iff $\sigma_{t}$ converges to $0$. The convergence of  formula (\ref{f12}) is based on the convergence of inexact Krasnosel'skil-Mann fixed-poin iteration (\cite{book1},Proposition 5.34). The detail convergence analysis of $\Bar{w}_{t} = \varepsilon(\Bar{w}_{t-1} + \sigma_{t})$ has been discussed in \cite{2019Liang}(Proposition 4.2).

Then we discuss the convergence of composite gradients update scheme, which is written as follow:
\begin{equation}
\begin{split}
    w_{t+s}^{1}=w_{t}^{1}-\alpha G_{w^{1}};\\
    w_{t+s}^{2}=w_{t}^{2}+\alpha G_{w^{2}},
\end{split}
\end{equation}
where the $G_{w^{1}},G_{w^{2}}$ are defined as:
\begin{equation}
\begin{split}
   G_{w^{1}}=\nabla_{w^{1}} \mathscr{L}((w_{t}^{1},& w_{t}^{2})) + \frac{\beta_{1}}{\alpha}(\vec{a}_{1}-\vec{b}_{1}) + \frac{\beta_{2}}{\alpha}\Bar{w}_{t+s}^{1};\\
   G_{w^{2}}=\nabla_{w^{2}} \mathscr{L}((w_{t}^{1},& w_{t}^{2})) + \frac{\beta_{1}}{\alpha}(\vec{a}_{2}-\vec{b}_{2})  +\frac{\beta_{2}}{\alpha}\Bar{w}_{t+s}^{2};\\
\end{split}
\end{equation}
\vspace{-0.5cm}
\begin{proposition}\label{p1}
Any given two vector $\vec{a},\vec{b}$, the projection of vector $\vec{b}$ onto the vector $\vec{a}$ can be denoted as $\gamma \vec{a},$ where the $\gamma \in \mathbb{R}.$
\end{proposition}
\end{multicols}
\vspace{-0.3cm}
According to proposition \ref{p1}, our dynamics of the bilinear game, the composite gradients update scheme is reduced to be
\begin{equation}\label{f13}
\begin{split}
    \theta_{t+1}&=\theta_{t}- (\alpha+\beta_{1}) \emph{A} \phi_{t} + \beta_{1}(1+\gamma)\emph{A}\phi_{t-1} - \beta_{2} \Bar{w}_{\theta_{t}};\\
    \phi_{t+1}&=\phi_{t}+(\alpha +\beta_{1})\emph{A}^{\mathrm{T}} \theta_{t} - \beta_{1}(1+\gamma) \emph{A}^{\mathrm{T}} \theta_{t-1} + \beta_{2} \Bar{w}_{\phi_{t}}.
\end{split}
\end{equation}
We can obtain the iterative matrix  as:
\begin{equation}\label{matrix}
    F:= \begin{bmatrix} I_{d} & -(\alpha+\beta_{1})\emph{A} & 0 & \beta_{1}(1+\gamma)\emph{A} & -\beta_{2}I_{d} & 0 \\ 
         (\alpha+\beta_{1})\emph{A}^{\mathrm{T}} & I_{d} & -\beta_{1}(1+\gamma)\emph{A}^{\mathrm{T}} & 0 & 0 & \beta_{2}I_{d} \\ 
         I_{d} & 0 & 0 & 0 & 0 & 0 \\ 
         0 & I_{d} & 0 & 0 & 0 & 0 \\ 
         0 & 0 & 0 & 0 & \tau I_{d}  & 0 \\ 
         0 & 0 & 0 & 0 & 0 & \tau I_{d} \\ 
    \end{bmatrix},
\end{equation}
where $\tau$ is defined by (\ref{f12}).

According to the iterative matrix, it is easy to obtain that 
$[\theta_{t+1},\phi_{t+1},\theta_{t},\phi_{t},\Bar{w}_{\theta_{t+1}},\Bar{w}_{\phi_{t+1}}]^{\mathrm{T}}=F[\theta_{t},\phi_{t},\theta_{t-1},\phi_{t-1},\Bar{w}_{\theta_{t}},\Bar{w}_{\phi_{t}}]^{\mathrm{T}},$ where the $(\theta_{t},\phi_{t})$ are generated by (\ref{f13}). With the assumption that $\emph{A}$ is square and nonsingular in Proposition \ref{p2}, we use the well-known theorem \ref{t1} to illustrate the linear convergence for the update scheme (\ref{f13}).

\begin{proposition}\label{p2}
Suppose that \emph{A} is square and nonsingular. Then, the eigenvalues of $F$ are the roots of the sixth order polynomials:
\begin{equation}
    (\tau-\lambda)^{2}[\lambda^{2}(1-\lambda)^{2}+(\lambda(\alpha+\beta_{1})-\beta_{1}(1+\gamma))^{2}\xi^{2}], \quad \xi^{2} \in Sp(\emph{A}^{\mathrm{T}}\emph{A}),
\end{equation}

 where $Sp(\cdot)$ denotes the collection of all eigenvalues.
\end{proposition}
\begin{proposition}\label{p3}
Suppose that \emph{A} is square and nonsingular. The $\bigtriangleup_{t}:=||\theta_{t}||^2+||\phi_{t}||^2+||\theta_{t+1}||^2+||\phi_{t+1}||^2+||\Bar{w}_{\theta_{t+1}}||^2+||\Bar{w}_{\phi_{t+1}}||^2$ is linearly convergent to 0 for given $\gamma$ with $\alpha$ and $\beta_{1}$ satisfy 
\begin{equation}
    0< \alpha+\beta_{1} \leq \frac{1}{\sqrt{\lambda_{max}(\emph{A}^{\mathrm{T}}\emph{A})}}, \quad |\alpha+\beta_{1}| + |2\beta_{1}(1+\gamma)| \leq \frac{(\alpha+\beta_{1})^{2}\sqrt{\lambda_{min}(\emph{A}^{\mathrm{T}}\emph{A})}}{10}
\end{equation}
where the $\lambda_{max}(\cdot),\  \lambda_{min}(\cdot)$ denote the largest and the smallest eigenvalues of $\emph{A}^{\mathrm{T}}\emph{A}$.
\end{proposition}
\begin{multicols}{2}
\subsection{The Convergence of Composite Gradient Method for General N-player Game}
This subsection mainly discusses the convergence of the Adaptive Composite Gradient method in the general $n$-player game. The problem is described as in Definition \ref{D1}, according to Algorithm \ref{alg2}, the convergence analysis process in general $n$ players game is the same as that of the bilinear game with three parts and two cases. Before analyzing convergence property, we introduce some basic definitions.

\begin{definition}\label{D2}
 Suppose that $f: \mathbb{R}^{n} \rightarrow \mathbb{R}$ and it is convex, continuously. For $\forall x,y \in \mathbb{R}^{n}$, the gradient of $f$ is Lipschitz continuous with constant $L$ such that:
 \begin{equation*}
     0 \leq f(y)-f(x)- \langle \nabla f(x), y-x \rangle  \leq \frac{L}{2}||x-y||^{2},
 \end{equation*}
 we define that $f$ belongs to the class $\mathscr{F}_{L}^{1,1}.$ If $f$ is strongly convex with modulus $\mu >0$ and such that:
  \begin{equation*}
     \frac{\mu}{2}||x-y||^{2} \leq f(y)-f(x)- \langle \nabla f(x), y-x \rangle,
 \end{equation*}
 we define that $f$ belongs to $\mathscr{F}_{\mu,L}^{1,1}.$
\end{definition}

Next, we suppose that the all $\ell_{i},i=1,2,\cdots,n$ are belonging to $\mathscr{F}_{L}^{1,1}.$ We can get the definition of fixed point which is also called Local Nash Equilibrium in game.
\begin{definition}\label{D3}
 $W^{*}$ is a Local Nash Equilibrium(fixed point) if  $W^{*}$ satisfy $g(W^{*})=0$. We say that it is stable if $g(W^{*}) \geq 0$, unstable if  $g(W^{*}) \leq 0.$
\end{definition}

\begin{theorem}\label{T1}[Nesterov 1983] Let $f$ be a convex and $\beta$-smooth function, and we can write the well-known Nesterov’s Accelerated Gradient Descent update scheme as
\begin{equation}\label{f15}
\begin{split}
    y_{t+1} &= x_{t} -\alpha \nabla f(x_{t})\\
    x_{t+1} &= y_{t+1}+\beta(y_{t+1}-y_{t}).
\end{split}
\end{equation}
Then Nesterov’s Accelerated Gradient Descent satisfies 
\begin{equation}
    f(x_{t})-f(x^{*}) \leq \frac{2\beta||x_{1}-x^{*}||^{2}}{t^{2}}.
\end{equation}
Nesterov (1983) proposed the accelerated gradient method which achieves the optimal $\mathscr{O}(\frac{1}{t^{2}})$ convergence rate.
\end{theorem}
The convergence of the ACG method for the general $n$ player game is also divided into two cases. Let $t$ represent the iterative step, and $k$ is the previous steps. The convergence property of Algorithm \ref{alg2} is as the following.

\textbf{Case 1.} $\mathbf{mod}(t,k) \neq 0$ or $\mathbf{mod}(t,k) =0 \& \rho(C_{k}) \geq 1$

From Algorithm \ref{alg2}, if $t$ and $k$ satisfy the conditions of \textbf{Case 1.} Our proposed ACG method is the same as classical Gradient Descent and the update scheme is 
\begin{equation*}
    W_{t+1}=W_{t}-\alpha g(W_{t}),
\end{equation*}
where the $\alpha$ is a positive step-size parameter. According to Definition \ref{D3}, let the Local Nash Equilibrium $W^{*}$ and $\mathscr{L}^{*}=\mathscr{L}(W^{*})$. Base the Definition \ref{D2}, if $\ell_{i},i=1,2,\cdots,n$ are belonging to $\mathscr{F}_{L}^{1,1},$ then $\mathscr{L}(W_{t})-\mathscr{L}^{*}$ associated $\{W_{t}\}$ converges at rate $\mathscr{O}(\frac{1}{t}).$ For more detail convergence of averaged iterates with generalized gradient descent update scheme in convex-concave games is analyzed in \cite{BUCK1977,NAO2009}.

\textbf{Case 2.} $\mathbf{mod}(t,k) = 0 \  \& \ \rho(C_{k}) < 1$

 According to Algorithm \ref{alg2}, firstly we should compute the  $\Bar{W}_{t+s}$ by: 
\begin{equation}\label{f14}
    \Bar{W}_{t+s}=\mathcal{F}(W_{t},W_{t-1},\cdots,W_{t-q}).
\end{equation}
The convergence of formula (\ref{f14}) is the same as that of \textbf{Case 2.} in previous section 5.1. For more detail about convergence of (\ref{f14}) has been discussed in  \cite{2019Liang}(Proposition 4.2).

In \textbf{Case 2.}, we mainly analyze the convergence of composite gradient update scheme in formula (\ref{f5}). Before illustrating our proposed method convergence property, The formula (\ref{f15}) can be equivalently written as 
\begin{equation}\label{f16}
    x_{t+1}=x_{t}-(\alpha+\alpha \beta)\nabla f(x_{t})+\alpha \beta \nabla f(x_{t-1})+\beta (x_{t}-x_{t-1}),
\end{equation}
where the $\alpha$ and $\beta$ are step-size parameters. However, our proposed composite gradient method (\ref{f5}) can be transfer to the similar formula as (\ref{f16}) which is based on the Proposition \ref{p1}. That is 
\begin{equation}\label{f17}
    W_{t+1}=W_{t+1}-(\alpha+\beta_{1})g(W_{t})+\beta_{1}(1+\gamma)g(W_{t-1})-\beta_{2}\Bar{W}_{t+s},
\end{equation}
where the $\Bar{W}_{t+s}$ is equivalent to  $(W_{t}-W_{t-1})$. Comparing (\ref{f16}) with (\ref{f17}) if the parameters are equivalently transformed, our proposed adaptive composite gradient method can be reduced to the Nesterov’s Accelerated Gradient (NAG) method by this way. In 1983, Nesterov had given the convergence rate at $\mathscr{O}(\frac{1}{t^{2}})$ for convex smooth optimization in \cite{Nesterov1983}. And also there had given the convergence bounds for convex,non-convex and smooth optimization in \cite{2016unified} (Theorem 1 $\sim$ Theorem 4). 

\textbf{Remark 4.1.} In \textbf{Case 2.}, our proposed Adaptive composite gradient method has the same convergence rate and the same convergence bounds as that of NAG method because of the assumption that  all $\ell_{i},i=1,2,\cdots,n$ are belonging to $\mathscr{F}_{L}^{1,1}.$ So we can naturally obtain the convergence rate of the ACG method, which is at $\mathscr{O}(\frac{1}{t^{2}})$ based on Theorem \ref{T1}. Here we will not repeat the similar description of the convergence of our algorithm, which is the same as that of the NAG method. 
\section{Experiments}
This article deploys three parts numerical simulation with the toy functions simulation, the mixture of Gaussians, and the four Prevalent Datasets. Meanwhile, we give more details on each experimental setup. Finally, we provide the detailed experimental environment for the last two parts of the experiments.
\subsection{Toy Functions Simulation}
\begin{figure}[ht]
    \centering
    \subfigure[$g1=\theta \cdot \phi$]{
    \begin{minipage}[c]{0.3\textwidth}
    \includegraphics[width=1\textwidth]{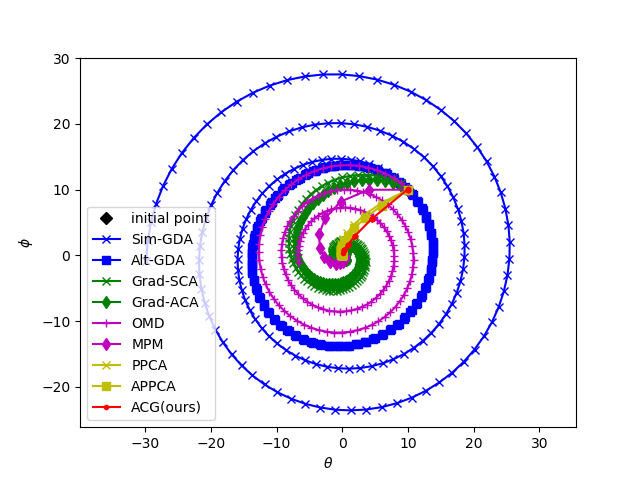}
    \end{minipage}}
    \subfigure[$g2=3\theta^{2}+\phi^{2}+4\theta \cdot \phi$]{
    \begin{minipage}[c]{0.3\textwidth}
    \includegraphics[width=1\textwidth]{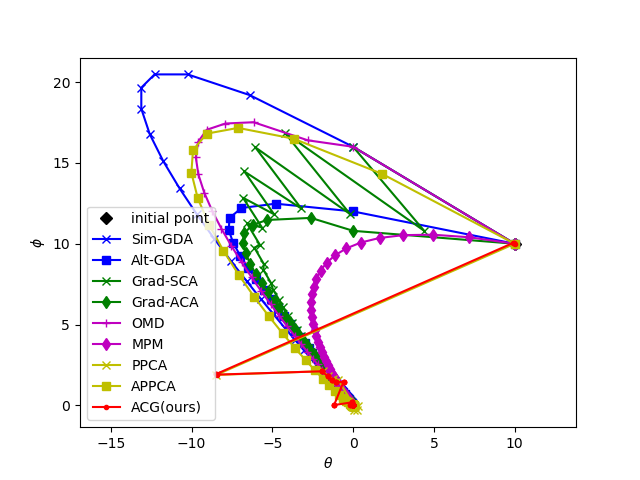}
    \end{minipage}}
     \subfigure[$g3=\theta^{2}+\phi^{2}-4\theta \cdot \phi$]{
    \begin{minipage}[c]{0.3\textwidth}
    \includegraphics[width=1\textwidth]{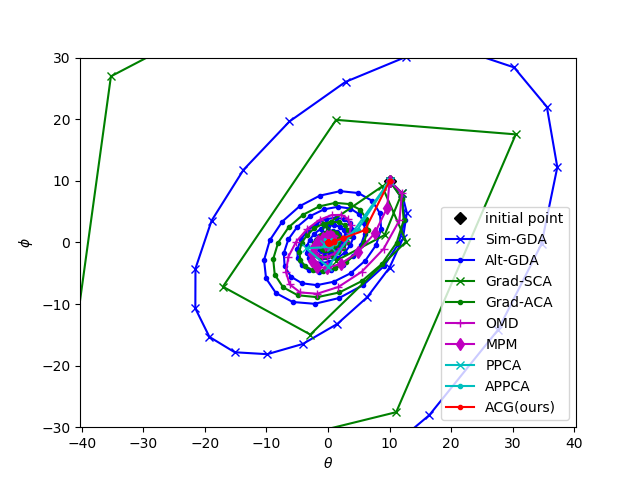}
    \end{minipage}}
    \captionsetup{font={small}}
    \caption{The effects of various compared methods in tow player games with 150 iterations. The parameters of compared methods are different in the three toy functions. In $g1$ function,the parameters are as following: Sim-GDA($\alpha=0.1$),Alt-GDA($\alpha=0.1$),Grad-SCA($\alpha=0.1,\beta=0.3$),Grad-ACA($\alpha=0.1,\beta=0.3$),OMD($\alpha=0.1,\beta=0.1$),MPM($\alpha=0.3,\gamma=1.0$),PPCA($\alpha=0.1,\beta=0.3,\gamma=1.0$),APPCA($\alpha=0.1,\beta=0.3,\gamma=1.0$),ACG($\alpha=0.05,\beta_{1}=0.5,\beta_{2}=1.0$). In $g2$ function,the parameters are as following: Sim-GDA($\alpha=0.1$),Alt-GDA($\alpha=0.1$),Grad-SCA($\alpha=0.1,\beta=0.3$),Grad-ACA($\alpha=0.1,\beta=0.03$),OMD($\alpha=0.1,\beta=0.1$),MPM($\alpha=0.3,\gamma=0.2$),PPCA($\alpha=0.1,\beta=0.3,\gamma=1.0$),APPCA($\alpha=0.1,\beta=0.02,\gamma=0.25$),ACG($\alpha=0.1,\beta_{1}=0.3,\beta_{2}=1.0$).In $g3$ function,the parameters are as following: Sim-GDA($\alpha=0.1$),Alt-GDA($\alpha=0.1$),Grad-SCA($\alpha=0.1,\beta=0.3$),Grad-ACA($\alpha=0.1,\beta=0.0.01$),OMD($\alpha=0.1,\beta=0.1$),MPM($\alpha=0.1,\gamma=0.2$),PPCA($\alpha=0.1,\beta=0.1,\gamma=1.0$),APPCA($\alpha=0.1,\beta=0.3,\gamma=0.2$),ACG($\alpha=0.1,\beta_{1}=0.05,\beta_{2}=1.0$).}
    \label{fig2}
\end{figure}

In this section, we mainly describe our experiment on toy functions. We tested our ACG methods in Algorithm \ref{alg1} and Algorithm \ref{alg2} corresponding with the bilinear game and general game with 3 players. we tested the ACG method in Algorithm \ref{alg1} on the following bilinear game, which can be written as 
\begin{equation*}
    \mathop{\min}\limits_{\theta \in \mathbb{R}^{d}} \mathop{\max} \limits_{\phi \in \mathbb{R}^{d}} \theta \cdot \phi, \quad d=1.
\end{equation*}
It is obvious that the Nash Equilibrium (stationary point) is $(0,0)$. We compared our ACG with some other methods whose results are presented in Figure \ref{fig2} (a). From the behaviors of Figure \ref{fig2}, the Sim-GDA method diverges, and the Alt-GDA method is rotating around the stationary point. However, the other methods all converge to the Nash Equilibrium. We proposed the ACG method converges faster than other convergence methods.

In Figure \ref{fig2} (b), we test our ACG method on the following general zero-sum game 
\begin{equation*}
    \mathop{\min}\limits_{\theta \in \mathbb{R}^{d}} \mathop{\max} \limits_{\phi \in \mathbb{R}^{d}} 3\theta^{2} + \phi^{2} + 4\theta \cdot \phi, \quad d=1.
\end{equation*}
The effects of all the compared methods on this game show that all methods converge to the origin. Significantly, the cyclic behavior of the Alt-GDA method has disappeared, and the Sim-GDA method converges. It is worth noting that the trajectory of our ACG method is the same as that of PPCA \cite{keke2020}. Both our ACG and PPCA \cite{keke2020} seem faster than others. We also compared ACG with other methods on the following general game
\begin{equation}\label{g2}
    \mathop{\min}\limits_{\theta \in \mathbb{R}^{d}} \mathop{\max} \limits_{\phi \in \mathbb{R}^{d}} \theta^{2} + \phi^{2} - 4\theta \cdot \phi, \quad d=1.
\end{equation}
Its effects are presented in Figure \ref{fig2} (c), which shows that Sim-GDA and Grad-SCA diverge, the rest methods converge. The APPCA \cite{keke2020} is faster than our ACG method in this game.

\begin{figure}[ht]
    \centering
    \subfigure[$\alpha=0.01,\beta_{1}=0.02, \beta_{2}=1$]{
    \begin{minipage}[c]{0.3\textwidth}
    \includegraphics[width=1\textwidth]{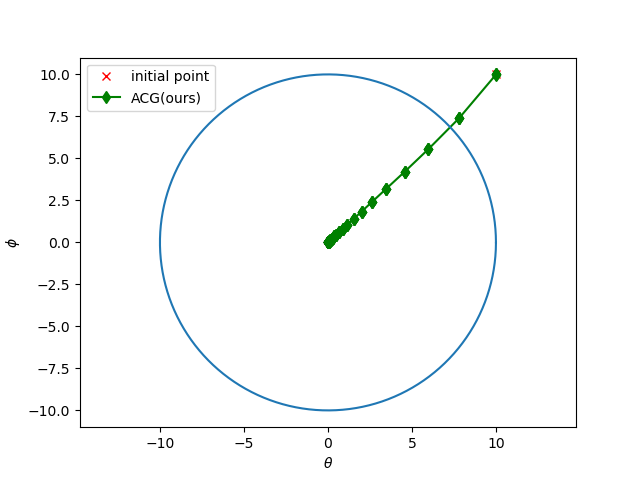}
    \end{minipage}}
    \subfigure[$\alpha=0.05,\beta_{1}=0.02, \beta_{2}=1$]{
    \begin{minipage}[c]{0.3\textwidth}
    \includegraphics[width=1\textwidth]{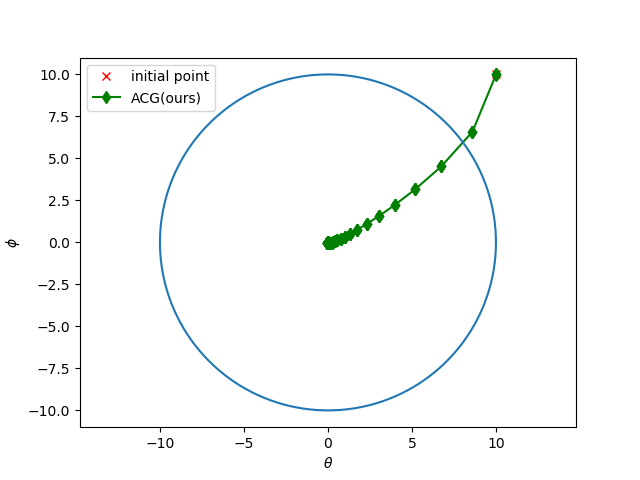}
    \end{minipage}}
     \subfigure[$\alpha=0.1,\beta_{1}=0.02, \beta_{2}=1$]{
    \begin{minipage}[c]{0.3\textwidth}
    \includegraphics[width=1\textwidth]{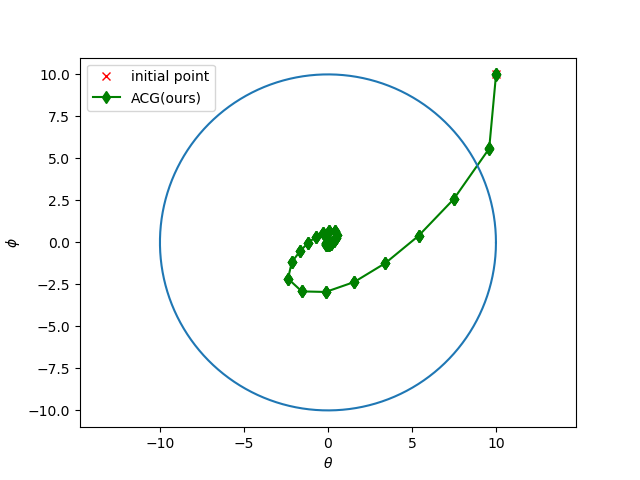}
    \end{minipage}}
    \captionsetup{font={small}}
    \caption{The robustness of the Adaptive Composite Gradient method on $g3$ toy function.}
    \label{fig3}
\end{figure}

We used the last general zero-sum game (\ref{g2}) to test the robustness of the proposed ACG method in Figure \ref{fig3}. As the learning rate $\alpha$ increases through $\{0.01,0.05,0.1\}$ and other parameters keeping same. ACG method converges faster when the learning rates setting with $\alpha=0.01$ and $\alpha=0.05$, although converge slower with learning rate $\alpha=0.1$, ACG method still converge to the origin rapidly.

\begin{figure}[ht]
    \centering
    \subfigure{
    \begin{minipage}[c]{0.9\textwidth}
    \includegraphics[width=0.3\textwidth]{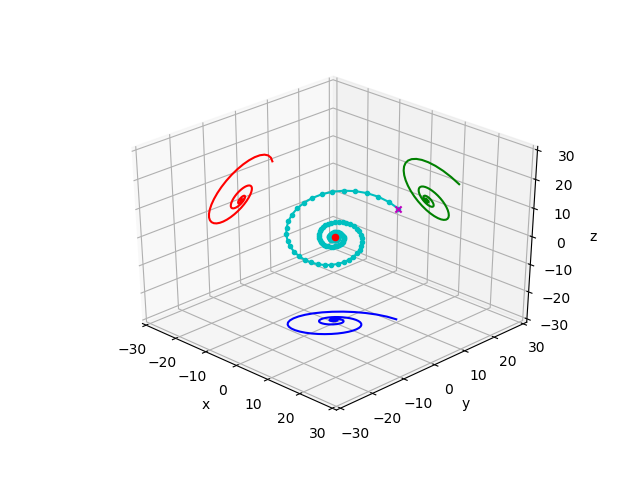}\quad
    \includegraphics[width=0.3\textwidth]{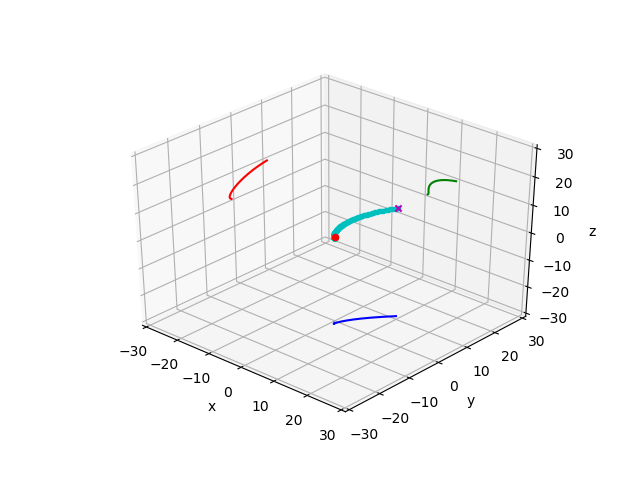}\quad
    \includegraphics[width=0.3\textwidth]{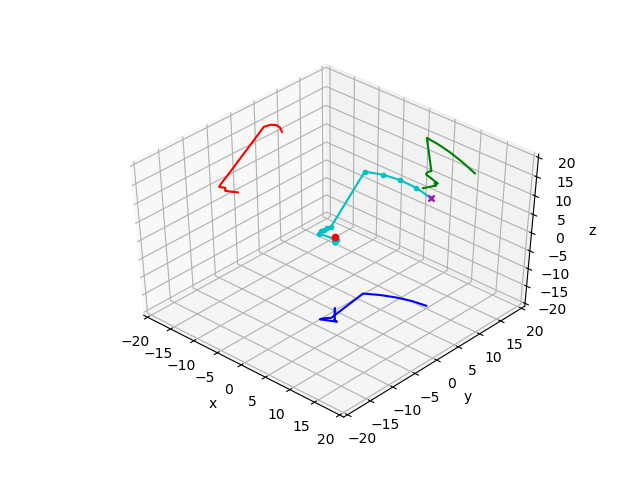}
    \end{minipage}}\\
    \subfigure{
    \begin{minipage}[c]{0.25\textwidth}
    \includegraphics[width=1\textwidth]{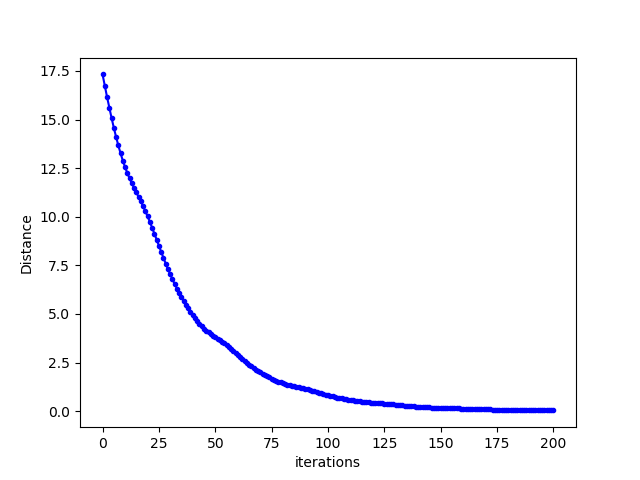}
    \caption*{SGD}
    \end{minipage}}\quad
    \subfigure{
    \begin{minipage}[c]{0.25\textwidth}
    \includegraphics[width=1\textwidth]{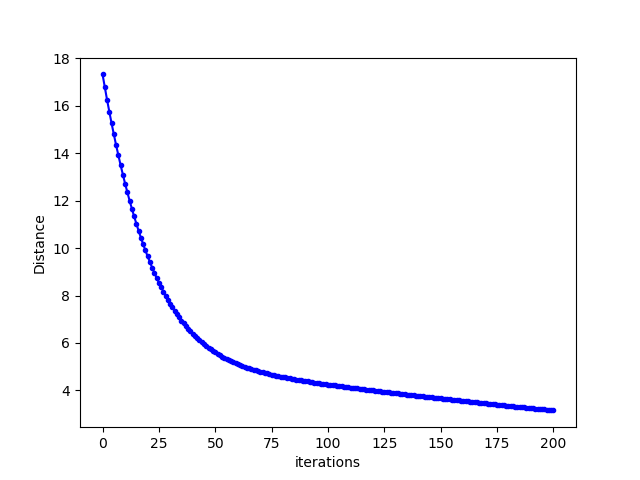}
    \caption*{SGA}
    \end{minipage}}\quad
    \subfigure{
    \begin{minipage}[c]{0.25\textwidth}
    \includegraphics[width=1\textwidth]{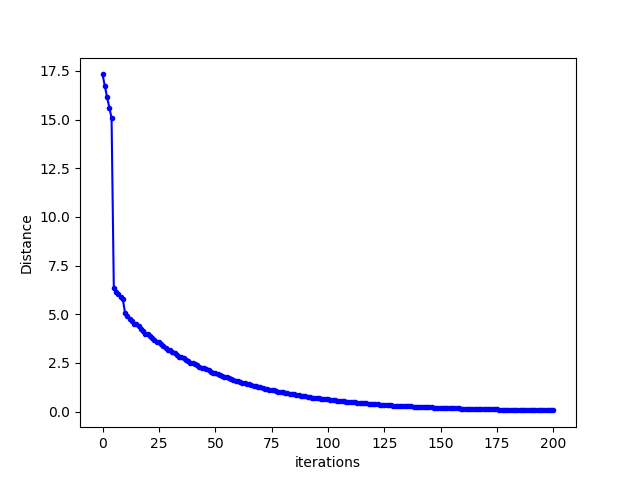}
    \caption*{ACG(Ours)}
    \end{minipage}}
    \captionsetup{font={small}}
    \caption{The effects of SGD, SGA, proposed ACG method in general 3 players' game.}
    \label{fig4}
\end{figure}

The proposed Adaptive Composite Gradient method(ACG) is also suitable for the general game with $n$ players. However, it is challenging to present the effect of the general $n$ player game by toy function. To illustrate our proposed method of Algorithm \ref{alg2} adaptive for $n$ players game, we show the effects of  Algorithm \ref{alg2} by a general 3 players game. The payoff functions can be written as 
\begin{equation*}
\begin{split}
   \ell_{1}(x,y,z)&=\frac{1}{4}x^{2}+xy+xz\\
   \ell_{2}(x,y,z)&=-xy+\frac{1}{10}y^{2}+yz\\
   \ell_{1}(x,y,z)&=-xz-yz+\frac{1}{10}z^{2}.
\end{split}
\end{equation*}
Where the local Nash Equilibrium is $(0,0,0)$. The effects are shown in Figure \ref{fig4}, the top row of Figure \ref{fig4} are the trajectory of the compared methods, and the second row of Figure \ref{fig4} are the Euclidean distances of each iteration away from the origin for compared methods. Figure \ref{fig4} presented that SGD, SGA, and our ACG method all converge to origin. There is cyclic behavior in SGD, which leads to converging slowly. The second row of Figure \ref{fig4} presents that the proposed ACG method approaches the origin faster than SGA and SGD.
\subsection{Mixtures of Gaussians}
In this section, we concentrate on the mixture of Gaussians experiments. GANs are the typical example of two players game in deep learning. We tested the proposed ACG method by training a toy GAN model, and we compared our method with other well-known methods on learning 5 Gaussians and 16 Gaussians. All the mixture of 16 Gaussians and 5 Gaussians are appointed with a standard deviation of 0.02. The Ground truths for the mixture of 16 Gaussians and 5 Gaussians are present in  Appendix Figure \ref{fig5}.

\textbf{Details on Network architecture.} GANs consist of a generator network and a discriminator network. We set up both the generator and discriminator networks with six fully connected layers, and each layer has 256 neurons. We used the fully connected layer to replace the sigmoid function layer, appended to the discriminator. We adopt the ReLU function layer appended to each of the six layers in the generator and discriminator networks. The generator network has two output neurons, while the discriminator network has one output. The input data of the generator is a random noise sampled from a standard Gaussian distribution. The output of the generator is used as the input for the discriminator. The output of the discriminator is used to evaluate the quality of the generated points by the generator.

\textbf{Experimental environments.} We deploy the mixture Gaussians experiments on the computer with \textbf{CPU AMD Ryzen 7 3700}, \textbf{GPU RTX 2060}, \textbf{6GB RAM}, Python(version 3.6.7), Keras(version 2.3.1), TensorFlow(version 1.13.1), PyTorch(version 1.3.1). We conducted each of the compared methods with 10,000 iterations.

We conducted the experiments on the mixture of 5 Gaussians with the proposed ACG method and several other methods as shown in Appendix Figure \ref{fig6}. The training set of the all compared algorithms are as follows: 
\vspace{-0.2cm}
\begin{itemize}[leftmargin=*]
 \item RMSP: We employ the TensorFlow to provide Simultaneous RMSPropOptimizer and set the learning rate $\alpha=5 \times 10^{-4}.$
  \item RMSP-alt: The alternating RMSPropOptimizer realized by TensorFlow  with learning rate $\alpha=5 \times 10^{-4}.$ 
  \item ConOpt\cite{mescheder2018}: The Consensus Optimizer realized by TensorFlow with $h=1 \times 10^{-4},\gamma =1$.
  \item RMSP-SGA\cite{pmlr-v80}: The Symplectic Gradient Adjusted RMSPropOptimizer realized by TensorFlow with leanring rate $\alpha=1 \times 10^{-4},\xi =1$.
  \item RMSP-ACA\cite{Peng_2020}: The Alternating Centripetal Acceleration on RMSPropOptimizer realized by TensorFlow with learning rate  $\alpha=5 \times 10^{-4}, \beta =0.5$.
  \item SGA-ACG(ours): Our proposed ACG method in Algorithm \ref{alg1} on the SGA Optimizer relaized by PyTorch with learning rate $\alpha=5 \times 10^{-4},\beta_{1}=5 \times 10^{-7},\beta_{2}=\alpha$.
\end{itemize}
\end{multicols}
\vspace{-0.7cm}
\vspace{-0.5cm}

\begin{multicols}{2}
Figure \ref{fig6} shows that the RMSP, RMSP-alt, RMSP-ACA do not converge after 10,000 iterations. In contrast, the ConOpt, RMSP-SGA, SGA-ACG algorithms all converge, and the generated mixture of 5 Gaussians is almost approaching the ground truth in Figure \ref{fig5}. It seems that they have the same convergence speed among  ConOpt, RMSP-SGA, SGA-ACG(Ours). To compare the convergence speed among all six algorithms, we employ the same training settings as the mixture of 5 Gaussians to conduct the mixture of 16 Gaussians as shown in Figure \ref{fig7}.

\begin{figure}[H]
    \centering
    \includegraphics[width=0.85\linewidth]{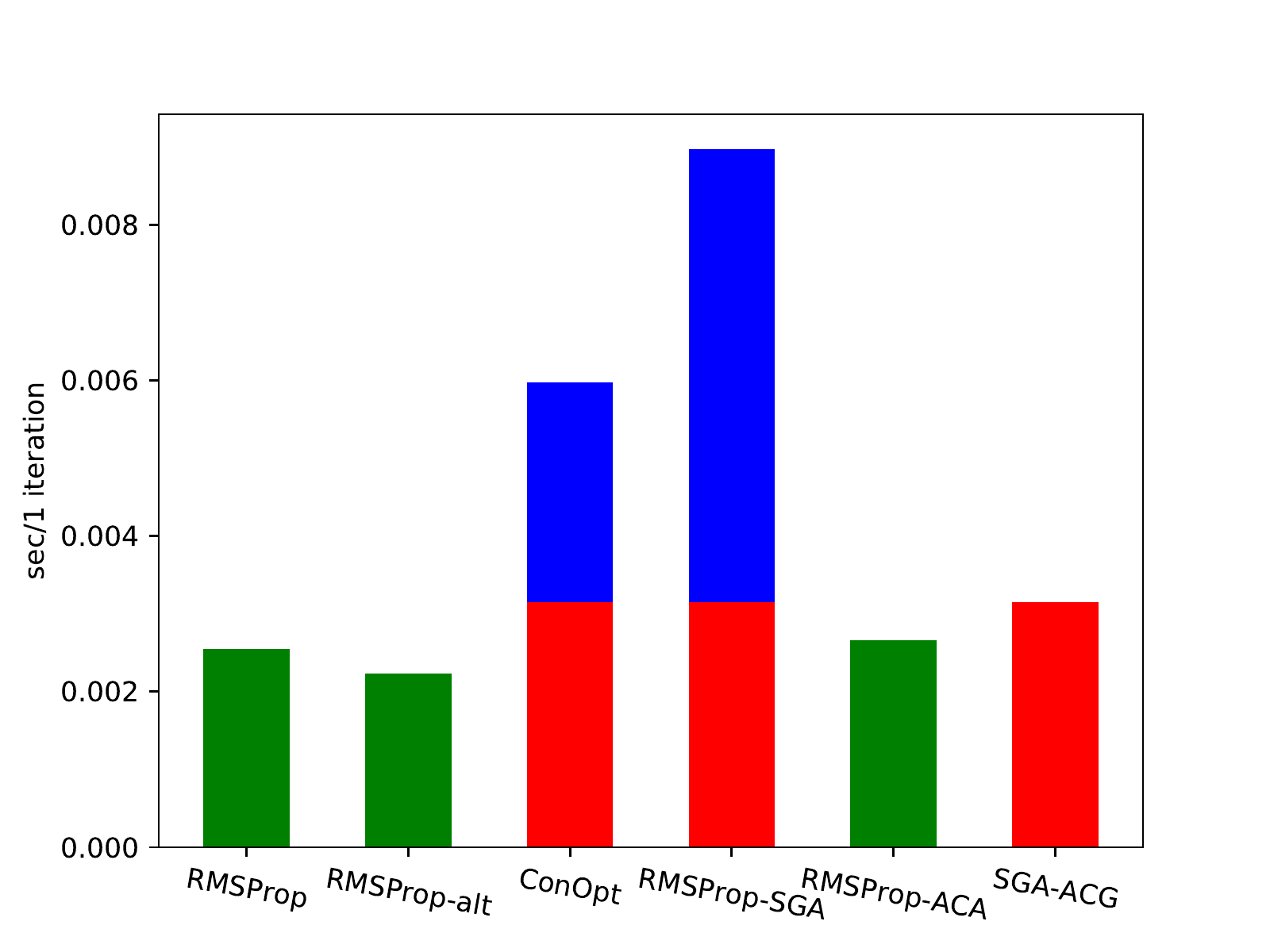}
    \captionsetup{font={small}}
    \caption{The time consuming of compared methods on the mixture of 16 Gaussians in Appendix Figure \ref{fig7-1}. Our proposed method SGA-ACG takes more time than RMSProp, RMSProp-alt, and RMSProp-ACA. However, it takes less time than ConOpt and RMSProp-SGA methods.}
    \label{fig:time}
\end{figure}

\begin{figure}[H]\tiny
(ConOpt)\qquad \quad \;
\subfigure{
\begin{minipage}[c]{0.95\textwidth}
\includegraphics[width=0.19\textwidth]{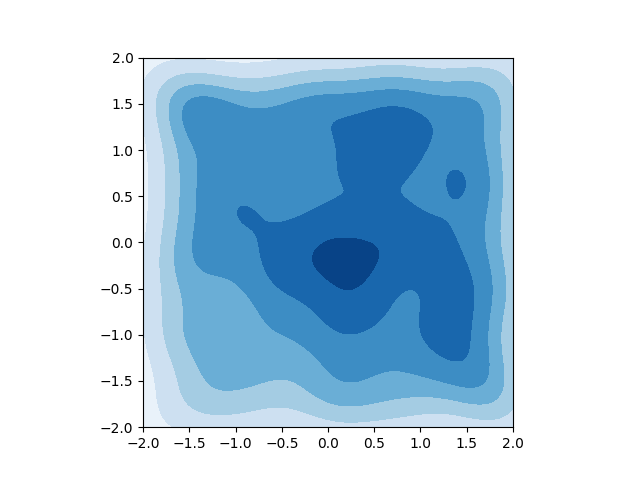}%
\includegraphics[width=0.19\textwidth]{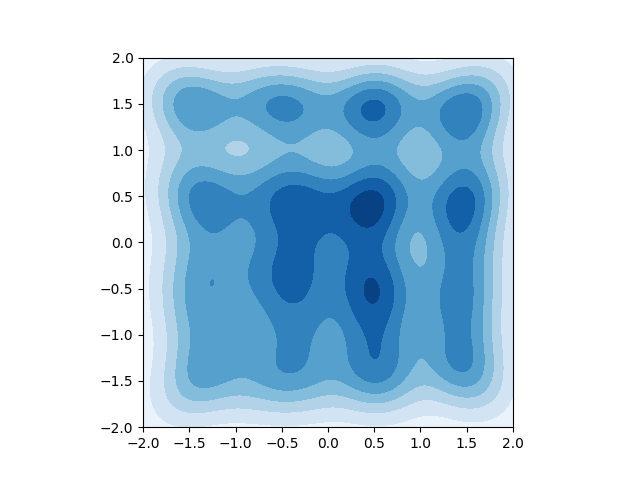}%
\includegraphics[width=0.19\textwidth]{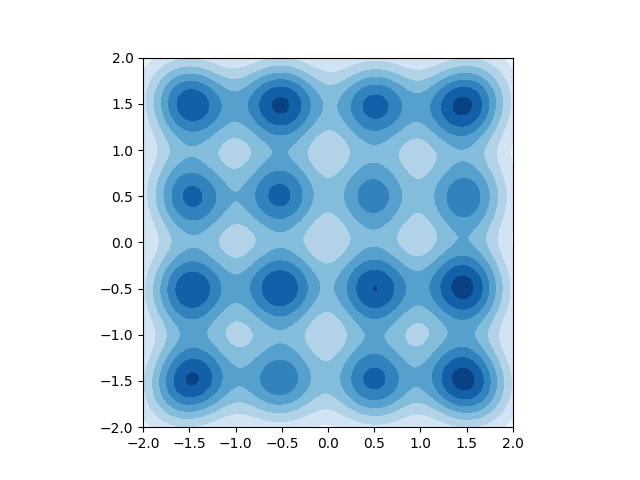}%
\includegraphics[width=0.19\textwidth]{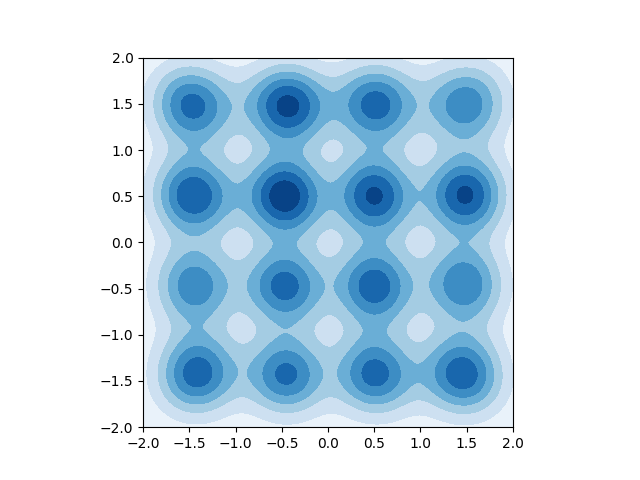}%
\includegraphics[width=0.19\textwidth]{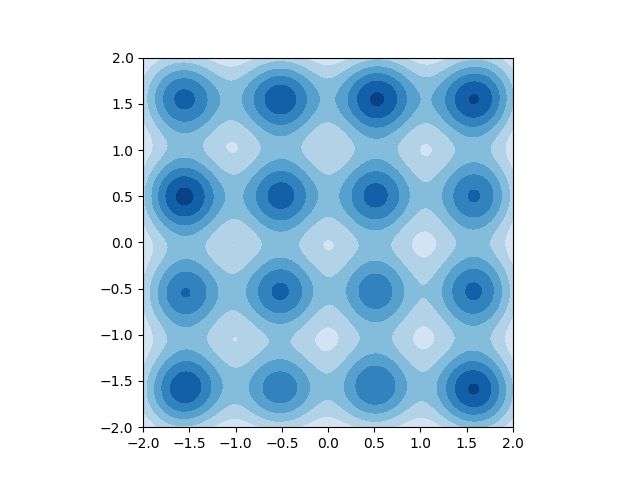}%
\end{minipage}}\\
RMSP-SGA \qquad
\subfigure{
\begin{minipage}[c]{0.95\textwidth}
\includegraphics[width=0.19\textwidth]{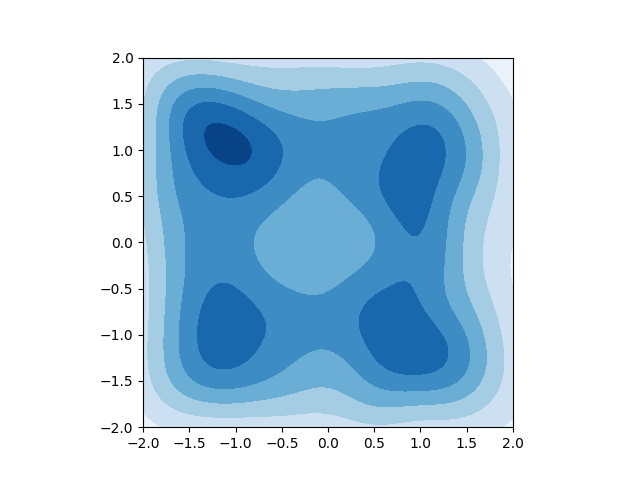}%
\includegraphics[width=0.19\textwidth]{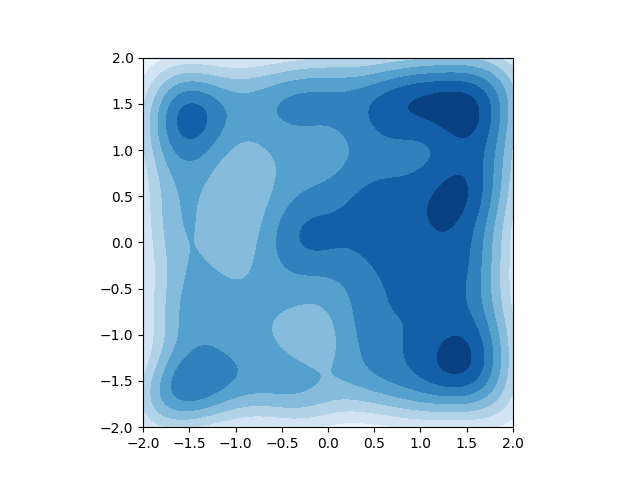}%
\includegraphics[width=0.19\textwidth]{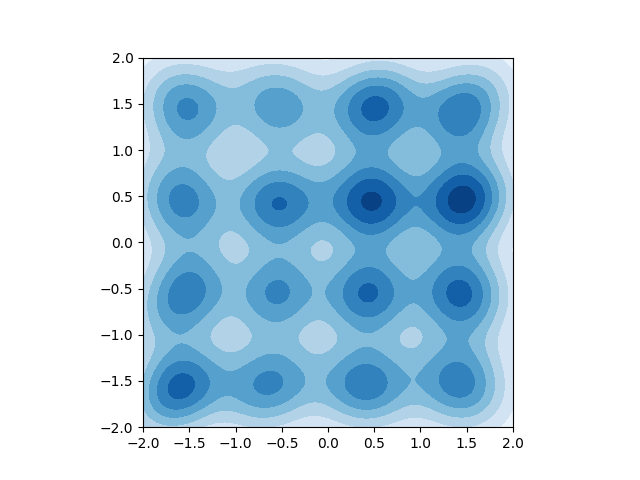}%
\includegraphics[width=0.19\textwidth]{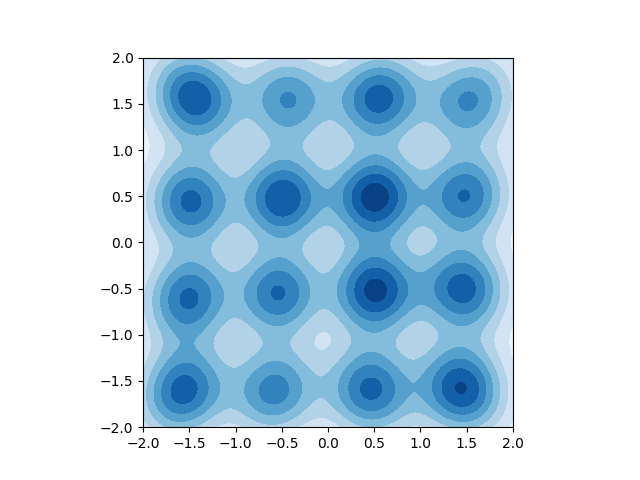}%
\includegraphics[width=0.19\textwidth]{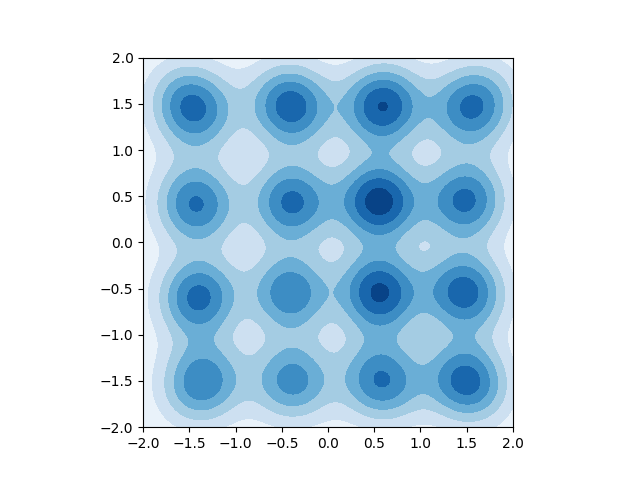}%
\end{minipage}}\\
SGA-ACG(Ours)
\subfigure{
\begin{minipage}[c]{0.179\textwidth}
\includegraphics[width=1\textwidth]{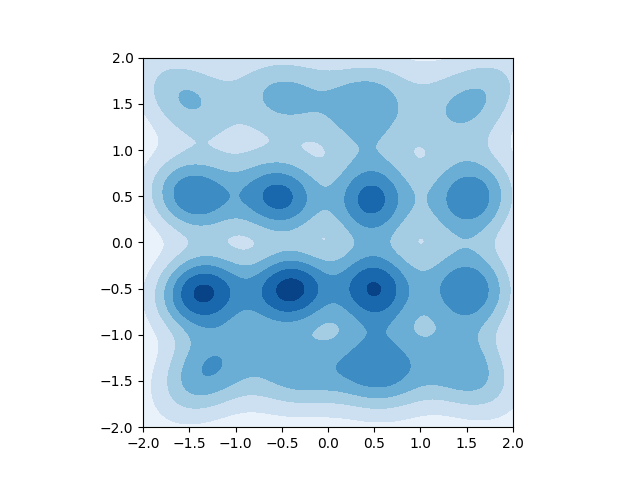}
\caption*{2000}
\end{minipage}}%
\subfigure{
\begin{minipage}[c]{0.179\textwidth}
\includegraphics[width=1\textwidth]{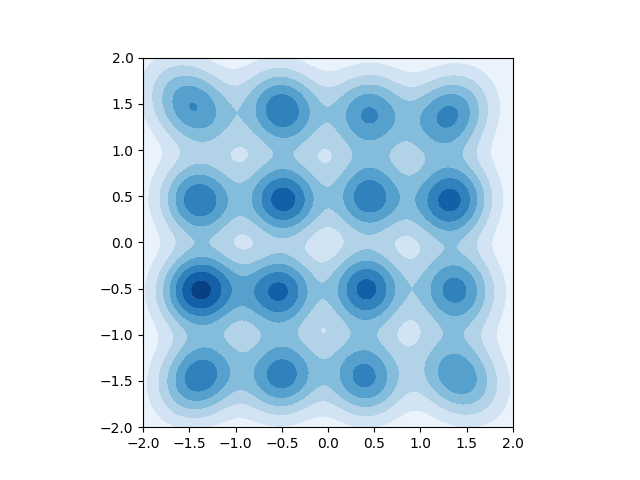}
\caption*{4000}
\end{minipage}}%
\subfigure{
\begin{minipage}[c]{0.179\textwidth}
\includegraphics[width=1\textwidth]{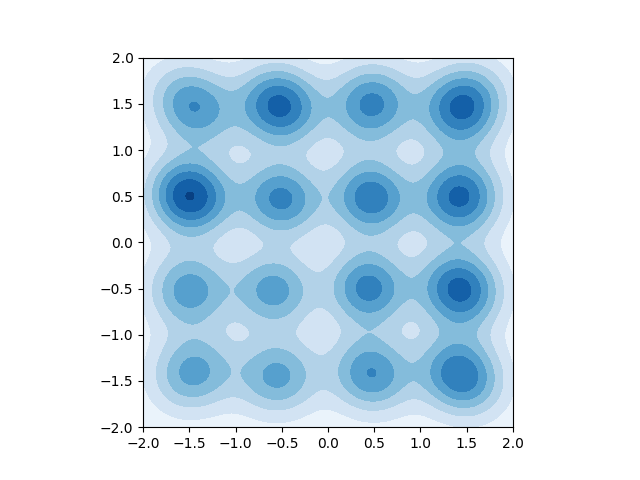}
\caption*{6000}
\end{minipage}}%
\subfigure{
\begin{minipage}[c]{0.179\textwidth}
\includegraphics[width=1\textwidth]{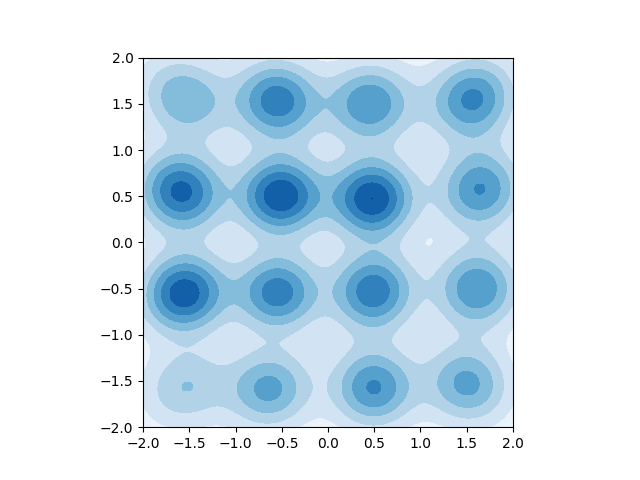}
\caption*{8000}
\end{minipage}}%
\subfigure{
\begin{minipage}[c]{0.179\textwidth}
\includegraphics[width=1\textwidth]{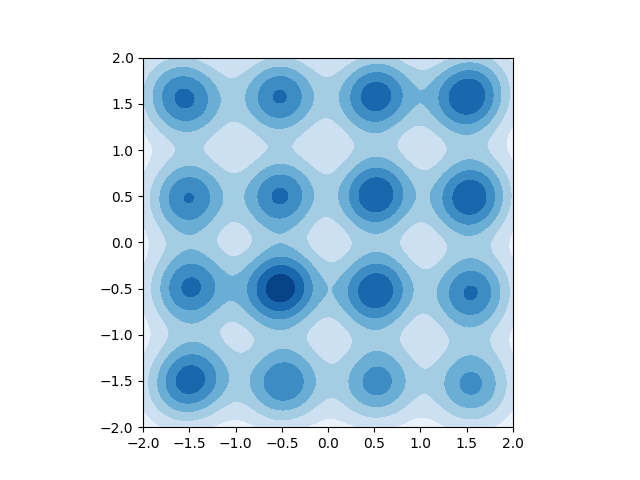}
\caption*{10000}
\end{minipage}}%
\captionsetup{font=small}
\caption{Compared results on the mixture of 16 Gaussians. Each row represents a kind of algorithms, and the columns are each algorithm in 2000,4000,6000,8000,10000 iterations, respectively.}
\label{fig7}
\end{figure}

From Figure \ref{fig7}, it is obvious that our proposed SGA-ACG method converges faster than ConOpt, RMSP-SGA. More comparisons are shown in Appendix Figure \ref{fig7-1}. From Figure \ref{fig7-1},RMSP, RMSP-alt, RMSP-ACA still do not converge after 10,000 iterations. To present the convergence speed, Figure \ref{fig:time} shows the time-consuming of the compared methods in Figure \ref{fig7-1}. There is a parameter $s$ in our proposed algorithms in Algorithm \ref{alg1} and Algorithm \ref{alg2}. To explore the influence of $s$ on final results, we conduct a series of experiments as the $s$ through $\{50,100,150,200\}$ on the mixture of 16 Gaussians as shown in Appendix Figure \ref{fig8}. From Figure \ref{fig8}, it shows that the proposed SGA-ACG method converges faster with the $s$ increasing.

\subsection{Experiments on Prevalent Datasets}
This section conducts the third experiment that tested our proposed ACG method on image generation tasks. We employ four prevalent datasets to illustrate our ACG method can be applied in deep learning. We choose the standard MNIST \cite{Minist}, Fashion-MNIST \cite{fashion}, CIFAR-10 \cite{cifar10}, CelebA \cite{celebA} datasets to conduct realistic experiments.

\subsubsection{Network Architecture}
We choose two kinds of network architectures for GANs on the MNIST dataset. For the first kind of network structure, we employ 2 fully connected layers with 256 and 512 neurons for the generator network, Each of the 2 layers is appended to a LeakyReLU layer with $\alpha=0.2$. We adopt a Tanh activation function layer as the last layer in the generator network. The input data for the generator is a random noise with dimensions 100 sampled from a standard Gaussian distribution. The output of the generator is an image with shape $(28,28,1)$. For the discriminator network, we also use 2 fully connected layers with 512 and 256 neurons. After each layer, there is appended to a LeakyReLU layer with $\alpha=0.2$, which is the same as the generator network. However, in the last layer of the discriminator, we adopt a Sigmoid activation function. The input data of the discriminator includes the generated image and the ground truth image on the MNIST dataset. The output of the discriminator is used to evaluate the quality of the image made by the generator network. For the second kind of network structure, we used the architecture of DCGANs \cite{DCGAN}, we just used 4 layers of DCGANs \cite{DCGAN} for both generator network and discriminator network.
\subsubsection{Implementation Details}

\textbf{Experimental environments.} We conduct experiments of this section on a server equipped with \textbf{CPU E5-2698}, \textbf{4*GPU GTX 3090 aero}, \textbf{24GB RAM }, Python(version 3.6.13), PyTorch(version 1.8.0). 

\textbf{Training setting.} We realized the all compared algorithms by the PyTorch, and the training set of these algorithms on the four datasets are as follows:
\begin{itemize}[leftmargin=*]
    \item SGD: The learning rate of linear GANs on MNIST dataset is $\alpha=2 \times 10^{-4}$, excepted for the learning rate of DCGANs on MNIST is  $\alpha=5 \times 10^{-4}$, Fashion-MNIST (learning rate $\alpha=2 \times 10^{-4}$ ), CIFAR-10(learning rate $\alpha=2 \times 10^{-4}$ ), CelebA(learning rate $\alpha=2 \times 10^{-4}$ ).
    \item Adam: Linear GANs on MNIST(learning rate $\alpha=3 \times 10^{-4}$), DCGANs on MNIST(learning rate $\alpha=2 \times 10^{-4}$), Fashion-MNIST (learning rate $\alpha=2 \times 10^{-4}$ ), CIFAR-10(learning rate $\alpha=2 \times 10^{-4}$ ), CelebA(learning rate $\alpha=2 \times 10^{-4}$ ).
    \item RMSP: Linear GANs on MNIST(learning rate $\alpha=2 \times 10^{-4}$), DCGANs on MNIST(learning rate $\alpha=5 \times 10^{-4}$), Fashion-MNIST (learning rate $\alpha=5 \times 10^{-4}$ ), CIFAR-10(learning rate $\alpha=5 \times 10^{-4}$ ), CelebA(learning rate $\alpha=5 \times 10^{-4}$ ).
    \item RMSP-ACG: Only on the linear GANS(learning rate $\alpha=5 \times 10^{-4},\beta_{1}=5 \times 10^{-7},\beta_{2}=\alpha$).
    \item Adam-ACG: On the all datasets, our proposed Adam-ACG method applied in linear GANs and DCGANs with learning rate $\alpha=5 \times 10^{-4},\beta_{1}=5 \times 10^{-7},\beta_{2}=\alpha$.
\end{itemize}
In the experiment of linear GANs on MNIST data, we set the batch size as 64, and the epoch number is 324. The generation results of our proposed methods are shown in Figure \ref{fig9}. More comparisons among these algorithms on MNIST are shown in Appendix Figure \ref{fig9-1}. For DCGANs experiments, the batch size is 64, and the epoch number is 110 on the MNIST dataset. The same batch size and epoch number are assigned to the Fashion-MNIST and CIFAR-10 data. In contrast, the batch size and number of epoch on CelebA data set are 128 and 70, respectively. The results of our methods on the four datasets are shown in Figure \ref{fig10}, and more comparisons results among several algorithms on the same datasets are shown in Figure \ref{fig10-1}.

\section{Conclusion}
This article proposed the Adaptive Composite Gradients(ACG) method to find a local Nash Equilibrium in a general zero-sum smooth game. Inspired by PPAC, OMD, and A3DMM, the ACG method can alleviate the cyclic behaviors in the bilinear game and training GANs. The proposed algorithm has Strong compatibility and robustness, which can easily integrate with SGD, Adam, RMSP, SGA, and other gradient-based optimizers. Since the ACG method employs the predicted information in future $s$ iterations, this is a novel semi-gradient-free algorithm. The ACG method has a linear convergence rate in a general zero-sum game, and the three parts of the experiments show that our algorithm is more preferred and faster than previous works. Furthermore, we offer that the SGA-ACG can be competitive to ConOpt and SGA methods on the mixture of Gaussians generated tasks. Finally, we prove our ACG method can be promoted and applied in a general zero-sum game with $n$ players by toy function experiment.

However, our research objectives just are limited to the convex and smoothness of simple zero-sum games. The non-convex and non-smooth games are more complex and challenging to find a local Nash Equilibrium. Therefore, optimizing and finding local solutions for non-convex and non-smooth games is still a challenging task worth researching in the future.
\end{multicols}

\begin{figure}[H]\tiny
RMSP-ACG(ours)
\subfigure{
\begin{minipage}[c]{0.92\textwidth}
\includegraphics[width=0.23\textwidth]{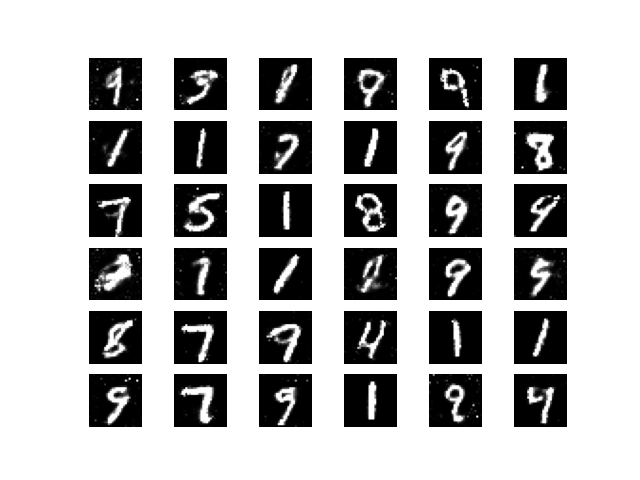}%
\includegraphics[width=0.23\textwidth]{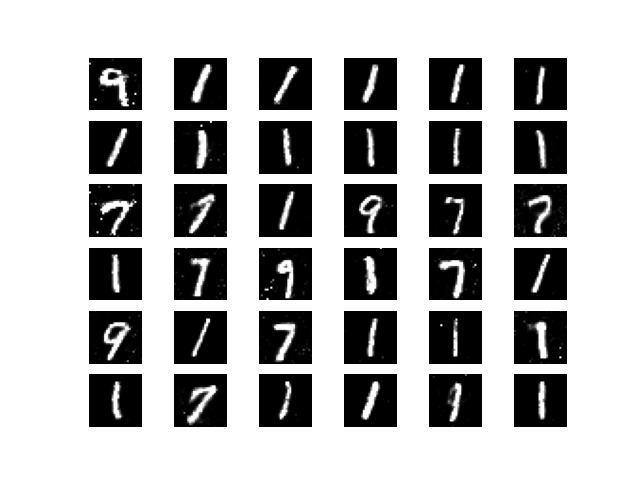}%
\includegraphics[width=0.23\textwidth]{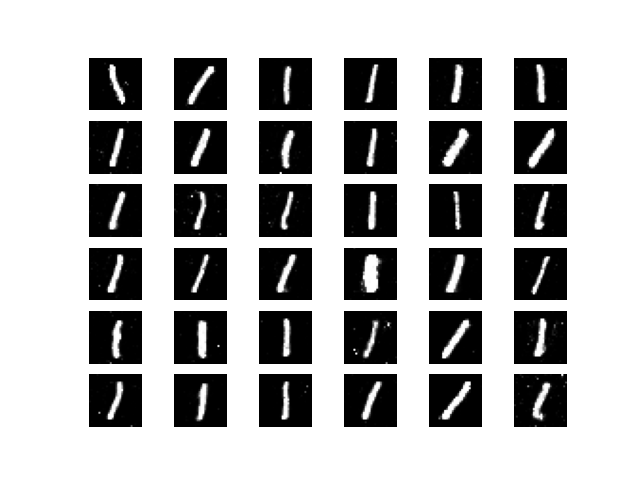}%
\includegraphics[width=0.23\textwidth]{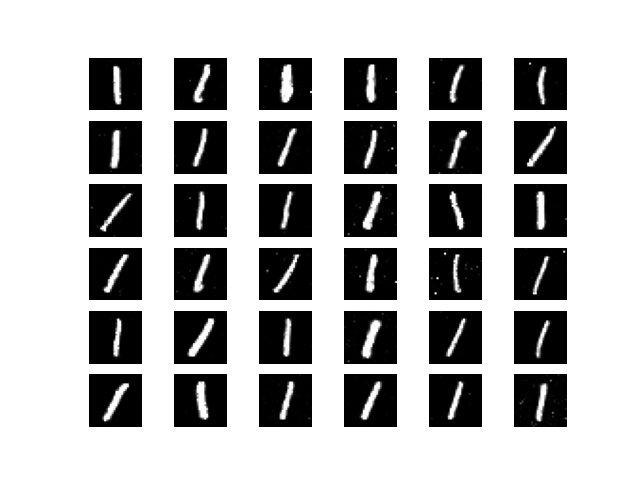}%
\end{minipage}}\\
Adam-ACG(ours)
\subfigure{
\begin{minipage}[c]{0.21\textwidth}
\includegraphics[width=1\textwidth]{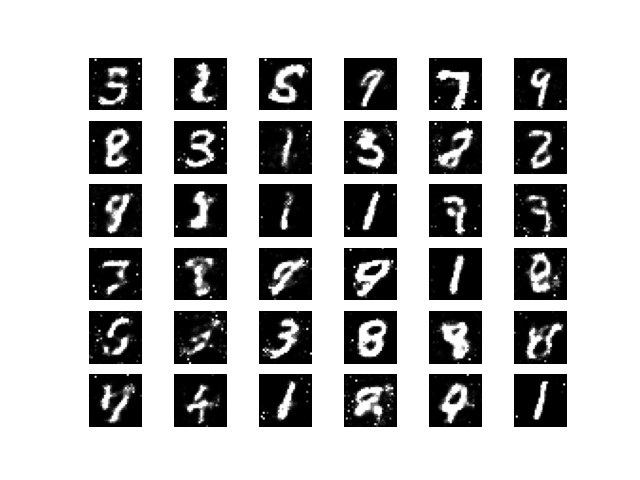}%
\caption*{50k}
\end{minipage}}%
\subfigure{
\begin{minipage}[c]{0.21\textwidth}
\includegraphics[width=1\textwidth]{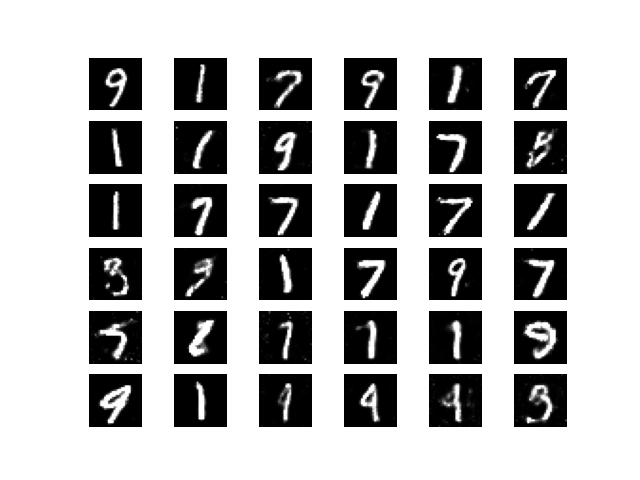}%
\caption*{150k}
\end{minipage}}%
\subfigure{
\begin{minipage}[c]{0.21\textwidth}
\includegraphics[width=1\textwidth]{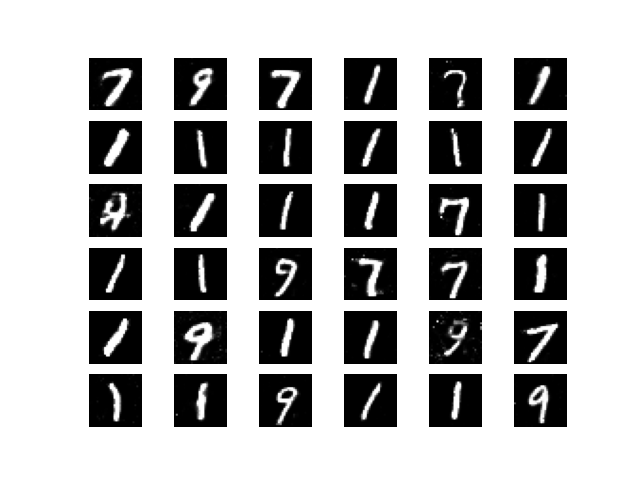}%
\caption*{250k}
\end{minipage}}%
\subfigure{
\begin{minipage}[c]{0.21\textwidth}
\includegraphics[width=1\textwidth]{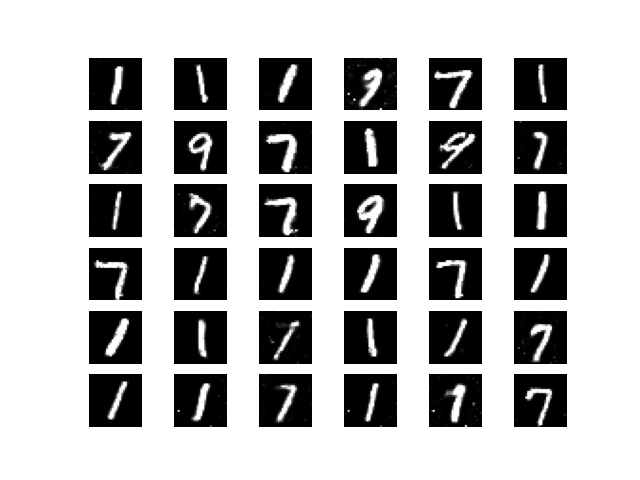}%
\caption*{300k}
\end{minipage}}%
\captionsetup{font={small}}
\caption{Compared results of Linear GANs on MNIST dataset. The first and second rows are the results of RMSP-ACG, Adam-ACG, The first, second, third, and fourth columns are the results in 50000,150000,250000, and 300000 iterations,respectively.}
\label{fig9}
\end{figure}
\begin{figure}[H]\tiny
Adam-ACG(ours)
\subfigure{
\begin{minipage}[c]{0.21\textwidth}
\includegraphics[width=1\textwidth]{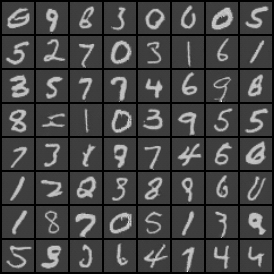}
\caption*{MNIST 100k}
\end{minipage}}%
\subfigure{
\begin{minipage}[c]{0.21\textwidth}
\includegraphics[width=1\textwidth]{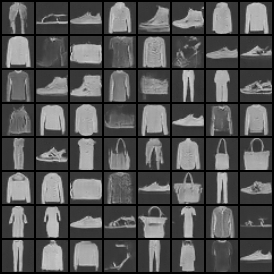}
\caption*{Fashion-MNIST 100k}
\end{minipage}}%
\subfigure{
\begin{minipage}[c]{0.21\textwidth}
\includegraphics[width=1\textwidth]{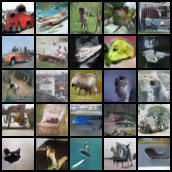}
\caption*{CIFAR10 80k}
\end{minipage}}%
\subfigure{
\begin{minipage}[c]{0.21\textwidth}
\includegraphics[width=1\textwidth]{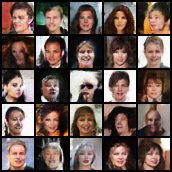}
\caption*{CelebA 100k}
\end{minipage}}%
\captionsetup{font={small}}
\caption{Comparison of DCGAN for our proposed method on the four datasets. The first column is the results of the MNIST dataset in 100000 iterations. The second column is the results of the Fashion-MNIST dataset in 100000 iterations. The third column is the results of the CIFAR10 dataset in 80000 iterations, and the last column is the results of CelebA dataset in 100000 iterations.}
\label{fig10}
\end{figure}

\begin{multicols}{2}
\section*{Acknowledgement}
This work is supported by the National Key Research and Development Program of China (No.2018AAA0101001), Science and Technology Commission of Shanghai Municipality (No.20511100200), and supported in part by the Science and Technology Commission of Shanghai Municipality (No.18dz2271000).

%\bibliography{references}

\begin{thebibliography}{10}

\bibitem{pmlr-v80}
D.~Balduzzi, S.~Racaniere, J.~Martens, J.~Foerster, K.~Tuyls, and T.~Graepel.
\newblock The mechanics of n-player differentiable games.
\newblock In J.~Dy and A.~Krause, editors, {\em Proceedings of the 35th
  International Conference on Machine Learning}, volume~80 of {\em Proceedings
  of Machine Learning Research}, pages 354--363. PMLR, 10--15 Jul 2018.

\bibitem{book1}
H.~H. Bauschke, P.~L. Combettes, et~al.
\newblock {\em Convex analysis and monotone operator theory in Hilbert spaces},
  volume 408.
\newblock Springer, 2011.

\bibitem{BUCK1977}
R.~E. Bruck.
\newblock On the weak convergence of an ergodic iteration for the solution of
  variational inequalities for monotone operators in hilbert space.
\newblock {\em Journal of Mathematical Analysis and Applications},
  61(1):159--164, 1977.

\bibitem{chavdarova2020}
T.~Chavdarova, M.~Pagliardini, S.~U. Stich, F.~Fleuret, and M.~Jaggi.
\newblock Taming gans with lookahead-minmax, 2020.

\bibitem{Croce2020}
D.~Croce, G.~Castellucci, and R.~Basili.
\newblock {GAN}-{BERT}: Generative adversarial learning for robust text
  classification with a bunch of labeled examples.
\newblock In {\em Proceedings of the 58th Annual Meeting of the Association for
  Computational Linguistics}, pages 2114--2119, Online, July 2020. Association
  for Computational Linguistics.

\bibitem{dai2017}
W.~Dai, J.~Doyle, X.~Liang, H.~Zhang, N.~Dong, Y.~Li, and E.~P. Xing.
\newblock Scan: Structure correcting adversarial network for organ segmentation
  in chest x-rays, 2017.

\bibitem{gidel2020}
G.~Gidel, H.~Berard, G.~Vignoud, P.~Vincent, and S.~Lacoste-Julien.
\newblock A variational inequality perspective on generative adversarial
  networks, 2020.

\bibitem{lee2018}
S.~gil Lee, U.~Hwang, S.~Min, and S.~Yoon.
\newblock Polyphonic music generation with sequence generative adversarial
  networks, 2018.

\bibitem{goodfellow2017}
I.~Goodfellow.
\newblock Nips 2016 tutorial: Generative adversarial networks, 2017.

\bibitem{goodfellow2014}
I.~J. Goodfellow, J.~Pouget-Abadie, M.~Mirza, B.~Xu, D.~Warde-Farley, S.~Ozair,
  A.~Courville, and Y.~Bengio.
\newblock Generative adversarial networks, 2014.

\bibitem{guimaraes2018}
G.~L. Guimaraes, B.~Sanchez-Lengeling, C.~Outeiral, P.~L.~C. Farias, and
  A.~Aspuru-Guzik.
\newblock Objective-reinforced generative adversarial networks (organ) for
  sequence generation models, 2018.

\bibitem{haidar2019}
M.~A. Haidar and M.~Rezagholizadeh.
\newblock Textkd-gan: Text generation using knowledgedistillation and
  generative adversarial networks, 2019.

\bibitem{Hong2019}
Y.~Hong, U.~Hwang, J.~Yoo, and S.~Yoon.
\newblock How generative adversarial networks and their variants work.
\newblock {\em ACM Computing Surveys}, 52(1):1–43, Feb 2019.

\bibitem{hsu2017}
C.-C. Hsu, H.-T. Hwang, Y.-C. Wu, Y.~Tsao, and H.-M. Wang.
\newblock Voice conversion from unaligned corpora using variational
  autoencoding wasserstein generative adversarial networks, 2017.

\bibitem{keke2020}
L.~Keke, Z.~Ke, L.~Qiang, and Y.~Xinmin.
\newblock Training gans with predictive projection centripetal acceleration,
  2020.

\bibitem{kim2017}
T.~Kim, M.~Cha, H.~Kim, J.~K. Lee, and J.~Kim.
\newblock Learning to discover cross-domain relations with generative
  adversarial networks, 2017.

\bibitem{cifar10}
A.~Krizhevsky, G.~Hinton, et~al.
\newblock Learning multiple layers of features from tiny images.
\newblock 2009.

\bibitem{DeblurGAN}
O.~Kupyn, V.~Budzan, M.~Mykhailych, D.~Mishkin, and J.~Matas.
\newblock Deblurgan: Blind motion deblurring using conditional adversarial
  networks.
\newblock {\em ArXiv e-prints}, 2017.

\bibitem{Minist}
Y.~Lecun, L.~Bottou, Y.~Bengio, and P.~Haffner.
\newblock Gradient-based learning applied to document recognition.
\newblock {\em Proceedings of the IEEE}, 86(11):2278--2324, 1998.

\bibitem{ledig2017}
C.~Ledig, L.~Theis, F.~Huszar, J.~Caballero, A.~Cunningham, A.~Acosta,
  A.~Aitken, A.~Tejani, J.~Totz, Z.~Wang, and W.~Shi.
\newblock Photo-realistic single image super-resolution using a generative
  adversarial network, 2017.

\bibitem{lee2017}
J.~D. Lee, I.~Panageas, G.~Piliouras, M.~Simchowitz, M.~I. Jordan, and
  B.~Recht.
\newblock First-order methods almost always avoid saddle points, 2017.

\bibitem{Li2020}
K.~{Li}, T.~{Zhang}, and R.~{Wang}.
\newblock Deep reinforcement learning for multiobjective optimization.
\newblock {\em IEEE Transactions on Cybernetics}, pages 1--12, 2020.

\bibitem{liang19b}
T.~Liang and J.~Stokes.
\newblock Interaction matters: A note on non-asymptotic local convergence of
  generative adversarial networks.
\newblock In K.~Chaudhuri and M.~Sugiyama, editors, {\em Proceedings of Machine
  Learning Research}, volume~89 of {\em Proceedings of Machine Learning
  Research}, pages 907--915. PMLR, 16--18 Apr 2019.

\bibitem{Sim2018}
T.~Liang and J.~Stokes.
\newblock Interaction matters: A note on non-asymptotic local convergence of
  generative adversarial networks, 2019.

\bibitem{lin2018}
K.~Lin, D.~Li, X.~He, Z.~Zhang, and M.-T. Sun.
\newblock Adversarial ranking for language generation, 2018.

\bibitem{celebA}
Z.~Liu, P.~Luo, X.~Wang, and X.~Tang.
\newblock Deep learning face attributes in the wild.
\newblock In {\em Proceedings of International Conference on Computer Vision
  (ICCV)}, December 2015.

\bibitem{8417964}
M.~{Mardani}, E.~{Gong}, J.~Y. {Cheng}, S.~S. {Vasanawala}, G.~{Zaharchuk},
  L.~{Xing}, and J.~M. {Pauly}.
\newblock Deep generative adversarial neural networks for compressive sensing
  mri.
\newblock {\em IEEE Transactions on Medical Imaging}, 38(1):167--179, 2019.

\bibitem{mescheder2018}
L.~Mescheder, S.~Nowozin, and A.~Geiger.
\newblock The numerics of gans, 2018.

\bibitem{pmlr-v108}
K.~Mishchenko, D.~Kovalev, E.~Shulgin, P.~Richtarik, and Y.~Malitsky.
\newblock Revisiting stochastic extragradient.
\newblock In S.~Chiappa and R.~Calandra, editors, {\em Proceedings of the
  Twenty Third International Conference on Artificial Intelligence and
  Statistics}, volume 108 of {\em Proceedings of Machine Learning Research},
  pages 4573--4582. PMLR, 26--28 Aug 2020.

\bibitem{pmlr-v108M}
A.~Mokhtari, A.~Ozdaglar, and S.~Pattathil.
\newblock A unified analysis of extra-gradient and optimistic gradient methods
  for saddle point problems: Proximal point approach.
\newblock In S.~Chiappa and R.~Calandra, editors, {\em Proceedings of the
  Twenty Third International Conference on Artificial Intelligence and
  Statistics}, volume 108 of {\em Proceedings of Machine Learning Research},
  pages 1497--1507. PMLR, 26--28 Aug 2020.

\bibitem{NAO2009}
A.~Nedi{\'c} and A.~Ozdaglar.
\newblock Subgradient methods for saddle-point problems.
\newblock {\em Journal of Optimization Theory and Applications},
  142(1):205--228, Mar. 2009.

\bibitem{Nesterov1983}
Y.~Nesterov.
\newblock A method for solving the convex programming problem with convergence
  rate $o(1/k^2)$.
\newblock {\em Proceedings of the USSR Academy of Sciences}, 269:543--547,
  1983.

\bibitem{neyshabur2018}
B.~Neyshabur, S.~Bhojanapalli, and A.~Chakrabarti.
\newblock Stabilizing gan training with multiple random projections, 2018.

\bibitem{NIPS2016}
S.~Nowozin, B.~Cseke, and R.~Tomioka.
\newblock f-gan: Training generative neural samplers using variational
  divergence minimization.
\newblock In D.~Lee, M.~Sugiyama, U.~Luxburg, I.~Guyon, and R.~Garnett,
  editors, {\em Advances in Neural Information Processing Systems}, volume~29.
  Curran Associates, Inc., 2016.

\bibitem{odena2019}
A.~Odena.
\newblock Open questions about generative adversarial networks.
\newblock {\em Distill}, 2019.
\newblock https://distill.pub/2019/gan-open-problems.

\bibitem{Peng_2020}
W.~Peng, Y.-H. Dai, H.~Zhang, and L.~Cheng.
\newblock Training gans with centripetal acceleration.
\newblock {\em Optimization Methods and Software}, 35(5):955–973, Apr 2020.

\bibitem{2019Liang}
C.~{Poon} and J.~{Liang}.
\newblock {Trajectory of Alternating Direction Method of Multipliers and
  Adaptive Acceleration}.
\newblock {\em arXiv e-prints}, page arXiv:1906.10114, June 2019.

\bibitem{qin2020}
C.~Qin, Y.~Wu, J.~T. Springenberg, A.~Brock, J.~Donahue, T.~P. Lillicrap, and
  P.~Kohli.
\newblock Training generative adversarial networks by solving ordinary
  differential equations, 2020.

\bibitem{DCGAN}
A.~Radford, L.~Metz, and S.~Chintala.
\newblock Unsupervised representation learning with deep convolutional
  generative adversarial networks, 2016.

\bibitem{tulyakov2017}
S.~Tulyakov, M.-Y. Liu, X.~Yang, and J.~Kautz.
\newblock Mocogan: Decomposing motion and content for video generation, 2017.

\bibitem{vezhnevets2017}
A.~S. Vezhnevets, S.~Osindero, T.~Schaul, N.~Heess, M.~Jaderberg, D.~Silver,
  and K.~Kavukcuoglu.
\newblock Feudal networks for hierarchical reinforcement learning, 2017.

\bibitem{NIPS2016_04025959}
C.~Vondrick, H.~Pirsiavash, and A.~Torralba.
\newblock Generating videos with scene dynamics.
\newblock In D.~Lee, M.~Sugiyama, U.~Luxburg, I.~Guyon, and R.~Garnett,
  editors, {\em Advances in Neural Information Processing Systems}, volume~29.
  Curran Associates, Inc., 2016.

\bibitem{walker2017}
J.~Walker, K.~Marino, A.~Gupta, and M.~Hebert.
\newblock The pose knows: Video forecasting by generating pose futures, 2017.

\bibitem{Wang_2018}
C.~Wang, C.~Xu, C.~Wang, and D.~Tao.
\newblock Perceptual adversarial networks for image-to-image transformation.
\newblock {\em IEEE Transactions on Image Processing}, 27(8):4066–4079, Aug
  2018.

\bibitem{9054014}
J.~{Wang}, V.~{Tantia}, N.~{Ballas}, and M.~{Rabbat}.
\newblock Lookahead converges to stationary points of smooth non-convex
  functions.
\newblock In {\em ICASSP 2020 - 2020 IEEE International Conference on
  Acoustics, Speech and Signal Processing (ICASSP)}, pages 8604--8608, 2020.

\bibitem{wang2020}
Y.~Wang.
\newblock A mathematical introduction to generative adversarial nets (gan),
  2020.

\bibitem{wu2017}
J.~Wu, C.~Zhang, T.~Xue, W.~T. Freeman, and J.~B. Tenenbaum.
\newblock Learning a probabilistic latent space of object shapes via 3d
  generative-adversarial modeling, 2017.

\bibitem{fashion}
H.~Xiao, K.~Rasul, and R.~Vollgraf.
\newblock Fashion-mnist: a novel image dataset for benchmarking machine
  learning algorithms, 2017.

\bibitem{yang2017}
D.~Yang, T.~Xiong, D.~Xu, Q.~Huang, D.~Liu, S.~K. Zhou, Z.~Xu, J.~Park,
  M.~Chen, T.~D. Tran, S.~P. Chin, D.~Metaxas, and D.~Comaniciu.
\newblock Automatic vertebra labeling in large-scale 3d ct using deep
  image-to-image network with message passing and sparsity regularization,
  2017.

\bibitem{2016unified}
T.~Yang, Q.~Lin, and Z.~Li.
\newblock Unified convergence analysis of stochastic momentum methods for
  convex and non-convex optimization, 2016.

\bibitem{yi2018}
Z.~Yi, H.~Zhang, P.~Tan, and M.~Gong.
\newblock Dualgan: Unsupervised dual learning for image-to-image translation,
  2018.

\bibitem{yu2017}
L.~Yu, W.~Zhang, J.~Wang, and Y.~Yu.
\newblock Seqgan: Sequence generative adversarial nets with policy gradient,
  2017.

\bibitem{yue2020dual}
Z.~Yue, Q.~Zhao, L.~Zhang, and D.~Meng.
\newblock Dual adversarial network: Toward real-world noise removal and noise
  generation, 2020.

\bibitem{zhang2020}
G.~Zhang and Y.~Yu.
\newblock Convergence of gradient methods on bilinear zero-sum games, 2020.

\bibitem{zhu2020}
J.-Y. Zhu, T.~Park, P.~Isola, and A.~A. Efros.
\newblock Unpaired image-to-image translation using cycle-consistent
  adversarial networks, 2020.

\end{thebibliography}
\end{multicols}

%%%%%%%%%%%%%%%%%%%%%%%%%%%%%%%%%%%
%%%%%%%%%%Appendiences%%%%%%%%%%%%%
%%%%%%%%%%%%%%%%%%%%%%%%%%%%%%%%%%%
\clearpage
\appendix
\setcounter{equation}{0}
\renewcommand\theequation{A.\arabic{equation}}
\section{Proofs in Section 5}
\subsection{Proof of Proposition \ref{p1}}
\begin{proof*}
Without loss of generality, let $\vec{a}=(a_{1},a_{2},a_{3},\cdots,a_{n}),\vec{b}=(b_{1},b_{2},b_{3},\cdots, b_{n})$, where $n\geq 2, n \in N^{*}.$ Then, we have 
\begin{align}\label{A1}
    cos<\vec{a},\vec{b}>=\frac{\vec{a}\cdot\vec{b}}{|\vec{a}||\vec{b}|}.
\end{align}
The projection $\vec{p}$ of $\vec{b}$ onto $\vec{a}$ can be written as 
\begin{align}\label{A2}
    \vec{p}=|\vec{b}|cos<\vec{a},\vec{b}> \frac{\vec{a}}{|\vec{a}|}. 
\end{align}
Incorporating \ref{A1} into \ref{A2}, we have 
\begin{align*}
    \vec{p}&=|\vec{b}| \cdot \frac{\vec{a}}{|\vec{a}|} \cdot \frac{\vec{a}\cdot\vec{b}}{|\vec{a}||\vec{b}|}\\
           &=\frac{\vec{a}\cdot\vec{b}}{||\vec{a}||_{2}^{2}} \cdot \vec{a}.
\end{align*}
Using the $\gamma$ to replace $\frac{\vec{a}\cdot\vec{b}}{||\vec{a}||_{2}^{2}}$ we can obtain 
$$\vec{p} = \gamma \vec{a},$$ 
where the $\gamma \in \mathbb{R}$.
\end{proof*}
\subsection{Proof of Proposition \ref{p2}}
\begin{proof}
The characteristic polynomial of the matrix (\ref{matrix}) is  
\begin{align}
      det \begin{pmatrix} (1-\lambda)I_{d} & -(\alpha+\beta_{1})\emph{A} & 0 & \beta_{1}(1+\gamma)\emph{A} & -\beta_{2}I_{d} & 0 \\ 
         (\alpha+\beta_{1})\emph{A}^{\mathrm{T}} & (1-\lambda)I_{d} & -\beta_{1}(1+\gamma)\emph{A}^{\mathrm{T}} & 0 & 0 & \beta_{2}I_{d} \\ 
         I_{d} & 0 & -\lambda I_{d} & 0 & 0 & 0 \\ 
         0 & I_{d} & 0 & -\lambda I_{d} & 0 & 0 \\ 
         0 & 0 & 0 & 0 & (\tau-\lambda) I_{d}  & 0 \\ 
         0 & 0 & 0 & 0 & 0 & (\tau-\lambda) I_{d} \\ 
    \end{pmatrix},
\end{align}
which is equivalent to 
\begin{align}\label{A4}
      (\tau-\lambda)^{2}\cdot det \begin{pmatrix} (1-\lambda)I_{d} & -(\alpha+\beta_{1})\emph{A} & 0 & \beta_{1}(1+\gamma)\emph{A} \\ 
         (\alpha+\beta_{1})\emph{A}^{\mathrm{T}} & (1-\lambda)I_{d} & -\beta_{1}(1+\gamma)\emph{A}^{\mathrm{T}} & 0  \\ 
         I_{d} & 0 & -\lambda I_{d} & 0 \\ 
         0 & I_{d} & 0 & -\lambda I_{d}\\ 
    \end{pmatrix}.
\end{align}
From (\ref{A4}) we can derive to 
\begin{align}\label{A5}
    (\tau-\lambda)^{2}\cdot \begin{pmatrix} \lambda(1-\lambda)I_{d} & -\lambda (\alpha+\beta_{1})\emph{A}+ \beta_{1}(1+\gamma)\emph{A} \\ 
         \lambda(\alpha+\beta_{1})\emph{A}^{\mathrm{T}}-\beta_{1}(1+\gamma)\emph{A}^{\mathrm{T}}& \lambda(1-\lambda)I_{d}  \\ 
    \end{pmatrix},
\end{align}
According to (\ref{A5}), 0 and 1 can not be its roots based on $\emph{A}$ is nonsingular and square. so (\ref{A5}) is equivalent to 
\begin{align}
    det((\tau-\lambda)^{2}[\lambda^{2}(1-\lambda)^{2}+(\lambda(\alpha+\beta_{1})-\beta_{1}(1+\gamma))^{2}\emph{A}^{\mathrm{T}}\emph{A}])
\end{align}
Then, we can obtain that the eigenvalues of $\emph{F}$ are the roots of the sixth order polynomials:
\begin{align*}
     (\tau-\lambda)^{2}[\lambda^{2}(1-\lambda)^{2}+(\lambda(\alpha+\beta_{1})-\beta_{1}(1+\gamma))^{2}\xi^{2}], \quad \xi^{2} \in Sp(\emph{A}^{\mathrm{T}}\emph{A}).
\end{align*}
\end{proof}

\subsection{Proof of Proposition \ref{p3}}
\begin{proof}
Let the characteristic polynomial of the matrix (\ref{matrix}) to be 0, which is written as follows:
\begin{align} \label{A6}
     (\tau-\lambda)^{2}[\lambda^{2}(1-\lambda)^{2}+(\lambda(\alpha+\beta_{1})-\beta_{1}(1+\gamma))^{2}\xi^{2}]=0, \quad \xi^{2} \in Sp(\emph{A}^{\mathrm{T}}\emph{A}).
\end{align}
It is obvious that (\ref{A6}) have 6 roots, and $\lambda_{1}=\lambda_{2}=\tau$ are two of the 6 roots. According to the convergence of formula (\ref{f12}), we can obtain the $\tau$ is almost small and $|\tau|<1$. However, in this case, whatever the values of the $\alpha,\beta_{1} and \gamma$ are, the dynamic system will converge to the Nash Equilibrium, which is meaningless. So we mainly discuss the following polynomial:
\begin{align*}
     [\lambda^{2}(1-\lambda)^{2}+(\lambda(\alpha+\beta_{1})-\beta_{1}(1+\gamma))^{2}\xi^{2}]=0.
\end{align*}
Using Proposition (\ref{p2}), we have 
\begin{align}\label{A7}
(\lambda^{2}-\lambda-i[\lambda(\alpha+\beta_{1})-\beta_{1}(1+\gamma)]\xi)(\lambda^{2}-\lambda+i[\lambda(\alpha+\beta_{1})-\beta_{1}(1+\gamma)]\xi)=0
\end{align}
Denote $a:=\alpha+\beta_{1}$ and $b:=\beta_{1}(1+\gamma)$, then (\ref{A7}) can be written as:
\begin{align}\label{A8}
[\lambda^{2}-\lambda-i(\lambda a-b)\xi][\lambda^{2}-\lambda+i(\lambda a-b)\xi]=0.
\end{align}
we can get the four roots of (\ref{A8}) are
\begin{align*}
    \lambda_{1}^{\pm} = \frac{1-ia\xi \pm \sqrt{1-a^{2}\xi^{2}-2ia\xi+4ib\xi}}{2};\\
    \lambda_{2}^{\pm} = \frac{1+ia\xi \pm \sqrt{1-a^{2}\xi^{2}+2ia\xi-4ib\xi}}{2}.\\
\end{align*}
Let $u:=a\xi+b\xi$ and $v:=a\xi-b\xi$, then we can obtain
\begin{align*}
    \lambda_{1}^{\pm} = \frac{1-(\frac{u+v}{2})i \pm \sqrt{1-(\frac{u+v}{2})^{2}-(3v-u)i}}{2};\\
    \lambda_{2}^{\pm} =  \frac{1+(\frac{u+v}{2})i \pm \sqrt{1-(\frac{u+v}{2})^{2}+(3v-u)i}}{2}.\\
\end{align*}
Denote $s:=\frac{u+v}{2}$ and $t:=\frac{3v-u}{2}$, then we have
\begin{align*}\label{A10}
    \lambda_{1}^{\pm} = \frac{1-si \pm \sqrt{1-2ti-s^{2}}}{2};\\
    \lambda_{2}^{\pm} = \frac{1+si \pm \sqrt{1+2ti-s^{2}}}{2}.\\
\end{align*}
The following proof process is the same as \textbf{(A.2)} in \cite{Peng_2020}. For a given complex number $z$, we can obtain the absolute value of the real part in $z$ is $\sqrt{\frac{|z|+R(z)}{2}}$ and the absolute value of the imaginary part in $z$ is $\sqrt{\frac{|z|-R(z)}{2}}$, However, According to this Proposition $s\leq 1$, all the real parts of the four roots lie in the interval $[-\mathcal{S},\mathcal{S}]$, where 
\begin{align}\label{A18}
    \mathcal{S}=\frac{1}{2}\sqrt{\frac{\sqrt{(1-s)^{2}+4t^{2}}+1-s^{2}}{2}}+\frac{1}{2},
\end{align}
all the imaginary parts of roots lie in the interval $[-\mathcal{T},\mathcal{T}]$, where
\begin{align}\label{A19}
    \mathcal{T} = \frac{1}{2}\sqrt{\frac{\sqrt{(1-s)^{2}+4t^{2}}-1+s^{2}}{2}}+\frac{s}{2}.
\end{align}
Using the Inequality $\sqrt{x+y} \leq \sqrt{x}+\frac{y}{2\sqrt{x}}, (x>0,y>0)$, we can obtain
\begin{align}\label{A15}
    \mathcal{S} &\leq \frac{1}{2}\sqrt{1-s^{2}+\frac{t^{2}}{1-s^{2}}}+\frac{1}{2}\\
\label{A16}    \mathcal{T} &\leq \frac{s}{2}+\frac{|t|}{2\sqrt{1-s^{2}}}
\end{align}
Then, we analyze the $s$ in $(0,\frac{1}{\sqrt{2}}]$ and $(\frac{1}{\sqrt{2}},1]$ two cases separately. \textbf{Case 1}, we suppose $0<s\leq \frac{1}{\sqrt{2}}$, According to this proposition $(|\alpha+\beta_{1}|+|2\beta_{1}(1+\gamma)|)/(\alpha+\beta_{1})^{2}\leq 0.1 \xi$, for all $\xi^{2} \in Sp(\emph{A}^{\mathrm{T}}\emph{A})$, we have
\begin{align*}
    |t|\leq\frac{s^{2}}{10}.
\end{align*}
Then, based on $\frac{s^{2}}{2}\leq 1-\sqrt{1-s^{2}}$, we have
\begin{align}\label{A11}
    |t| \leq \frac{1-\sqrt{1-s^{2}}}{5}. 
\end{align}
Integrating $s\leq \frac{1}{\sqrt{2}}$ with (\ref{A11}), we can obtain 
\begin{align}
    |t|\leq \frac{2(1-\sqrt{1-s^{2}})(1-s^{2})}{5}\leq \frac{(1-\sqrt{1-s^{2}})(1-s^{2})}{2\sqrt{1-s^{2}}+\frac{1}{2}},
\end{align}
which follows that 
\begin{align}\label{A14}
    1 &\geq \frac{|t|}{\sqrt{1-s^{2}}}+\sqrt{1-s^{2}}+\frac{|t|}{2(1-s^{2})}+\frac{|t|}{\sqrt{1-s^{2}}}\\ 
\label{A12}   &\geq \frac{t^{2}}{1-s^{2}}+\sqrt{1-s^{2}}+\frac{t^{2}}{2(1-s^{2})^{\frac{3}{2}}}+\frac{s|t|}{\sqrt{1-s^{2}}}\\ 
\label{A13}   &\geq \frac{t^{2}}{1-s^{2}}+\sqrt{1-s^{2}+\frac{t^{2}}{1-s^{2}}}+\frac{s|t|}{\sqrt{1-s^{2}}}. 
\end{align}
The inequality (\ref{A12}) follows by the fact that $\frac{|t|}{\sqrt{1-s^{2}}} \leq \frac{s}{\sqrt{1-s^{2}}} \leq 1$ and the inequality (\ref{A13}) uses $\sqrt{x+y} \leq \sqrt{x}+\frac{y}{2\sqrt{x}}$. The (\ref{A14} - \ref{A13}) can be written equivalently to
\begin{align*}
    (\frac{1}{2}\sqrt{1-s^{2}+\frac{t^{2}}{1-s^{2}}}+\frac{1}{2})^{2} + (\frac{s}{2}+\frac{|t|}{2\sqrt{1-s^{2}}})^{2} \leq 1.
\end{align*}
According (\ref{A15}) and (\ref{A16}), we have 
\begin{align}\label{A17}
    \rho(F) \leq \sqrt{\mathcal{S}^{2}+\mathcal{T}^{2}} \leq 1.
\end{align}
It is worth noting that $\sqrt{x+y} \leq \sqrt{x}+\frac{y}{2\sqrt{x}}$ holds equality if and only if $y=0$. Then, the (\ref{A17}) holds equality when $t=0$ and $s=0$. Since $s>0$,we have the strict inequality $\rho(F) \leq 1$ which suggests for the linear convergence of unit time $\Delta_{t}$.

\textbf{Case 2.} we suppose $\frac{1}{\sqrt{2}}<s \leq 1$, since $t \leq \frac{s^{2}}{10} \leq 0.1$. Combining (\ref{A18}) and (\ref{A19}) directly, we can obtain 
\begin{align}
    \rho(F) \leq \sqrt{\mathcal{S}^{2}+\mathcal{T}^{2}} < 1.
\end{align}
which is also linear convergence
\end{proof}

\section{The Appendix Figures of Experiments}
\begin{figure}[H]
    \centering
    \subfigure[A mixture of 5 Gaussians]{
    \begin{minipage}[c]{0.40\textwidth}
    \includegraphics[width=1\textwidth]{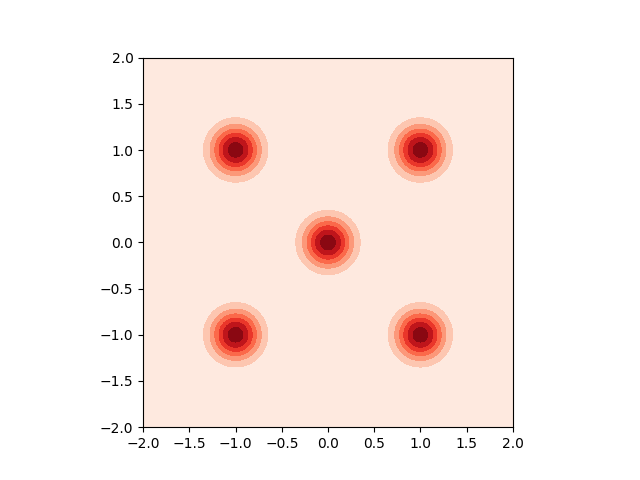}%
    \end{minipage}}%
    \subfigure[A mixture of 16 Gaussians]{
    \begin{minipage}[c]{0.40\textwidth}
    \includegraphics[width=1\textwidth]{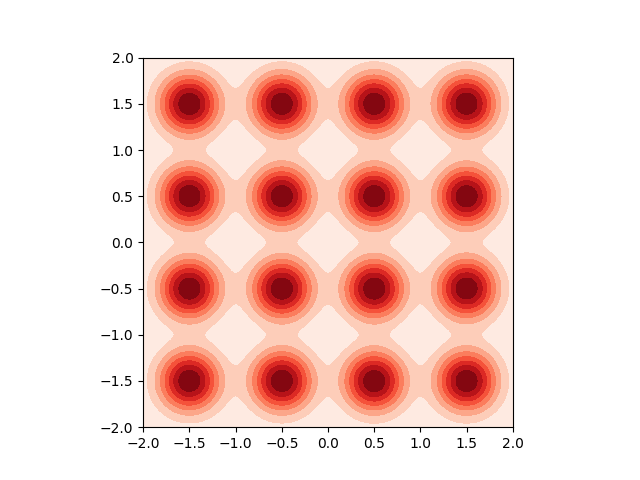}%
    \end{minipage}}%
    \caption{ Ground truth of  Mixture Gaussians}
    \label{fig5}
\end{figure}

\begin{figure}[H]\tiny
(s=50)
\subfigure{
\begin{minipage}[c]{0.95\textwidth}
\includegraphics[width=0.19\textwidth]{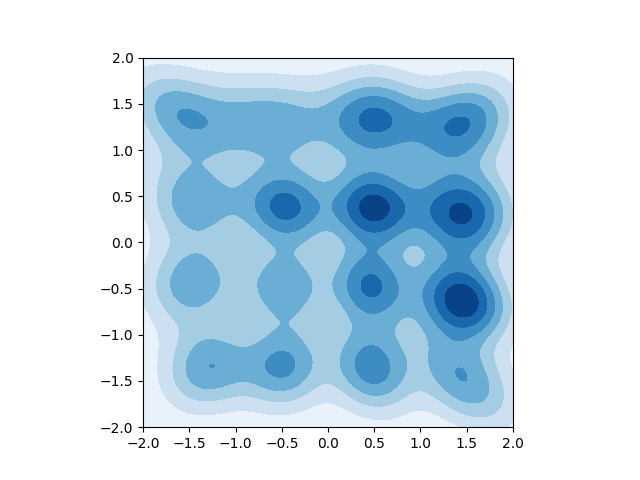}%
\includegraphics[width=0.19\textwidth]{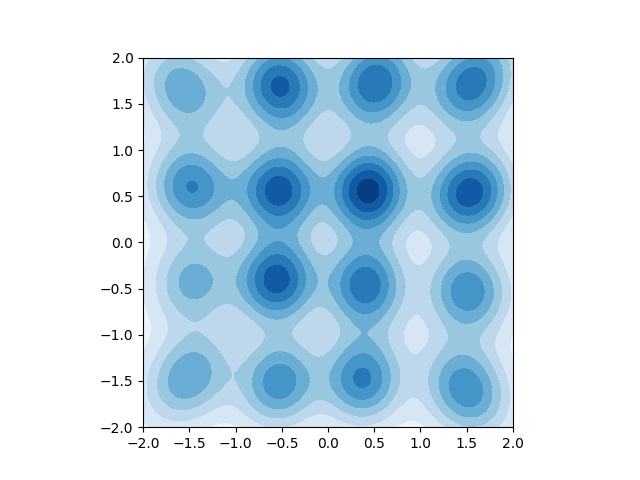}%
\includegraphics[width=0.19\textwidth]{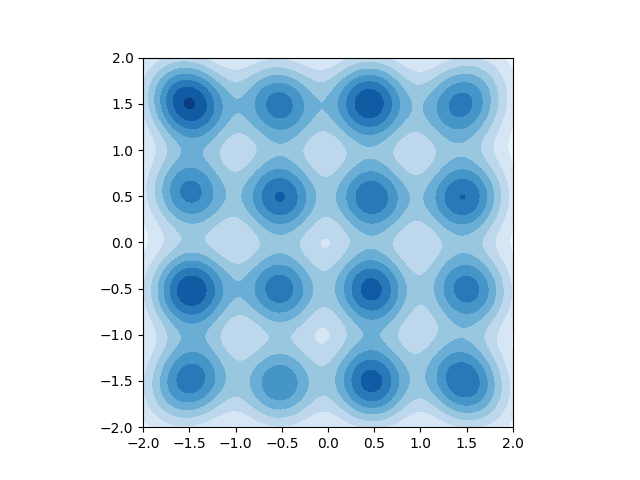}%
\includegraphics[width=0.19\textwidth]{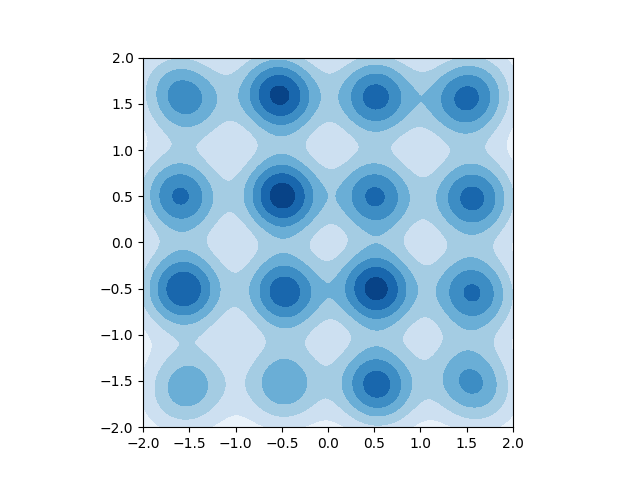}%
\includegraphics[width=0.19\textwidth]{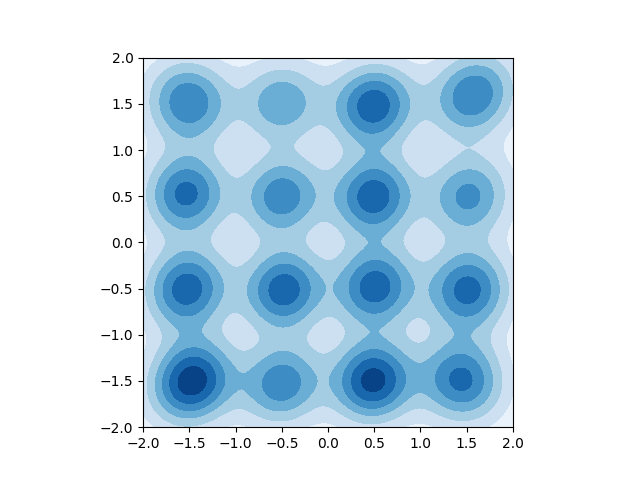}%
\end{minipage}}\\
(s=100)
\subfigure{
\begin{minipage}[c]{0.95\textwidth}
\includegraphics[width=0.19\textwidth]{ACG16_2000_5.png}%
\includegraphics[width=0.19\textwidth]{ACG16_4000_5.png}%
\includegraphics[width=0.19\textwidth]{ACG16_6000_5.png}%
\includegraphics[width=0.19\textwidth]{ACG16_8000_5.png}%
\includegraphics[width=0.19\textwidth]{ACG16_10000_5.png}%
\end{minipage}}\\
(s=150)
\subfigure{
\begin{minipage}[c]{0.95\textwidth}
\includegraphics[width=0.19\textwidth]{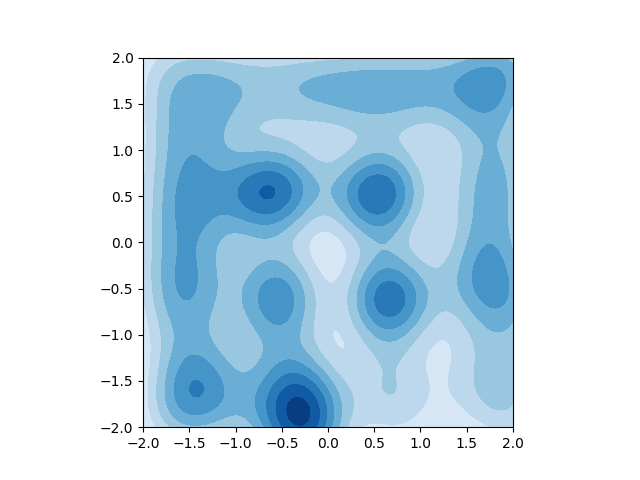}%
\includegraphics[width=0.19\textwidth]{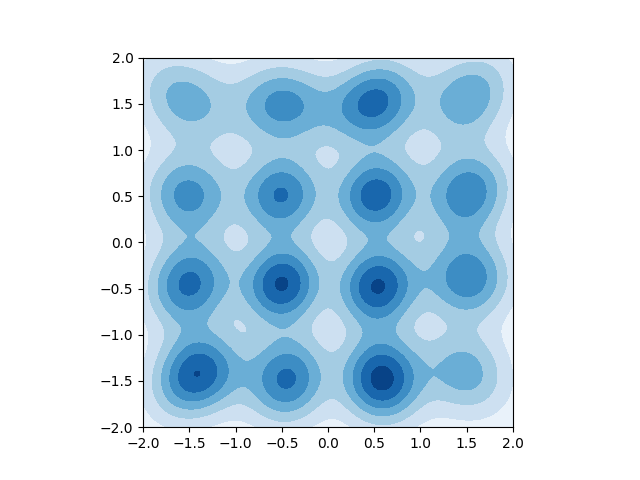}%
\includegraphics[width=0.19\textwidth]{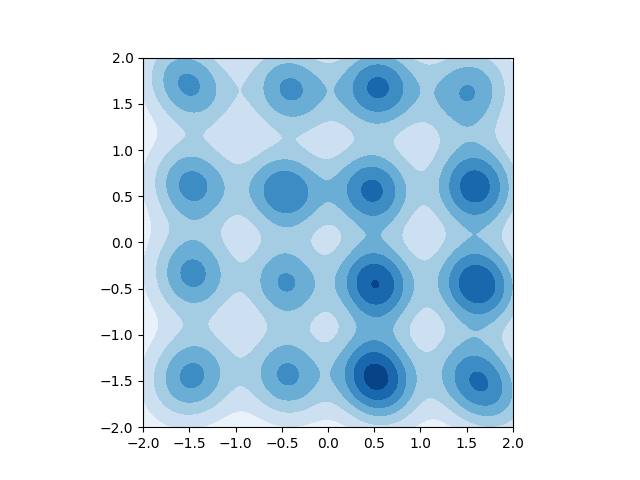}%
\includegraphics[width=0.19\textwidth]{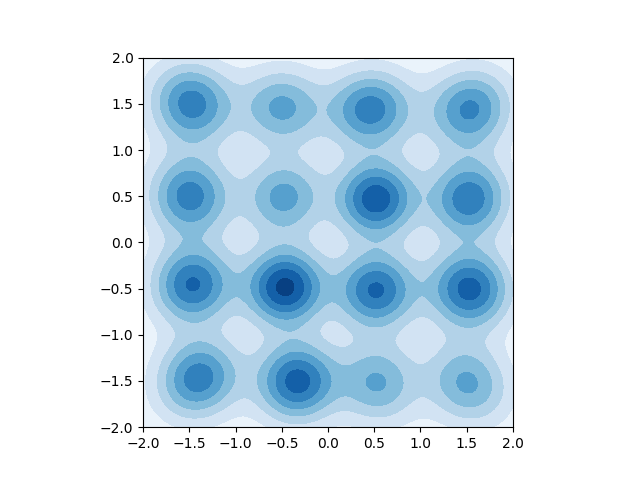}%
\includegraphics[width=0.19\textwidth]{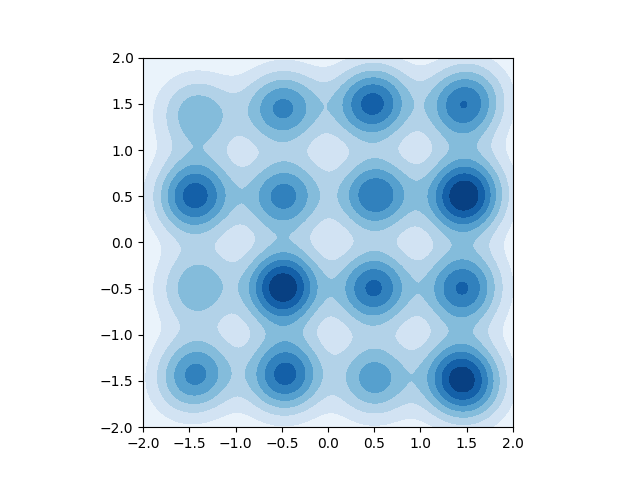}%
\end{minipage}}\\
(s=200)
\subfigure{
\begin{minipage}[c]{0.179\textwidth}
\includegraphics[width=1\textwidth]{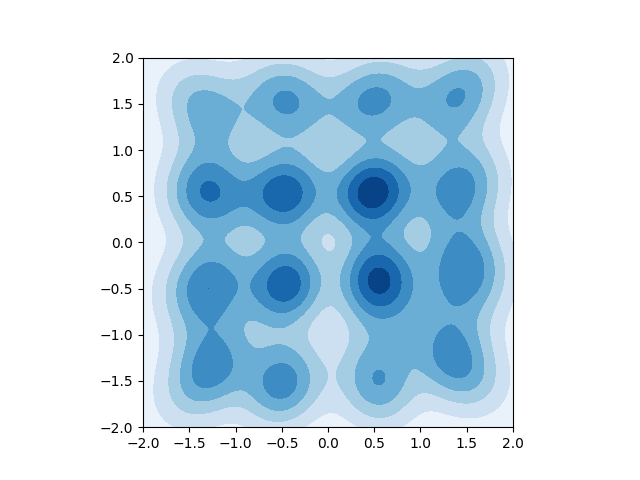}
\caption*{2000}
\end{minipage}}%
\subfigure{
\begin{minipage}[c]{0.179\textwidth}
\includegraphics[width=1\textwidth]{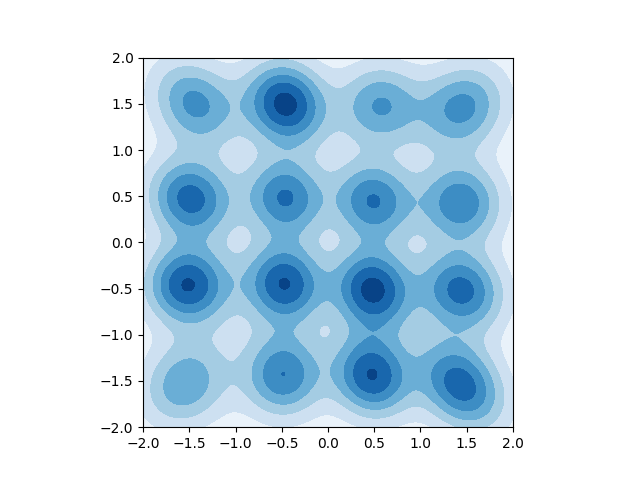}
\caption*{4000}
\end{minipage}}%
\subfigure{
\begin{minipage}[c]{0.179\textwidth}
\includegraphics[width=1\textwidth]{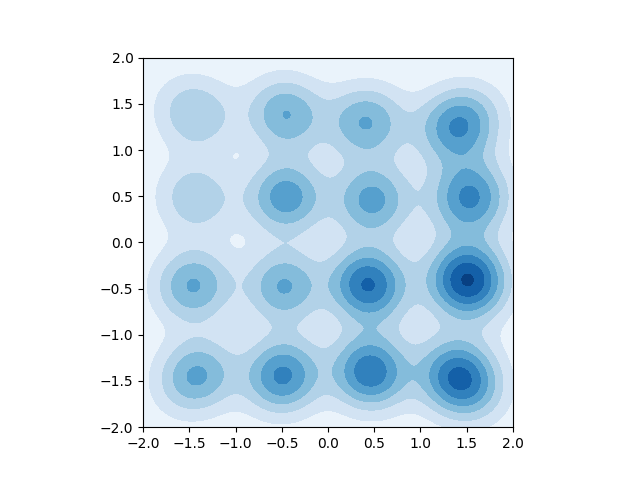}
\caption*{6000}
\end{minipage}}%
\subfigure{
\begin{minipage}[c]{0.179\textwidth}
\includegraphics[width=1\textwidth]{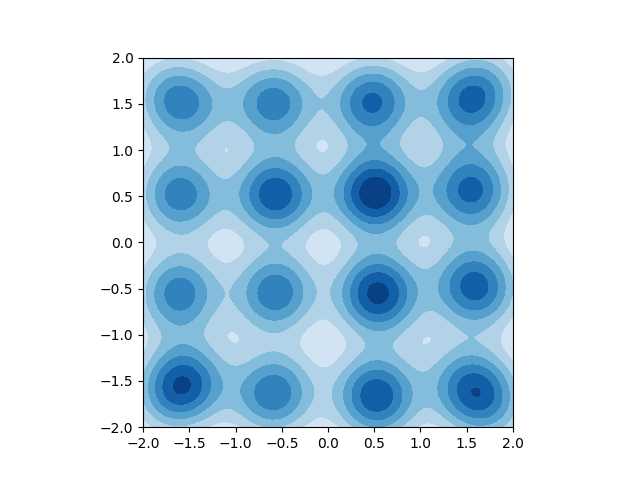}
\caption*{8000}
\end{minipage}}%
\subfigure{
\begin{minipage}[c]{0.179\textwidth}
\includegraphics[width=1\textwidth]{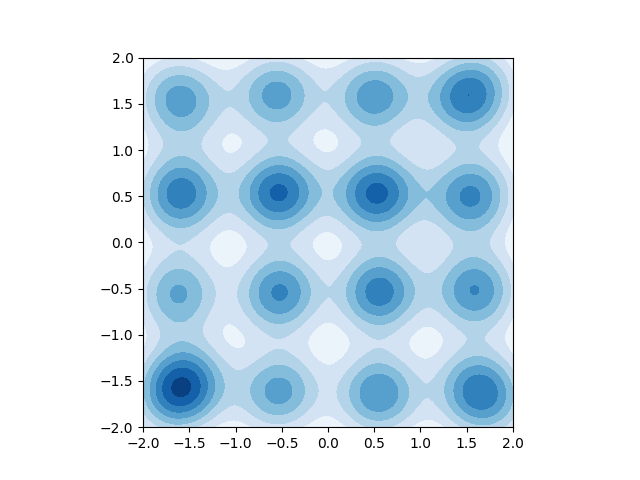}
\caption*{10000}
\end{minipage}}%
\caption{Exploring of $s$, the mixture of 16 Gaussians. Each row shows results with $s$ at different values, and then each column shows the results with iteration number through $\{2000,4000,6000,8000,10000\}$. This figure present that the method converges faster with $s$ increasing.}
\label{fig8}
\end{figure}

\begin{figure}[H]\tiny
RMSP \quad \qquad \;
\subfigure{
\begin{minipage}[c]{0.95\textwidth}
\includegraphics[width=0.19\textwidth]{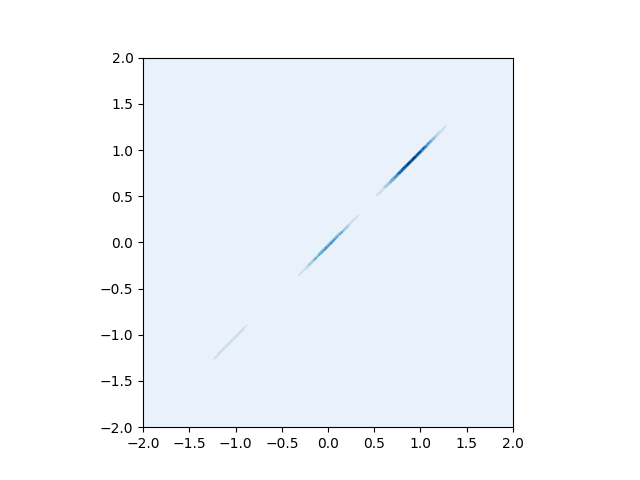}%
\includegraphics[width=0.19\textwidth]{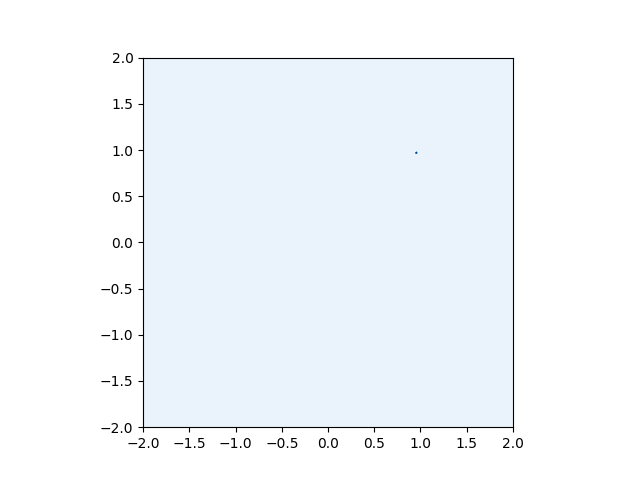}%
\includegraphics[width=0.19\textwidth]{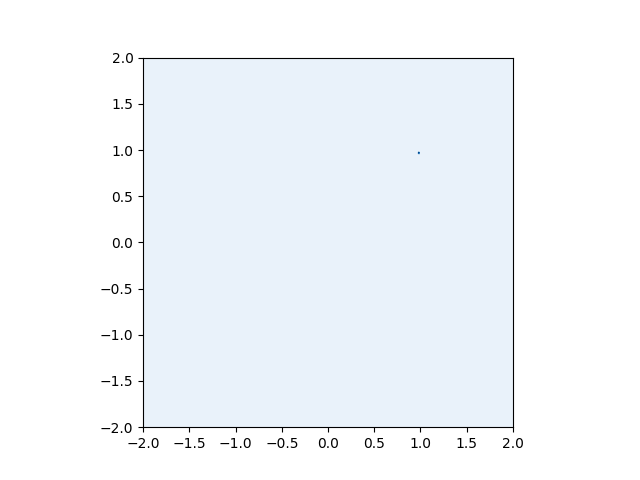}%
\includegraphics[width=0.19\textwidth]{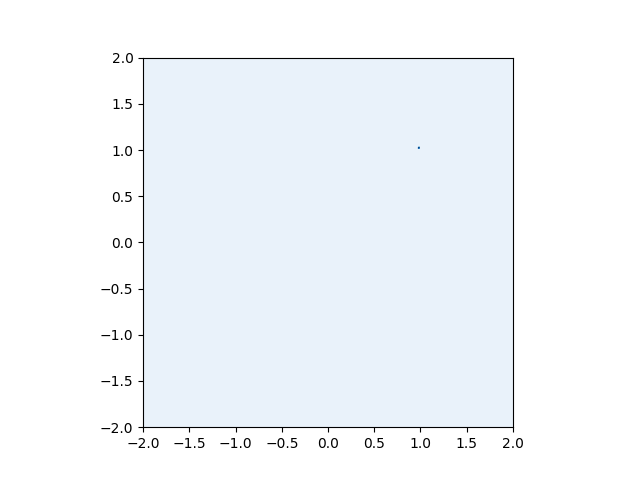}%
\includegraphics[width=0.19\textwidth]{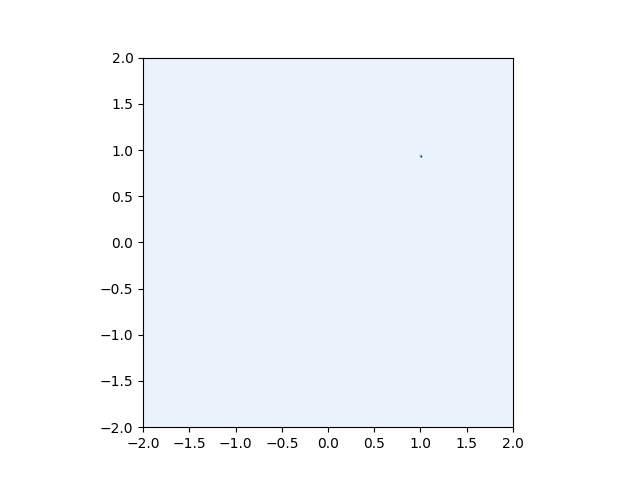}%
\end{minipage}}\\
RMSP-alt \quad \quad \;
\subfigure{
\begin{minipage}[c]{0.95\textwidth}
\includegraphics[width=0.19\textwidth]{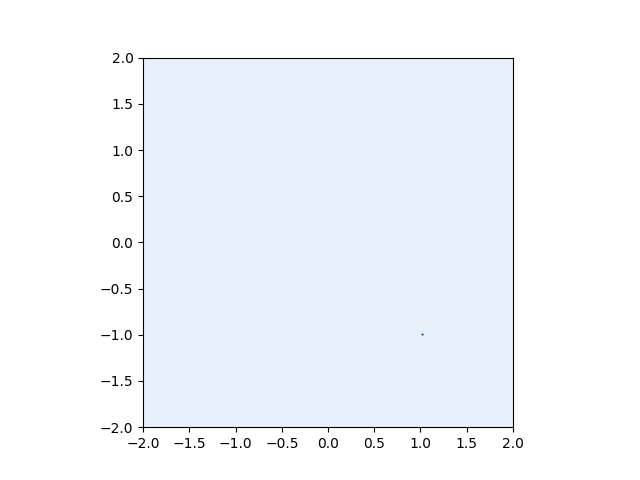}%
\includegraphics[width=0.19\textwidth]{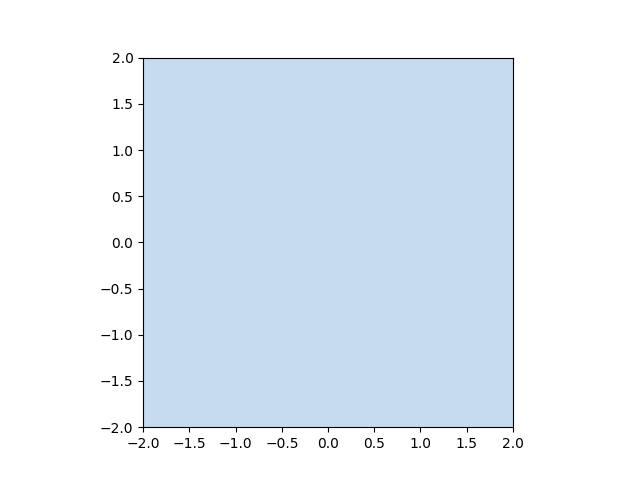}%
\includegraphics[width=0.19\textwidth]{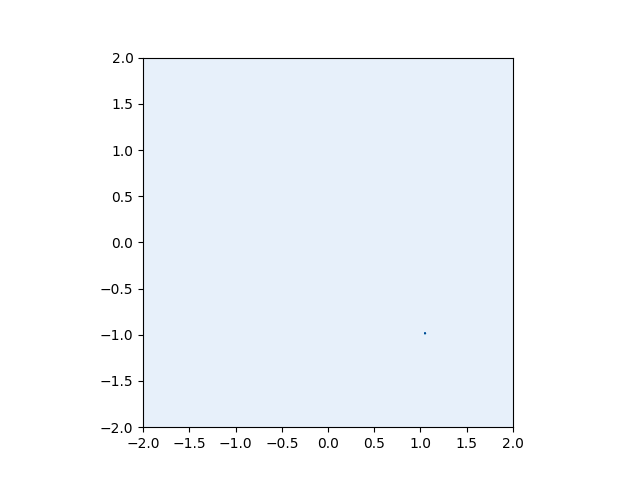}%
\includegraphics[width=0.19\textwidth]{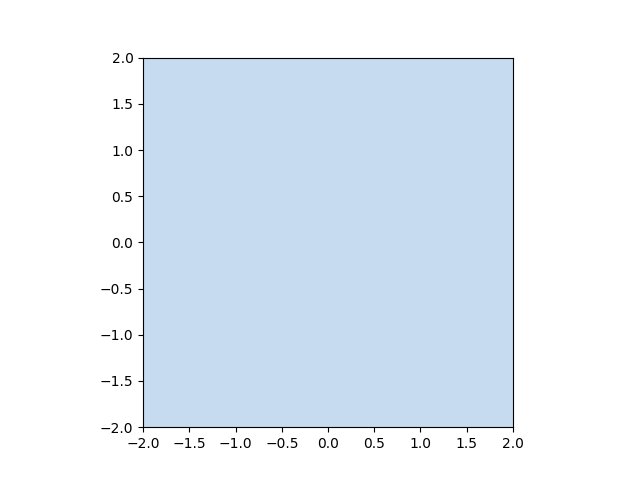}%
\includegraphics[width=0.19\textwidth]{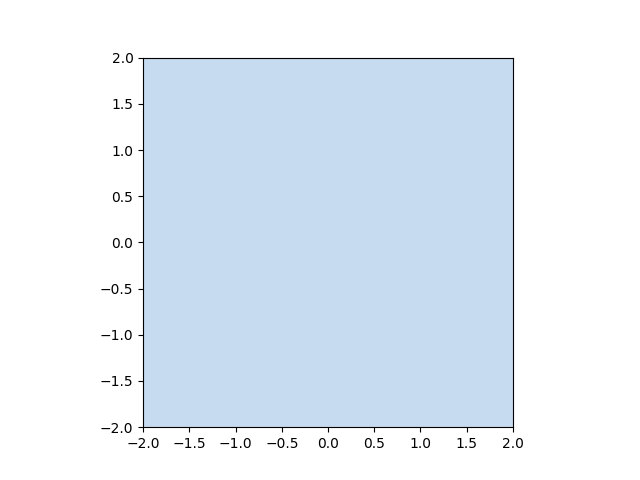}%
\end{minipage}}\\
ConOpt \qquad \quad \;
\subfigure{
\begin{minipage}[c]{0.95\textwidth}
\includegraphics[width=0.19\textwidth]{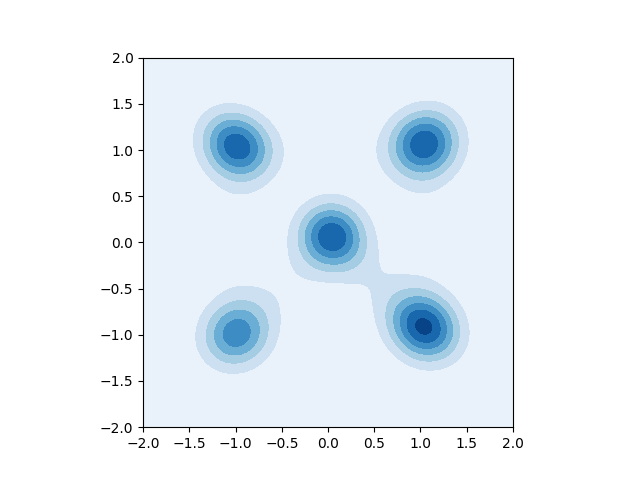}%
\includegraphics[width=0.19\textwidth]{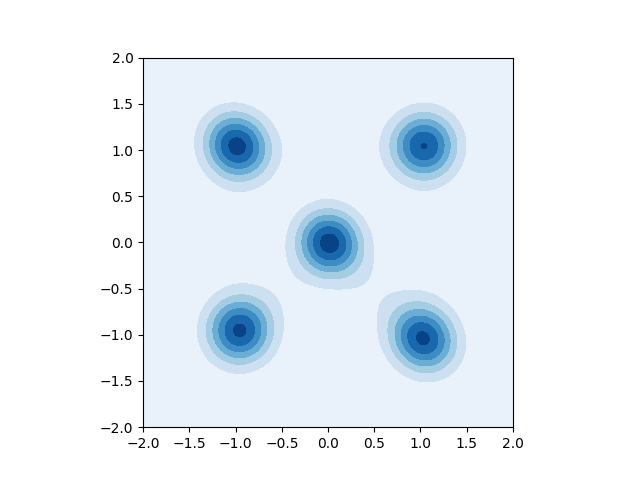}%
\includegraphics[width=0.19\textwidth]{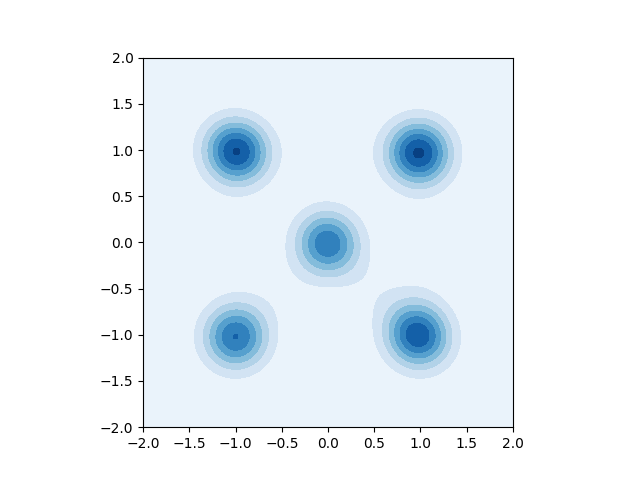}%
\includegraphics[width=0.19\textwidth]{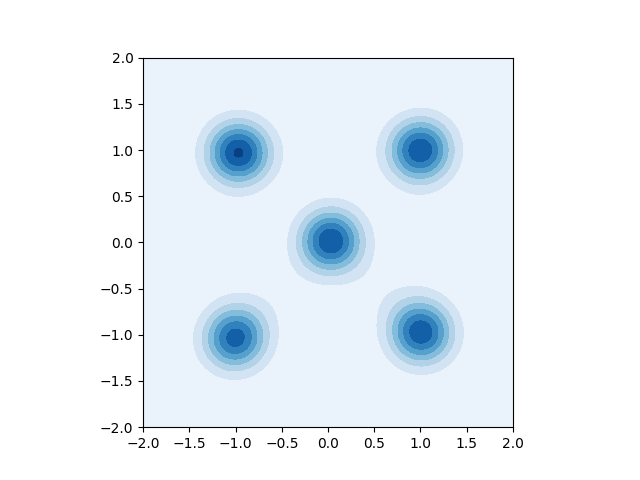}%
\includegraphics[width=0.19\textwidth]{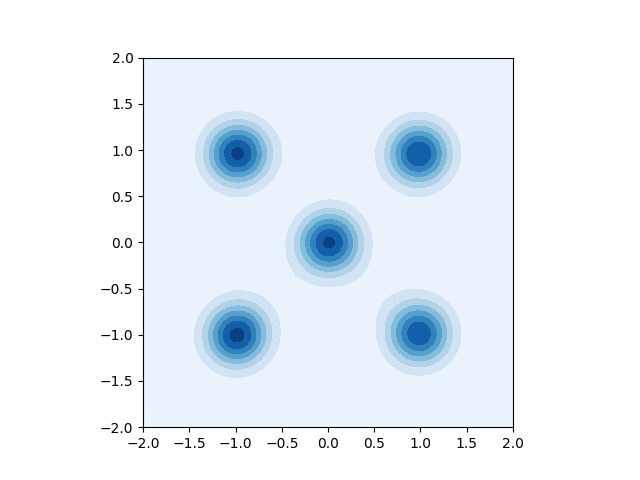}%
\end{minipage}}\\
RMSP-SGA \qquad
\subfigure{
\begin{minipage}[c]{0.95\textwidth}
\includegraphics[width=0.19\textwidth]{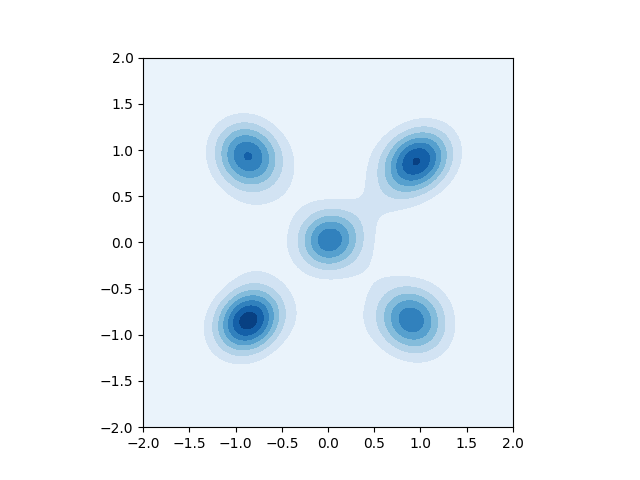}%
\includegraphics[width=0.19\textwidth]{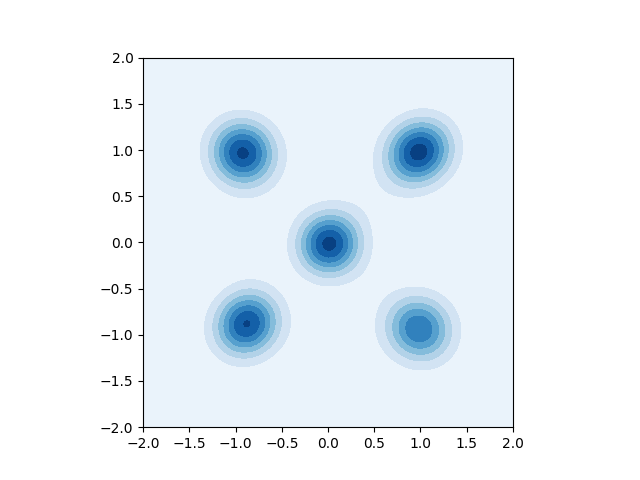}%
\includegraphics[width=0.19\textwidth]{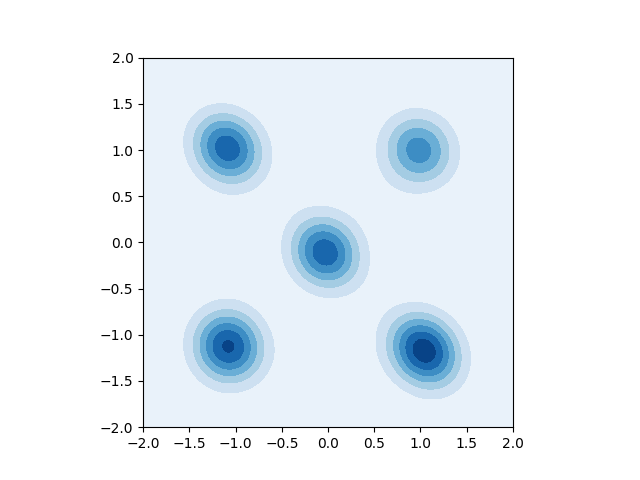}%
\includegraphics[width=0.19\textwidth]{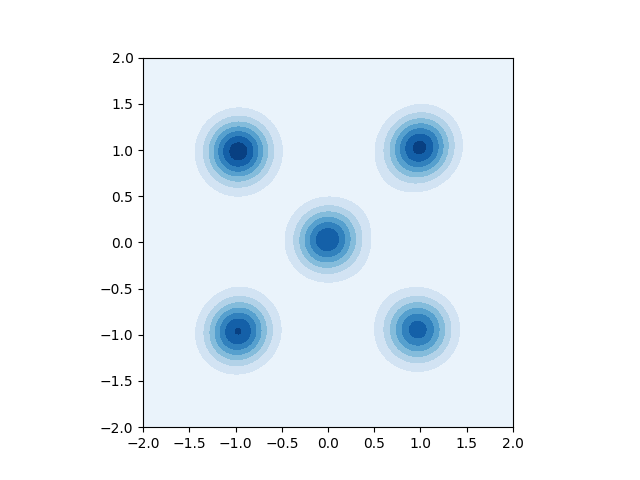}%
\includegraphics[width=0.19\textwidth]{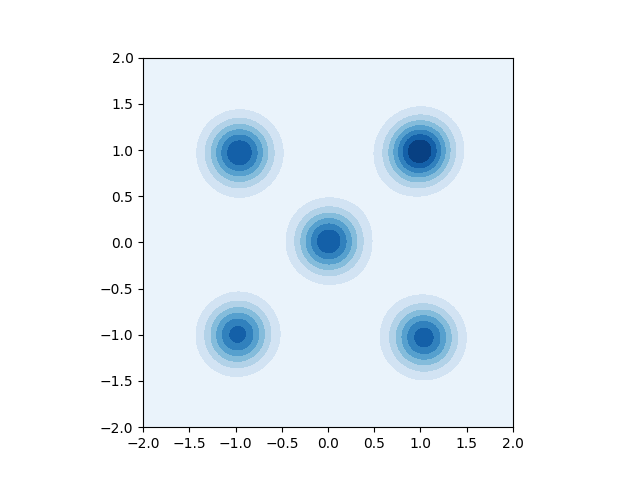}%
\end{minipage}}\\
RMSP-ACA \quad \quad 
\subfigure{
\begin{minipage}[c]{0.95\textwidth}
\includegraphics[width=0.19\textwidth]{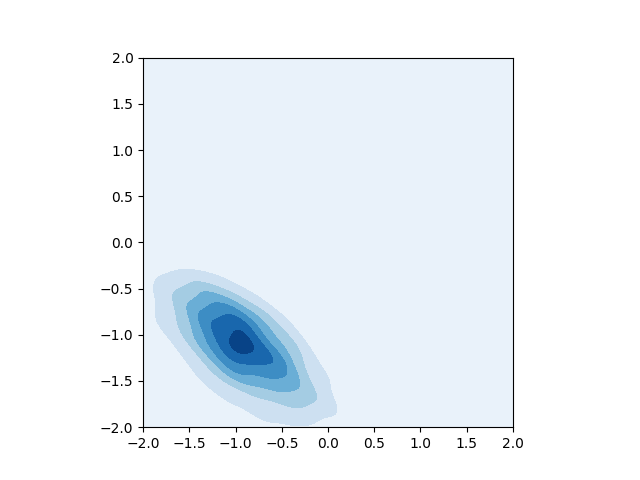}%
\includegraphics[width=0.19\textwidth]{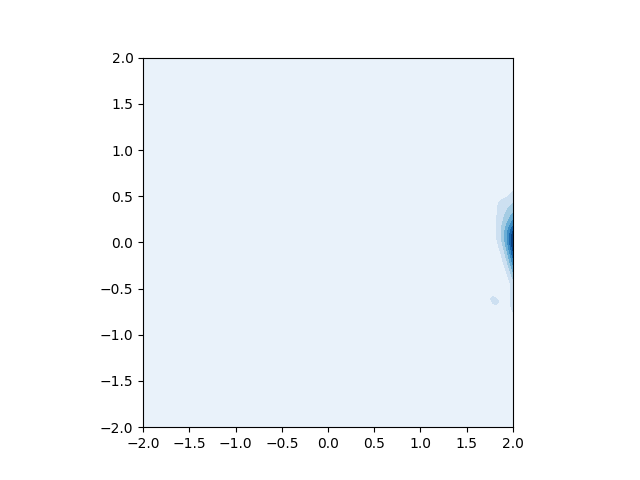}%
\includegraphics[width=0.19\textwidth]{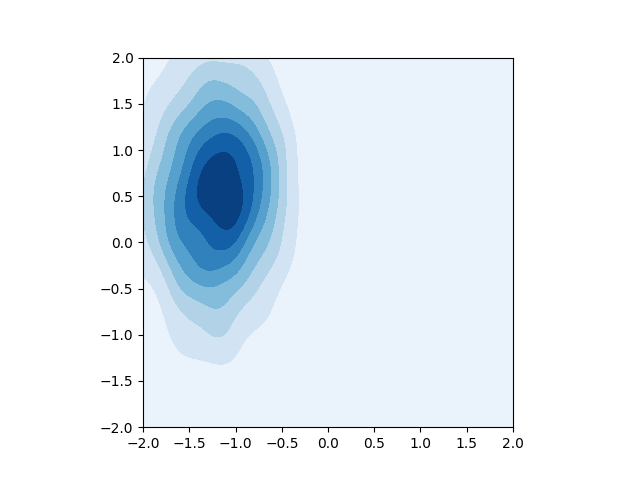}%
\includegraphics[width=0.19\textwidth]{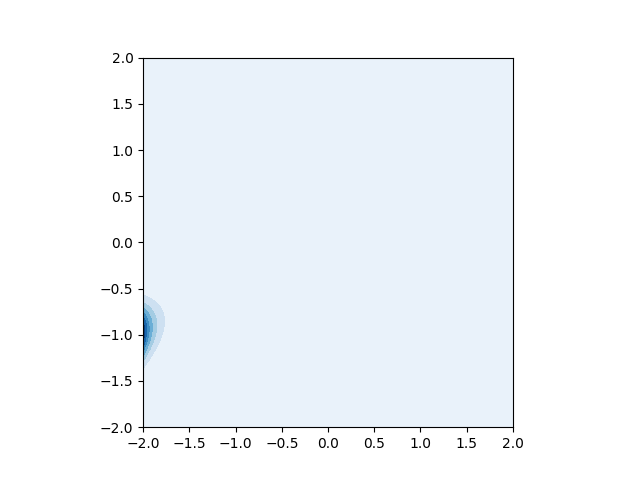}%
\includegraphics[width=0.19\textwidth]{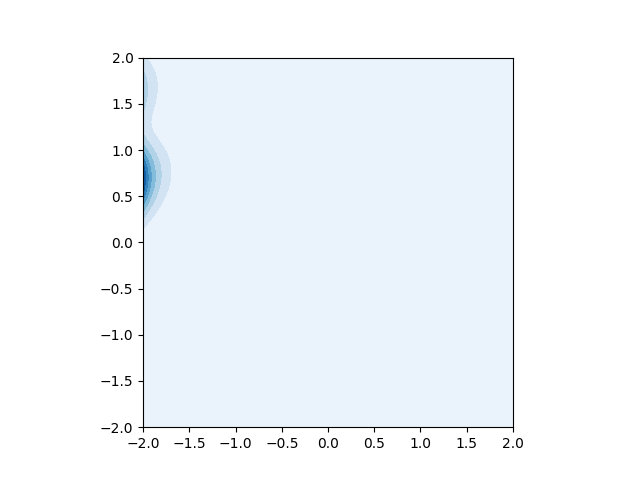}%
\end{minipage}}\\
SGA-ACG(Ours)
\subfigure{
\begin{minipage}[c]{0.179\textwidth}
\includegraphics[width=1\textwidth]{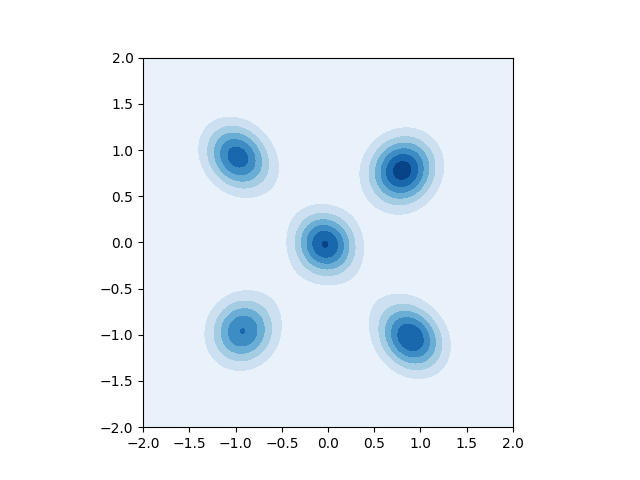}
\caption*{2000}
\end{minipage}}%
\subfigure{
\begin{minipage}[c]{0.179\textwidth}
\includegraphics[width=1\textwidth]{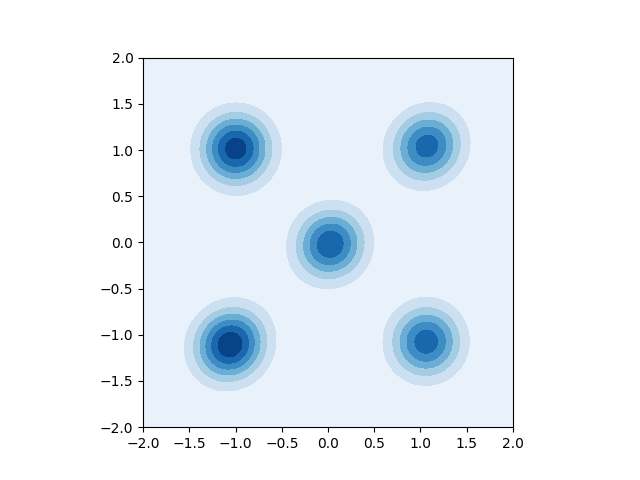}
\caption*{4000}
\end{minipage}}%
\subfigure{
\begin{minipage}[c]{0.179\textwidth}
\includegraphics[width=1\textwidth]{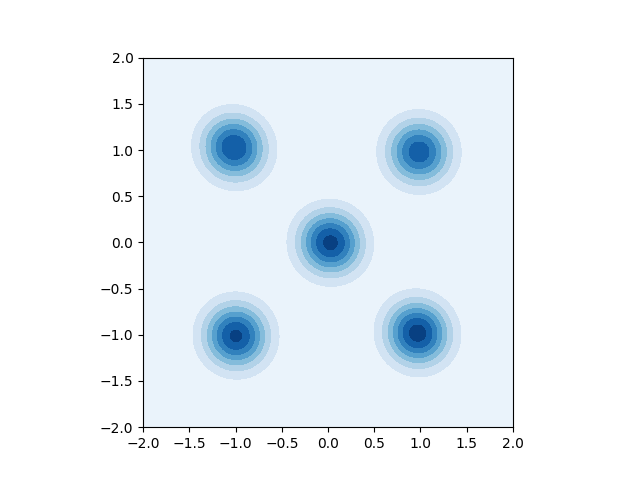}
\caption*{6000}
\end{minipage}}%
\subfigure{
\begin{minipage}[c]{0.179\textwidth}
\includegraphics[width=1\textwidth]{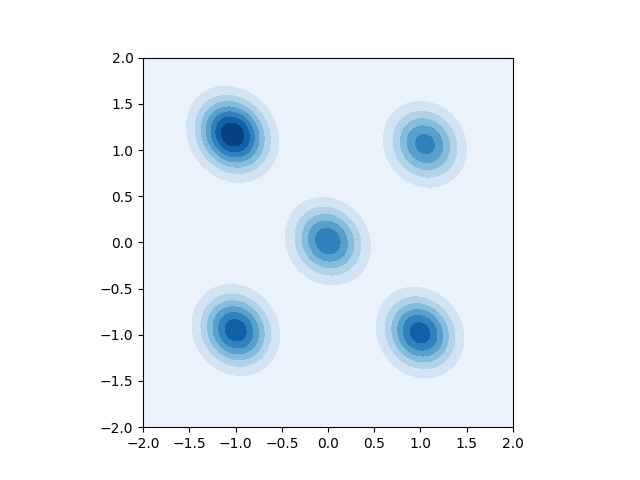}
\caption*{8000}
\end{minipage}}%
\subfigure{
\begin{minipage}[c]{0.179\textwidth}
\includegraphics[width=1\textwidth]{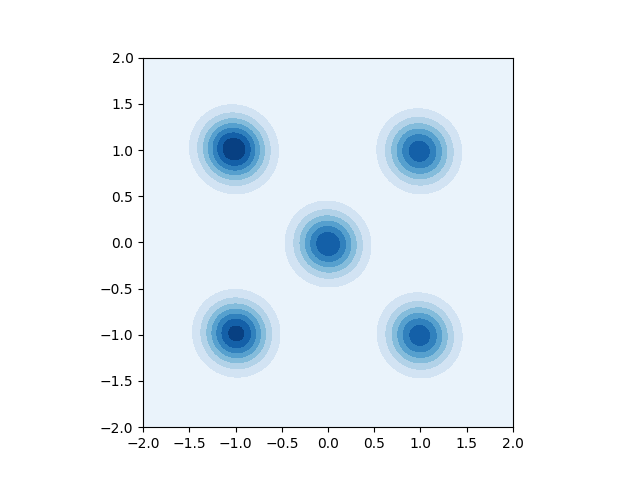}
\caption*{10000}
\end{minipage}}%
\caption{Compared results on the mixture of 5 Gaussians. Each row suggests a different method, and each column is the results at other iteration numbers through $\{2000,4000,6000,8000,10000\}$. Although from the figure, we can obtain that RMSP, RMSP-alt, RMSP-ACA can not converge to the ground truth, and ConOpt and RMSP-SGA, SGA-ACG all converge to the ground truth, our proposed method seems competitive to ConOpt and RMSP-SGA.} 
\label{fig6}
\end{figure}

\begin{figure}[H]\tiny
(RMSP)\quad \qquad \;
\subfigure{
\begin{minipage}[c]{0.95\textwidth}
\includegraphics[width=0.19\textwidth]{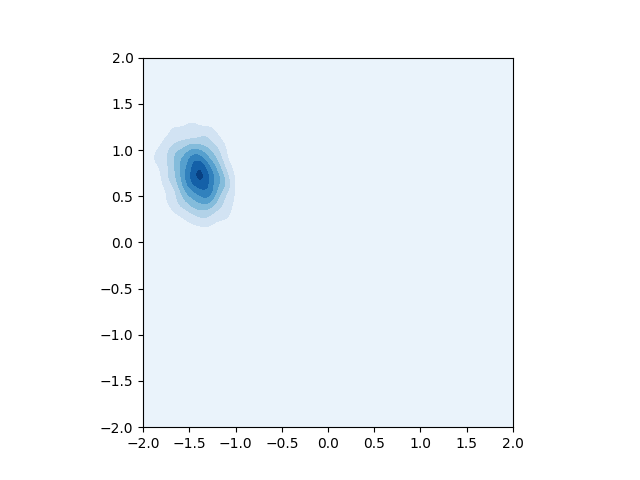}%
\includegraphics[width=0.19\textwidth]{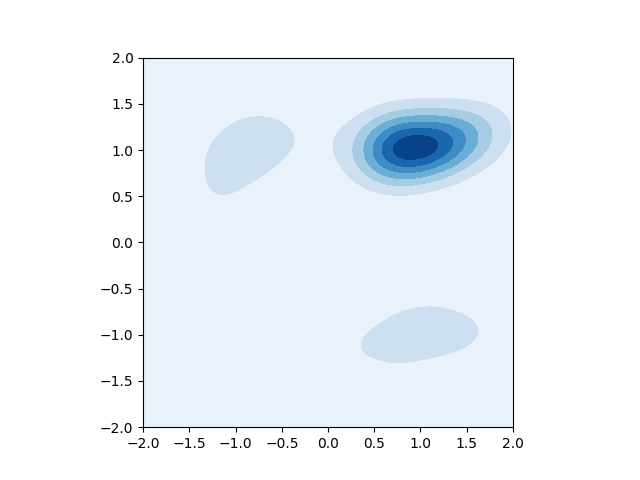}%
\includegraphics[width=0.19\textwidth]{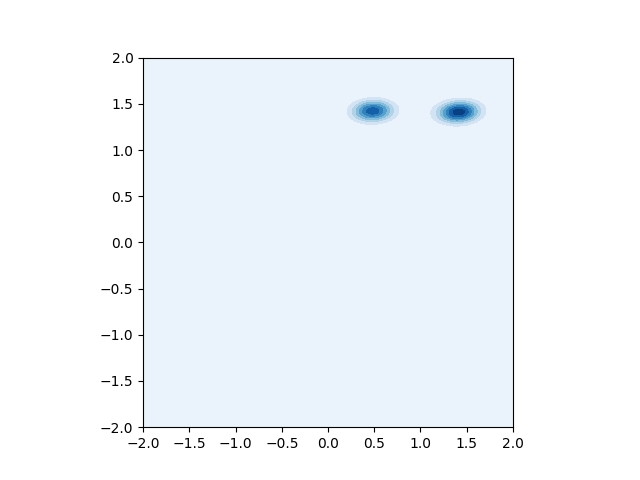}%
\includegraphics[width=0.19\textwidth]{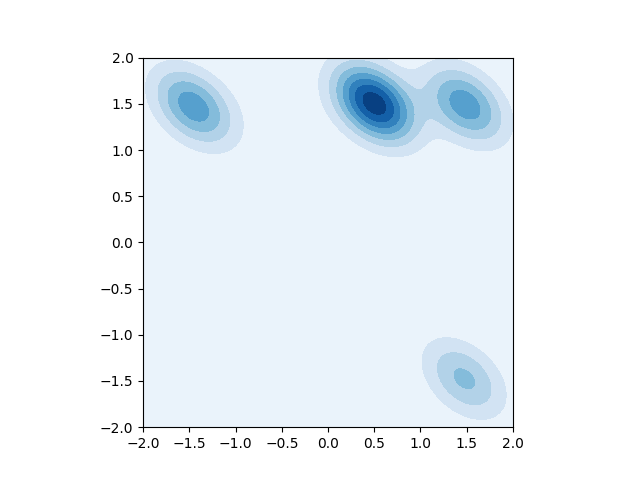}%
\includegraphics[width=0.19\textwidth]{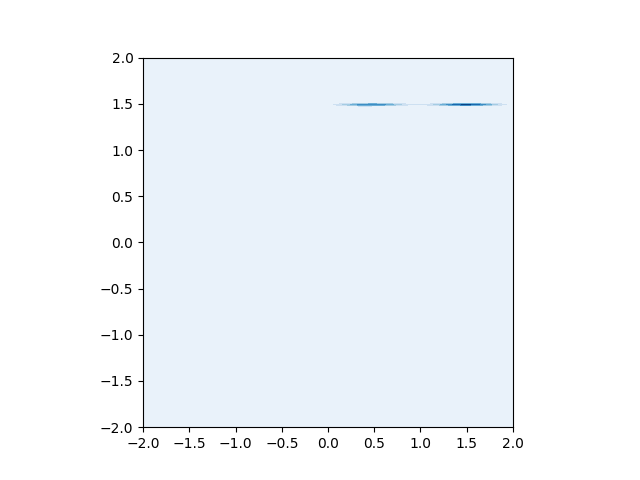}%
\end{minipage}}\\
(RMSP-alt) \quad \quad \;
\subfigure{
\begin{minipage}[c]{0.95\textwidth}
\includegraphics[width=0.19\textwidth]{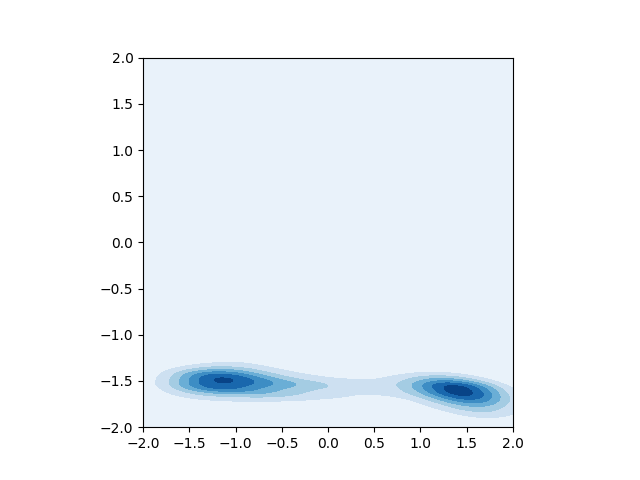}%
\includegraphics[width=0.19\textwidth]{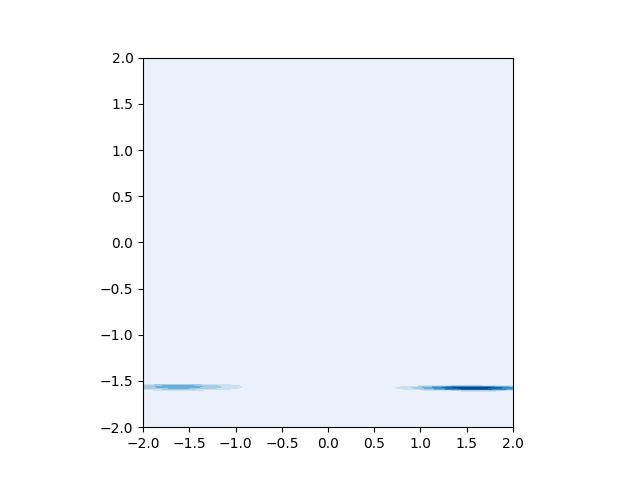}%
\includegraphics[width=0.19\textwidth]{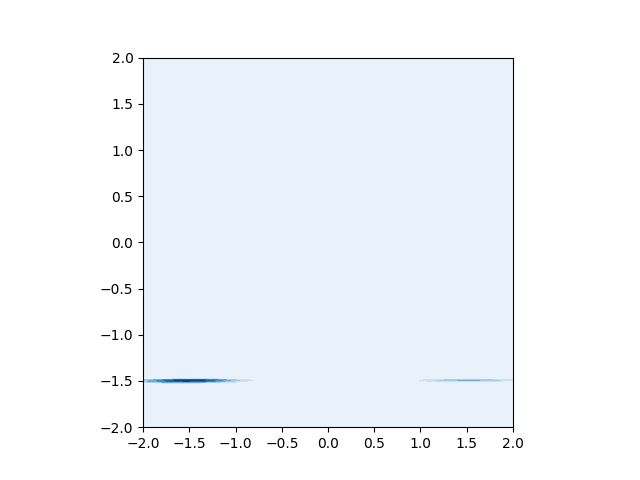}%
\includegraphics[width=0.19\textwidth]{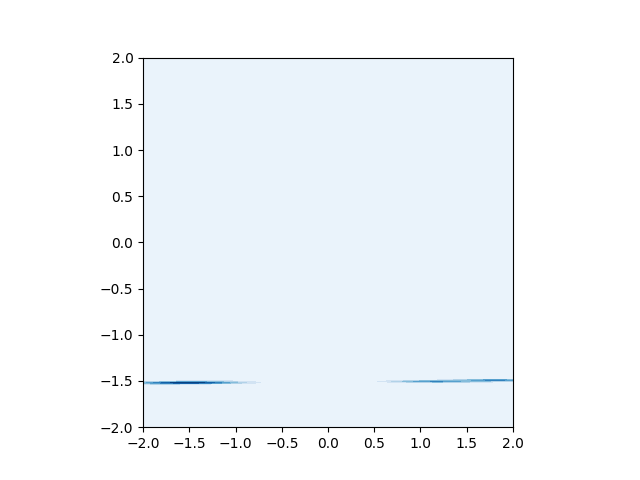}%
\includegraphics[width=0.19\textwidth]{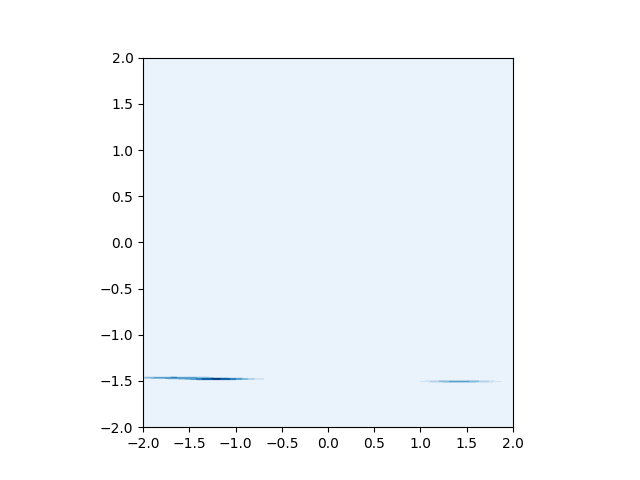}%
\end{minipage}}\\
(ConOpt)\qquad \quad \;
\subfigure{
\begin{minipage}[c]{0.95\textwidth}
\includegraphics[width=0.19\textwidth]{m16_ConOpt_6_256_it_2000.png}%
\includegraphics[width=0.19\textwidth]{m16_ConOpt_6_256_it_4000.png}%
\includegraphics[width=0.19\textwidth]{m16_ConOpt_6_256_it_6000.png}%
\includegraphics[width=0.19\textwidth]{m16_ConOpt_6_256_it_8000.png}%
\includegraphics[width=0.19\textwidth]{m16_ConOpt_6_256_it_10000.png}%
\end{minipage}}\\
RMSP-SGA \qquad
\subfigure{
\begin{minipage}[c]{0.95\textwidth}
\includegraphics[width=0.19\textwidth]{m16_SGA_6_256_it_2000.png}%
\includegraphics[width=0.19\textwidth]{m16_SGA_6_256_it_4000.png}%
\includegraphics[width=0.19\textwidth]{m16_SGA_6_256_it_6000.png}%
\includegraphics[width=0.19\textwidth]{m16_SGA_6_256_it_8000.png}%
\includegraphics[width=0.19\textwidth]{m16_SGA_6_256_it_10000.png}%
\end{minipage}}\\
RMSP-ACA \quad \quad 
\subfigure{
\begin{minipage}[c]{0.95\textwidth}
\includegraphics[width=0.19\textwidth]{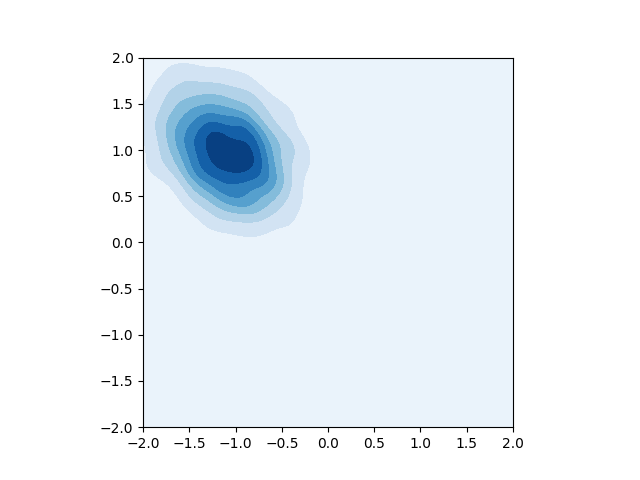}%
\includegraphics[width=0.19\textwidth]{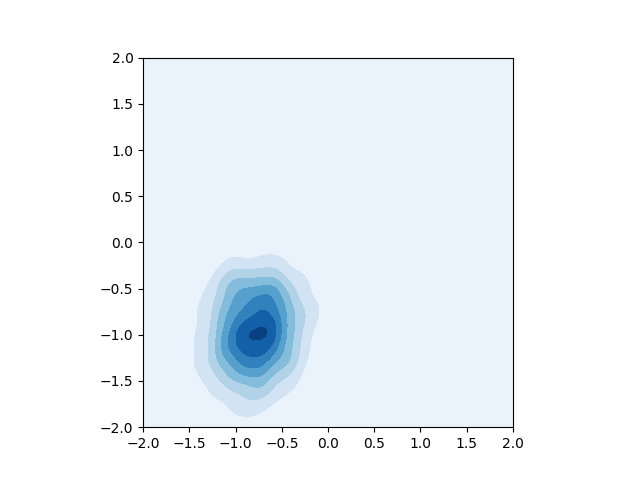}%
\includegraphics[width=0.19\textwidth]{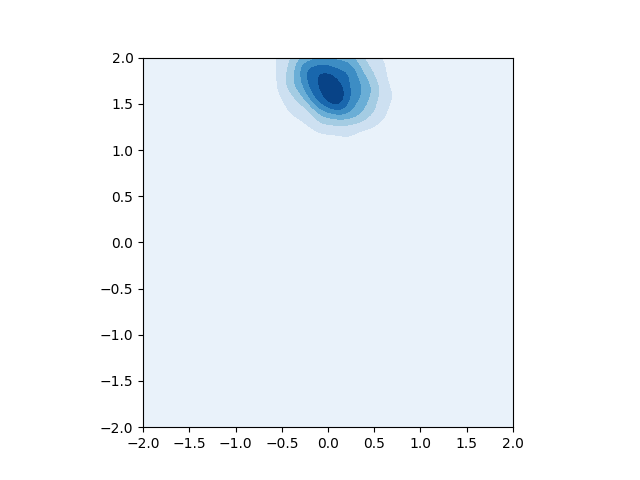}%
\includegraphics[width=0.19\textwidth]{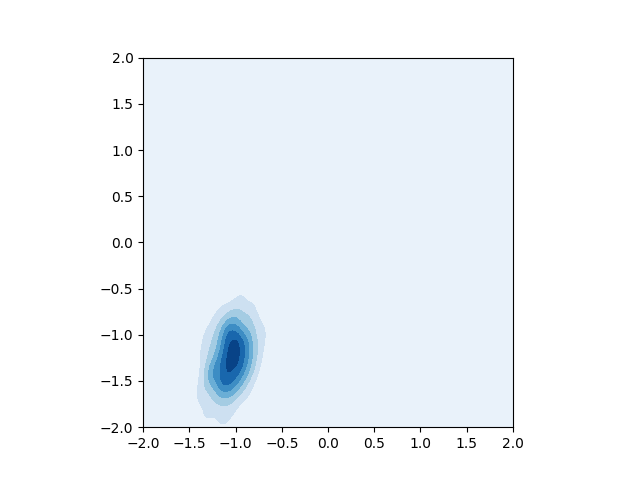}%
\includegraphics[width=0.19\textwidth]{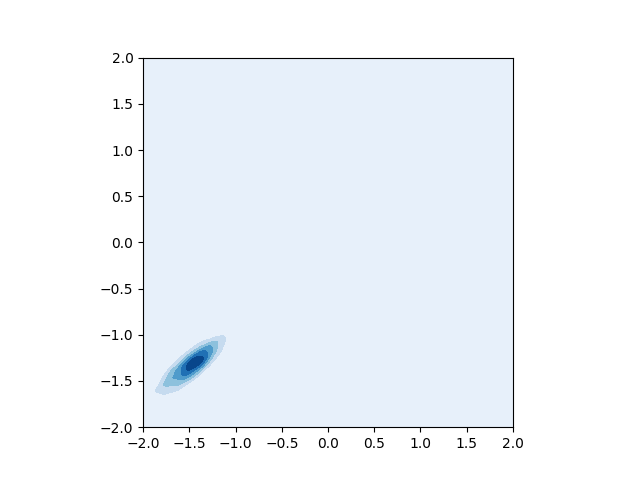}%
\end{minipage}}\\
SGA-ACG(Ours)
\subfigure{
\begin{minipage}[c]{0.179\textwidth}
\includegraphics[width=1\textwidth]{ACG16_2000_5.png}
\caption*{2000}
\end{minipage}}%
\subfigure{
\begin{minipage}[c]{0.179\textwidth}
\includegraphics[width=1\textwidth]{ACG16_4000_5.png}
\caption*{4000}
\end{minipage}}%
\subfigure{
\begin{minipage}[c]{0.179\textwidth}
\includegraphics[width=1\textwidth]{ACG16_6000_5.png}
\caption*{6000}
\end{minipage}}%
\subfigure{
\begin{minipage}[c]{0.179\textwidth}
\includegraphics[width=1\textwidth]{ACG16_8000_5.png}
\caption*{8000}
\end{minipage}}%
\subfigure{
\begin{minipage}[c]{0.179\textwidth}
\includegraphics[width=1\textwidth]{ACG16_10000_5.png}
\caption*{10000}
\end{minipage}}%
\caption{Compared results on the mixture of 16 Gaussians. Each row suggests a different method, and then each column is the results at different iteration numbers through $\{2000,4000,6000,8000,10000\}$. The figure shows that RMSP, RMSP-alt, RMSP-ACA can not converge to the ground truth, and ConOpt and RMSP-SGA, SGA-ACG all converge to the ground truth. However, our methods converge faster than ConOpt and RMSP-SGA at iteration 2000, which seems competitive to ConOpt and RMSP-SGA.}
\label{fig7-1}
\end{figure}

\begin{figure}[H]\tiny
SGD \quad \quad \quad \qquad
\subfigure{
\begin{minipage}[c]{0.92\textwidth}
\includegraphics[width=0.23\textwidth]{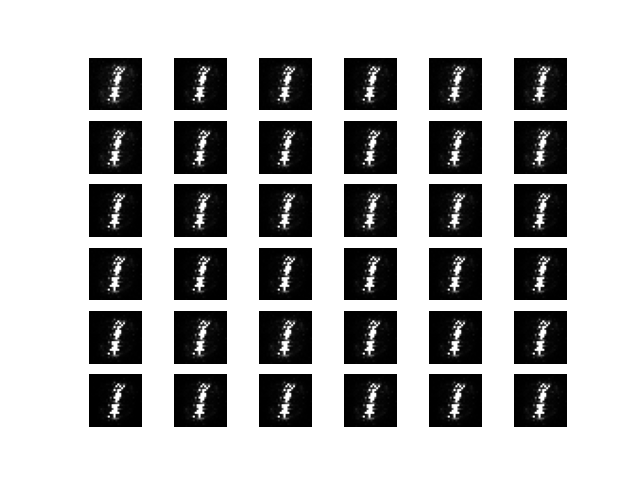}%
\includegraphics[width=0.23\textwidth]{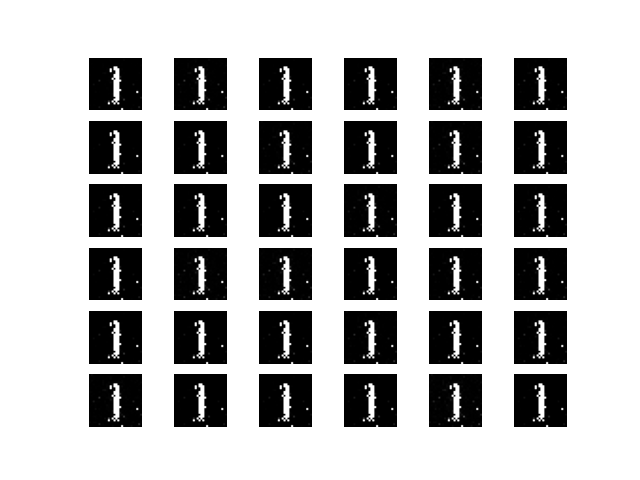}%
\includegraphics[width=0.23\textwidth]{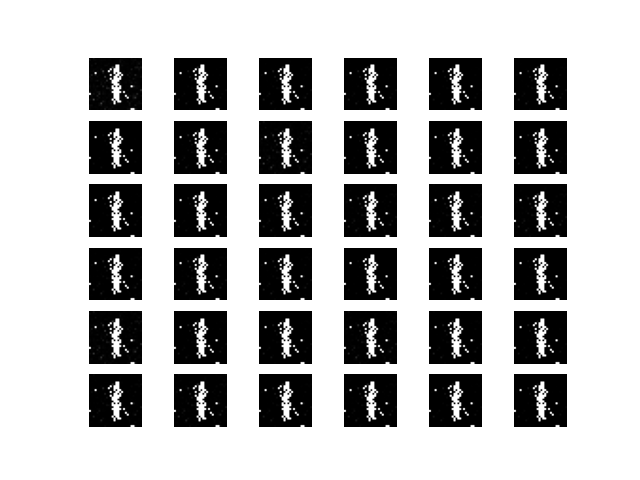}%
\includegraphics[width=0.23\textwidth]{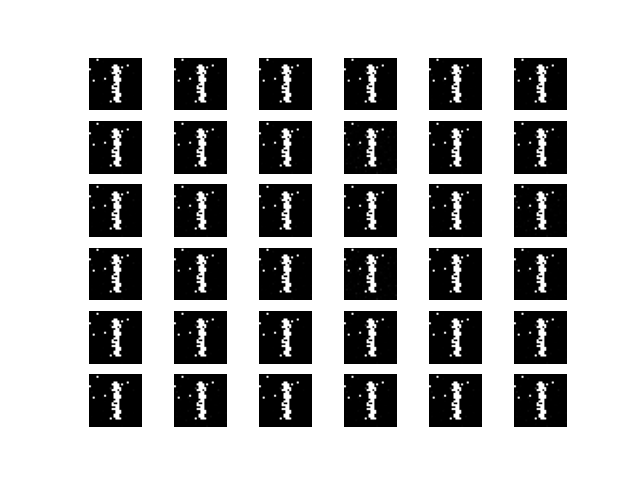}%
\end{minipage}}\\
Adam \quad \quad \quad \qquad
\subfigure{
\begin{minipage}[c]{0.92\textwidth}
\includegraphics[width=0.23\textwidth]{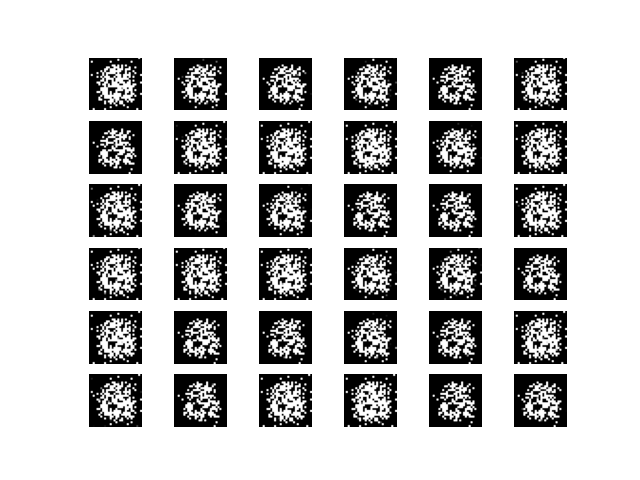}%
\includegraphics[width=0.23\textwidth]{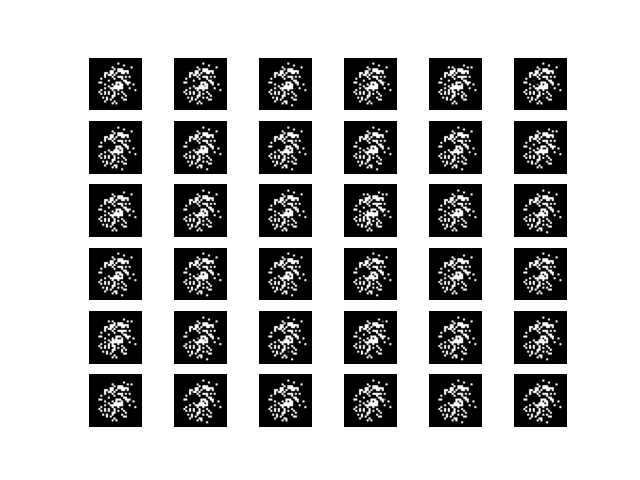}%
\includegraphics[width=0.23\textwidth]{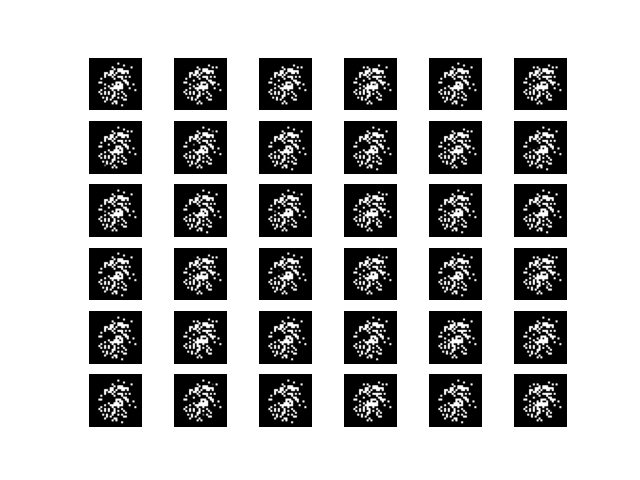}%
\includegraphics[width=0.23\textwidth]{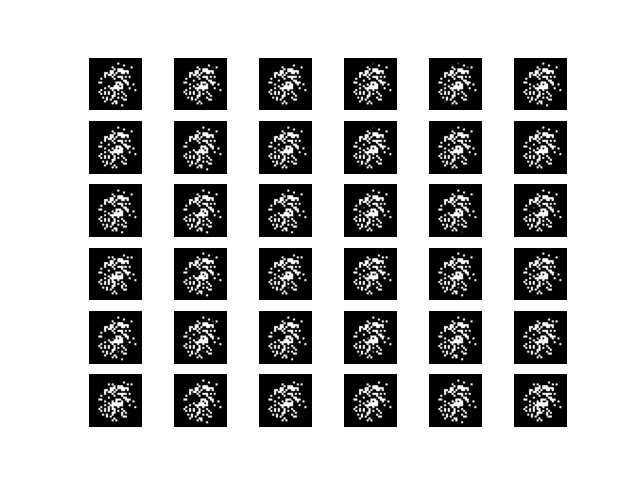}%
\end{minipage}}\\
RMSP \quad \quad \quad \qquad
\subfigure{
\begin{minipage}[c]{0.92\textwidth}
\includegraphics[width=0.23\textwidth]{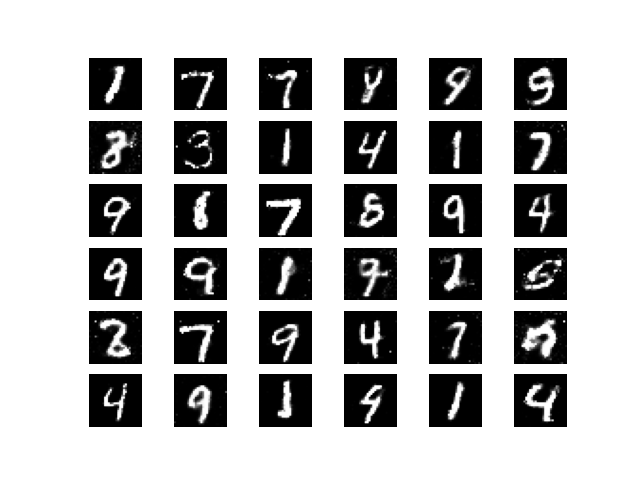}%
\includegraphics[width=0.23\textwidth]{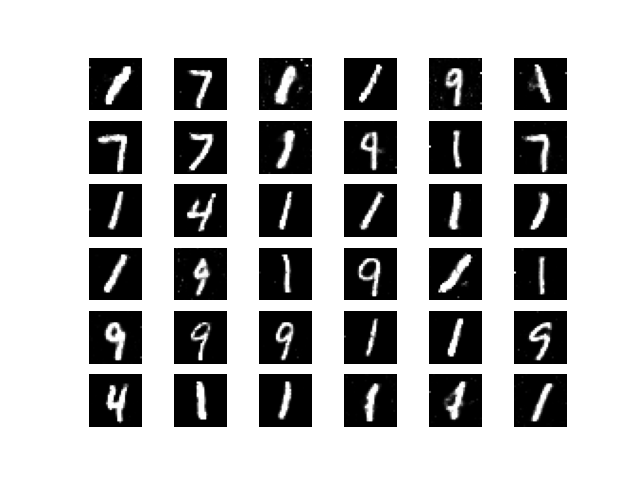}%
\includegraphics[width=0.23\textwidth]{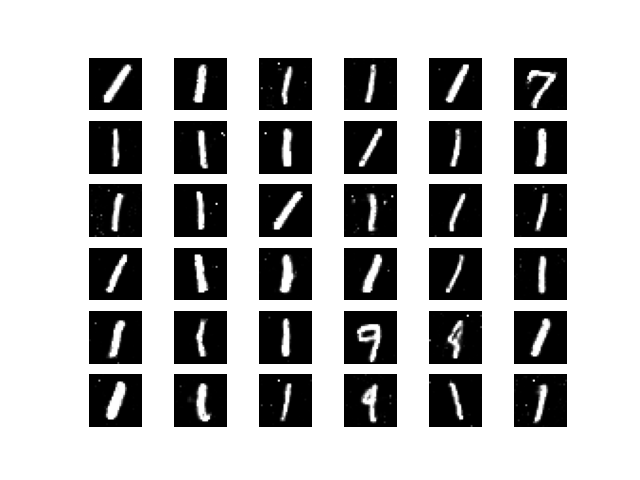}%
\includegraphics[width=0.23\textwidth]{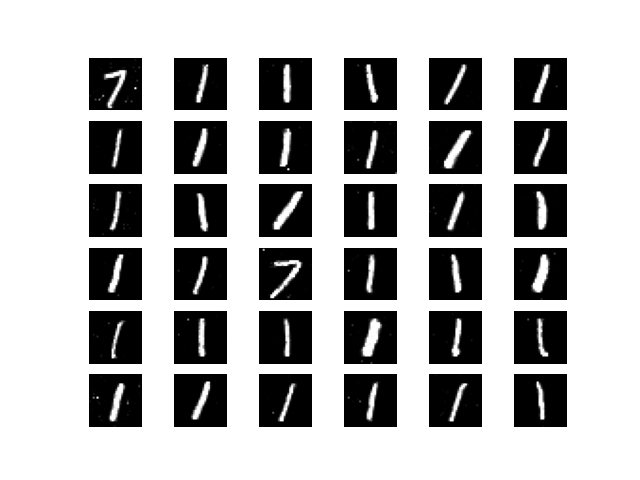}%
\end{minipage}}\\
RMSP-ACG(ours)
\subfigure{
\begin{minipage}[c]{0.92\textwidth}
\includegraphics[width=0.23\textwidth]{L_Rmsp_CG_50000.png}%
\includegraphics[width=0.23\textwidth]{L_Rmsp_CG_150000.png}%
\includegraphics[width=0.23\textwidth]{L_Rmsp_CG_250000.png}%
\includegraphics[width=0.23\textwidth]{L_Rmsp_CG_300000.png}%
\end{minipage}}\\
Adam-ACG(ours)
\subfigure{
\begin{minipage}[c]{0.21\textwidth}
\includegraphics[width=1\textwidth]{L_Adam_CG_50000.png}%
\caption*{50k}
\end{minipage}}%
\subfigure{
\begin{minipage}[c]{0.21\textwidth}
\includegraphics[width=1\textwidth]{L_Adam_CG_150000.png}%
\caption*{150k}
\end{minipage}}%
\subfigure{
\begin{minipage}[c]{0.21\textwidth}
\includegraphics[width=1\textwidth]{L_Adam_CG_250000.png}%
\caption*{250k}
\end{minipage}}%
\subfigure{
\begin{minipage}[c]{0.21\textwidth}
\includegraphics[width=1\textwidth]{L_Adam_CG_300000.png}%
\caption*{300k}
\end{minipage}}%
\caption{Compared results of Linear GANs on MNIST dataset. Each row suggests a different compared method, and each column is the results of iteration number through $\{50000,150000,250000,300000\}$. This figure show SGD and Adam can not generate correct handwritten numbers. While RMSP and RMSP-ACG(ours), Adam-ACG(ours) can generate the handwritten digits. However, all the compared methods, including our proposed, are faced with the mode collapse problem.}
\label{fig9-1}
\end{figure}

\begin{figure}[H]\tiny
 SGD \qquad \quad \quad \quad
\subfigure{
\begin{minipage}[c]{0.92\textwidth}
\includegraphics[width=0.23\textwidth]{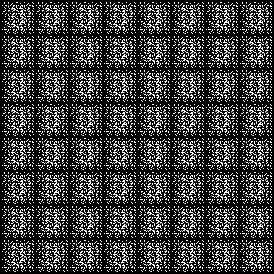}
\includegraphics[width=0.23\textwidth]{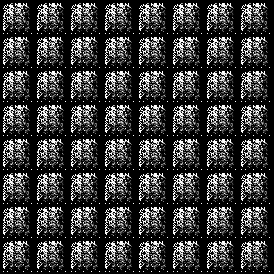}
\includegraphics[width=0.23\textwidth]{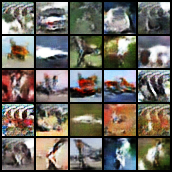}
\includegraphics[width=0.23\textwidth]{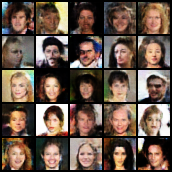}
\end{minipage}}\\
RMSP \qquad \quad \quad \;
\subfigure{
\begin{minipage}[c]{0.92\textwidth}
\includegraphics[width=0.23\textwidth]{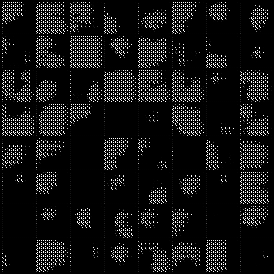}
\includegraphics[width=0.23\textwidth]{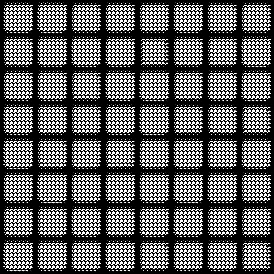}
\includegraphics[width=0.23\textwidth]{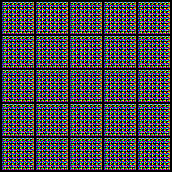}
\includegraphics[width=0.23\textwidth]{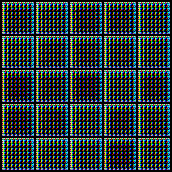}
\end{minipage}}\\
Adam \qquad \quad \quad \;
\subfigure{
\begin{minipage}[c]{0.92\textwidth}
\includegraphics[width=0.23\textwidth]{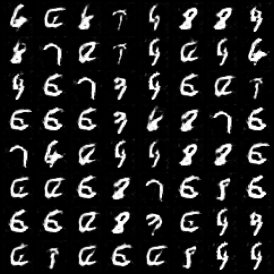}
\includegraphics[width=0.23\textwidth]{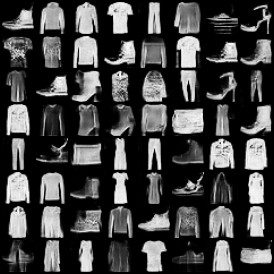}
\includegraphics[width=0.23\textwidth]{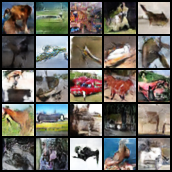}
\includegraphics[width=0.23\textwidth]{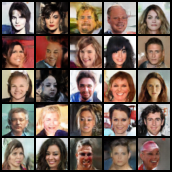}
\end{minipage}}\\
Adam-ACG(ours)
\subfigure{
\begin{minipage}[c]{0.21\textwidth}
\includegraphics[width=1\textwidth]{M_Adam-CG_100000.png}
\caption*{MNIST 100k}
\end{minipage}}%
\subfigure{
\begin{minipage}[c]{0.21\textwidth}
\includegraphics[width=1\textwidth]{F_Adam-CG_100000.png}
\caption*{Fashion-MNIST 100k}
\end{minipage}}%
\subfigure{
\begin{minipage}[c]{0.21\textwidth}
\includegraphics[width=1\textwidth]{CI_Adam-CG_80000.png}
\caption*{CIFAR10 80k}
\end{minipage}}%
\subfigure{
\begin{minipage}[c]{0.21\textwidth}
\includegraphics[width=1\textwidth]{C_Adam-CG_100000.png}
\caption*{CelebA 100k}
\end{minipage}}%
\caption{Comparison of DCGANs for several Algorithms on the four datasets. The first, second, third, and fourth rows are the results of SGD, RMSP, Adam, Adam-ACG(ours) on the four datasets. The first, second, third, and fourth columns are the results of the MNIST, Fashion-MNIST, CIFAR1O, and CelebA datasets, respectively. We conduct 100000 iterations for all experiments on the MNIST dataset, 100000 iterations for all experiments on the Fashion-MNIST dataset,80000 iterations for all experiments on the CIFAR10 dataset, and 100000 iterations for all experiments on the CelebA dataset, respectively. Thus, the SGD method is invalid on the MNIST and Fashion-MNIST datasets experiments. In contrast, SGD is valid on CIFAR10 and CelebA datasets experiments. Significantly, the RMSP method is invalid on the four datasets experiments. From this figure, our proposed method seems competitive to Adam on all four datasets experiments.}
\label{fig10-1}
\end{figure}

In this subsection, we mainly show more detailed figures in Section 6. Next, we describe all the figures and experiments settings. There is two figure in Figure \ref{fig5}, the left figure is the ground truth for the mixture of 5 Gaussians, and the right figure is the ground truth for the mixture of 16 Gaussians. We explore the influence of value $s$ through $\{50,100,150,200\}$ by our proposed method and the experimental results show in Figure \ref{fig8}. Figure \ref{fig6} is the comparison results of our proposed method and other SOTA algorithms on the mixture of 5 Gaussians experiments. Figure \ref{fig7-1} is the results of the compared methods in figure \ref{fig6} on the mixture of 16 Gaussians experiments. Figure \ref{fig9-1} is the experimental results of compared methods with linear GANs on the MNIST dataset. Figure \ref{fig10-1} is the experimental results of compared methods with DCGANs on the four Datasets (MNIST, Fashion-MNIST, CIFAR10, and CelebA). 
\vspace{-0.5cm}
\vspace{-0.5cm}

\end{document}